\documentclass[5p]{elsarticle}

\usepackage{graphicx}
\usepackage{caption}
\usepackage{subcaption}
\usepackage{amsmath}
\usepackage{algorithmic}
\usepackage{algorithm}
\usepackage{lineno}
\usepackage{multirow}
\usepackage{tabularx}
\usepackage{booktabs}
\usepackage{url}
\usepackage{color}
\usepackage[figuresright]{rotating}
\usepackage[utf8]{inputenc}
\usepackage[T1]{fontenc}

\graphicspath {{figures/}}
\newcommand{\tabhead}[1]{\textbf{#1}}

\journal{Knowledge-Based Systems}

\begin{document}

\begin{frontmatter}

\title{Combined Cleaning and Resampling Algorithm for Multi-Class Imbalanced Data with Label Noise}

\author[agh]{Micha\l{} Koziarski\corref{cor1}}
\ead{michal.koziarski@agh.edu.pl}
\cortext[cor1]{Corresponding author}

\author[pwr]{Micha\l{} Wo\'zniak}
\ead{michal.wozniak@pwr.edu.pl}

\author[vcu]{Bartosz Krawczyk}
\ead{bkrawczyk@vcu.edu}

\address[agh]{Department of Electronics, AGH University of Science and Technology, Al. Mickiewicza 30, 30-059 Krak\'ow, Poland}

\address[pwr]{Department of Systems and Computer Networks, Wroc\l{}aw University of Science and Technology, Wybrze\.ze Wyspia\'nskiego 27, 50-370 Wroc\l{}aw, Poland}

\address[vcu]{Department of Computer Science, School of Engineering, Virginia Commonwealth University, 401 West Main Street, P.O. Box 843019, Richmond, Virginia 23284-3019, USA}

\begin{abstract}
The imbalanced data classification is one of the most crucial tasks facing modern data analysis. Especially when combined with other difficulty factors, such as the presence of noise, overlapping class distributions, and small disjuncts, data imbalance can significantly impact the classification performance. Furthermore, some of the data difficulty factors are known to affect the performance of the existing oversampling strategies, in particular SMOTE and its derivatives. This effect is especially pronounced in the multi-class setting, in which the mutual imbalance relationships between the classes complicate even further. Despite that, most of the contemporary research in the area of data imbalance focuses on the binary classification problems, while their more difficult multi-class counterparts are relatively unexplored. In this paper, we propose a novel oversampling technique, a \emph{Multi-Class Combined Cleaning and Resampling} (MC-CCR) algorithm. The proposed method utilizes an energy-based approach to modeling the regions suitable for oversampling, less affected by small disjuncts and outliers than SMOTE. It combines it with a simultaneous cleaning operation, the aim of which is to reduce the effect of overlapping class distributions on the performance of the learning algorithms. Finally, by incorporating a dedicated strategy of handling the multi-class problems, MC-CCR is less affected by the loss of information about the inter-class relationships than the traditional multi-class decomposition strategies. Based on the results of experimental research carried out for many multi-class imbalanced benchmark datasets, the high robust of the proposed approach to noise was shown, as well as its high quality compared to the state-of-art methods.
\end{abstract}


\end{frontmatter}


\section{Introduction}
\label{sec:int}

The presence of data imbalance can significantly impact the performance of traditional learning algorithms \cite{Branco:2016}. The disproportion between the number of majority and minority observations influences the process of optimization concerning a zero-one loss function, leading to a bias towards the majority class and accompanying degradation of the predictive capabilities for the minority classes. While the problem of data imbalance is well established in the literature, it was traditionally studied in the context of binary classification problems, with the sole goal of reducing the degree of imbalance. However, recent studies point to the fact that it is not the imbalanced data itself, but rather other data difficulty factors, amplified by the data imbalance, that pose a challenge during the learning process \cite{stefanowski2016dealing,Fernandez:2018}. Such factors include small sample size, presence of disjoint and overlapping data distributions, and presence of outliers and noisy observations.

Furthermore, yet another important and often overlooked aspect is a multi-class nature of many classification problems, that can additionally amplify the challenges associated with the imbalanced data classification \cite{Krawczyk:2016}. For the two-class classification task determining relationships between classes is relatively simple. In the case of a multi-class task, the relationships mentioned are definitely more complex \cite{Wang:2012}. Developed classifiers dedicated to two-class problems cannot be easily adapted to multi-class tasks mainly because they are unable to model relationships among classes and the difficulties built into the multi-class problem, such as the occurrence of borderline objects among more than two classes, or multiple class overlapping. Many suggestions focus on decomposing multi-class tasks into binary ones, however, such a simplification of the multi-class imbalanced classification problem leads to the loss of valuable information about relationships among more than a selected pair of classes \cite{Krawczyk:2016,Fernandez:2018}.
This paper introduces a novel algorithm named \emph{Multi-Class Combined Cleaning and Resampling} (MC-CCR) to alleviate the identified drawbacks of the existing algorithms. MC-CCR is developed with the aim of handling the imbalanced problems with embedded data-level difficulties, i.e., atypical data distributions, overlapping classes, and small disjuncts, in the multi-class setting. The main strength of MC-CCR lies in the idea of originally proposed decomposition strategy and applying an idea of cleaning the neighborhood of minority class examples and generating new synthetic objects there. Therefore, we make an important step towards a new view on the oversampling scheme, by showing that utilizing information coming from all of the classes is highly beneficial. Our proposal is trying to depart from traditional methods based on the use of nearest neighbors to generate synthetic learning instances. Thanks to which we reduce the impact of existing algorithms’ drawbacks, and we enable smart oversampling of multiple classes in the guided manner.

To summarize, this work makes the following contributions:
\begin{itemize} 
\item Proposition of the \emph{Multi-Class Combined Cleaning and Resampling} algorithm, which allows for intelligent data oversampling that exploits local data characteristics of each class and is robust to atypical data distributions.

\item Utilization of the information about the inter-class relationships in the multi-class setting during the artificial instance generation procedure that offers better placement of new instances and more targeted empowering of minority classes.


\item Explanation of how the constraining of the oversampling using the proposed energy-based approach, as well as the guided cleaning procedure, alleviate the drawbacks of the SMOTE-based methods.

\item Presenting capabilities of the proposed method to handle challenging imbalanced data with label noise presence.

\item {Detailed analysis of computational complexity of our method, showcasing its reliable trade-off between preprocessing time and obtained improvements in handling imbalanced data.}

\item Experimental evaluation of the proposed approach based on diverse benchmark datasets and a detailed comparison with the state-of-art approaches.
\end{itemize} 

The paper is organized as follows. The next section discusses in detail the problem of learning from noisy and imbalanced data, as well it also emphasizes the unique characteristics of multi-class problems. Section~\ref{sec:ccr} introduces MC-CCR in details, while Section~\ref{sec:exp} depicts the conducted experimental study. The final section concludes the paper and offers insight into future directions in the field of multi-class imbalanced data preprocessing.

\section{Learning from imbalanced data}
\label{sec:imb}

In this section, we discuss the difficulties mentioned above, starting with an overview of binary imbalanced problems, and later progressing to the multi-class classification task and label noise.

\subsection{Binary imbalanced problems}
\label{sec:bin}

The strategies for dealing with data imbalance can be divided into two categories. First of all, the data-level methods: algorithms that perform data preprocessing with the aim of reducing the imbalance ratio, either by decreasing the number of majority observations (undersampling) or increasing the number of minority observations (oversampling). After applying such preprocessing, the transformed data can be later classified using traditional learning algorithms.

By far, the most prevalent data-level approach is SMOTE \cite{chawla2002smote} algorithm. It is a guided oversampling technique, in which synthetic minority observations are being created by interpolation of the existing instances. It is nowadays considered a cornerstone for the majority of the following oversampling methods \cite{Perez-Ortiz:2016,Bellinger:2018}. However, due to the underlying assumption about the homogeneity of the clusters of minority observations, SMOTE can inappropriately alter the class distribution when factors such as disjoint data distributions, noise, and outliers are present, which will be later demonstrated in Section~\ref{sec:ccr}. Numerous modifications of the original SMOTE algorithm have been proposed in the literature. The most notable include Borderline SMOTE \cite{Han:2005}, which focuses on the process of synthetic observation generation around the instances close to the decision border; Safe-level SMOTE~\cite{Bunkhumpornpat:2009} and LN-SMOTE~\cite{Maciejewski:2011}, which aim to reduce the risk of introducing synthetic observations inside regions of the majority class; and ADASYN \cite{He:2008}, that prioritizes the difficult instances.

The second category of methods for dealing with data imbalance consists of algorithm-level solutions. These techniques alter the traditional learning algorithms to eliminate the shortcomings they display when applied to imbalanced data problems. Notable examples of algorithm-level solutions include: kernel functions \cite{Mathew:2018}, splitting criteria in decision trees \cite{Li:2018}, and modifications off the underlying loss function to make it cost-sensitive \cite{Khan:2018}. However, contrary to the data-level approaches, algorithm-level solutions necessitate a choice of a specific classifier. Still, in many cases, they are reported to lead to a better performance than sampling approaches \cite{Fernandez:2018}.

\subsection{Multi-class imbalanced problems}
\label{sec:mci}

While in the binary classification, one can easily define the majority and the minority class, as well as quantify the degree of imbalance between the classes. This relationship becomes more convoluted when transferring to the multi-class setting. One of the earlier proposals for the taxonomy of multi-class problems used either the concept of multi-minority, a single majority class accompanied by multiple minority classes or multi-majority, a single minority class accompanied by multiple majority classes \cite{Wang:2012}. However, in practice, the relationship between the classes tends to be more complicated, and a single class can act as a majority towards some, a minority towards others, and have a similar number of observations to the rest of the classes. Such situations are not well-encompassed by the current taxonomies. Since categorizations such as the one proposed by Napierała and Stefanowski \cite{napierala2016types} played an essential role in the development of specialized strategies for dealing with data imbalance in the binary setting, the lack of a comparable alternative for the multi-class setting can be seen as a limiting factor for the further research.

The difficulties associated with the imbalanced data classification are also further pronounced in the multi-class setting, where each additional class increases the complexity of the classification problem. This includes the problem of overlapping data distributions, where multiple classes can simultaneously overlap a particular region, and the presence of noise and outliers, where on one hand a single outlier can affect class boundaries of several classes at once, and on the other can cease to be an outlier where some of the classes are excluded. Finally, any data-level observation generation or removal must be done by a careful analysis of how action on a single class influences different types of observations in remaining classes. All of the above lead to a conclusion that algorithms designed explicitly to handle the issues associated with multi-class imbalance are required to adequately address the problem.

The existing methods for handling multi-class imbalance can be divided into two categories. First of all, the binarization solutions, which decompose a multi-class problem into either $M(M-1)/2$ (one-vs-one, OVO) or $M$ (one-vs-all, OVA) binary sub-problems \cite{Fernandez:2013}. Each sub-problem can then be handled individually using a selected binary algorithm. An obvious benefit of this approach is the possibility of utilization of existing algorithms \cite{Zhang:2018}. However, binarization solutions have several significant drawbacks.

Most importantly, they suffer from the loss of information about class relationships. In essence, we either completely exclude the remaining classes in a single step of OVO decomposition or discard the inner-class relations by merging classes into a single majority in OVA decomposition. Furthermore, especially in the case of OVO decomposition associated computational cost can quickly grow with the number of classes and observations, making the approach ill-suited for dealing with the big data. Among the binarization solutions, the recent literature suggests the efficacy of using ensemble methods with OVO decomposition \cite{Zhang:2016}, augmenting it with cost-sensitive learning \cite{Krawczyk:2016ijcnn}, or applying dedicated classifier combination methods \cite{Japkowicz:2015}.

The second category of methods consists of ad-hoc solutions: techniques that treat the multi-class problem natively, proposing dedicated solutions for exploiting the complex relationships between the individual classes. Ad-hoc solutions require either a significant modification to the existing algorithms, or exploring a completely novel approaches to overcoming the data imbalance, both on the data and the algorithm level. However, they tend to significantly outperform binarization solutions, offering a promising direction for further research. Most-popular data-level approaches include extensions of the SMOTE algorithm into a multi-class setting \cite{Fernandez-Navarro:2011,Saez:2016,Zhu:2017}, strategies using feature selection \cite{Cao:2017,Wu:2017}, and alternative methods for instance generation by using Mahalanobis distance \cite{Abdi:2016,Yang:2017}. Algorithm-level solutions include decision tree adaptations \cite{Hoens:2012}, cost-sensitive matrix learning \cite{Bernard:2016}, and ensemble solutions utilizing Bagging \cite{Lango:2018,Collell:2018} and Boosting \cite{Wang:2012,Guo:2016}.

\subsection{Metrics for multi-class imbalance task}
\label{sec:metrics}
One of the important problems related to imbalanced data classification is the assessment of the predictive performance of the developed algorithms. It is obvious that in the case of imbalanced data, we cannot use \emph{Accuracy}, which prefers classes with higher prior probabilities. Currently, many metrics dedicated to imbalanced data tasks have been proposed for both binary and multi-class problems. 
Branco et al. \cite{Branco:2017} reported the following metrics which may be used in multi-class imbalanced data classification task: Average Accuracy (\emph{AvAcc}), Class Balance Accuracy (\emph{CBA}), multi-class G-measure (\emph{mGM}), and Confusion Entropy (\emph{CEN}). They are expressed as follows:

\begin{align}
AvAcc &= \frac{\sum_{i=1}^M TPR_i}{M}\\
CBA &= \frac{\sum_{i=1}^{M}\frac{mat_{i,i}}{max\left( \sum_{j=1}^{M} mat_{i,j},\sum_{j=1}^{M} mat_{j,i} \right)}}{M}\\
mGM &= \sqrt[M]{\prod_{i=1}^M recall_i}\\
CEN &= \sum_{i=1}^{M} P_i \cdot CEN_i,\\
\end{align}

\noindent where $M$ is the number of classes, $mat_{i,j}$ stands for the number of instances of the true class $i$ that were predicted as class $j$, 
\begin{align*}
P_i = \frac{\sum_{j=1}^{M} mat_{i,j} + mat_{j,i}}{2 \cdot \sum_{i,l=1}^{M} mat_{k,l}}, \end{align*}

\noindent and 

\begin{align*}
CEN_i&= \\&- \sum_{j=1, i\neq j}^{M} \left( P_{i,j}^{i} \text{log}_{2(C-1)}(P_{i,j}^{i}) + P_{j,i}^{i} \text{log}_{2(C-1)}(P_{j,i}^{i})\right)
\end{align*}

\noindent Additionally, for $CEN$ we have 

\begin{align*}    
P_{i,i}^{i}&= 0, \\
P_{i,j}^{i}&= \frac{mat_{i,j}}{\left( \sum_{j=1}^C mat_{i,j} + mat_{j,i} \right)},\\ \text{ and }& i \neq j. 
\end{align*}

It is also worth mentioning that choosing the right metrics is still an open problem. Currently, many works show that previously considered metrics may prefer majority classes, especially in the case of the so-called parametric metrics (e.g., $IBA_{\alpha}$ or $Fscore_{\beta}$) \cite{Brzezinski:2018,Brzezinski:2019}.

\subsection{Class label noise in the imbalanced problems}
\label{sec:noi}

Machine learning algorithms depend on the data, and for many problems, such as the classification task, they require labeled data. Therefore, the high quality labeled learning set is an important factor in building a high-quality predictive system. One of the most serious problems in data analysis is data noise. It can have a dual nature. On the one hand, it may relate to noise caused by a human operator (incorrect imputation) or measurement errors when acquiring attribute values. On the other hand, it may relate to incorrect data labels. In this work, we will examine the robustness of the proposed solution to label noise. This type of noise occurs whenever an observation is assigned incorrect label \cite{Zhu:2003b}, and can lead to the formation of contradictory learning instances: duplicate observations having different class label \cite{Hernandez:1998}. Some works have reported this problem \cite{scott2013classification, garcia2015effect}, including a survey by Fr\'{e}nay and Verleysen \cite{Frenay:2014}. However, relatively few papers are devoted to the impact of noise on the predictive performance of imbalanced data classifiers, in which label noise can become the most problematic.
Let’s firstly consider where the labels come from. The most common case is obtaining labels from human experts. Unfortunately, man is not infallible, e.g., considering the quality of medical diagnostics, we may conclude that the number of errors made by human experts is noticeable \cite{Donaldson:2000}. Another problem is the fact that the distribution of errors committed by experts is not uniform, because labeling may be subjective. After all, human experts may be biased.
Another approach is obtaining labels from non-experts as crowdsourcing provides a scalable and efficient way to construct labeled datasets for training machine learning systems. However, creating comprehensive label guidelines for crowd workers is often hard, even for seemingly simple concepts. Incomplete or ambiguous label guidelines can then result in differing interpretations of concepts and inconsistent labels.
Another reason for the noise in labels is data corruption \cite{Chang:2017}, which may be due to, e.g., \emph{data poisoning} \cite{Li:2016}. Both natural and malicious label corruptions tend to degrade the performance of classification systems sharply \cite{Hendrycks:2018}.
As mentioned, the distribution of label errors can have a different nature, usually dependant on their source. One can highlight the label noise that is:
\begin{itemize}
\item a completely random label noise,
\item a random label noise dependant on the true label (asymmetrical label noise),
\item label noise is not random, but depends on the true label and features.
\end{itemize}

There are many methods of dealing with label noise. One of the most popular ways is \emph{data cleaning}. An example of this solution is the use of SMOTE oversampling with cleaning using the \emph{Edited Nearest Neighbors} (ENN) \cite{batista2004study}. This approach keeps the total relatively high number of observations, and the number of mislabeled observations relatively low, allowing to detect improper labeling examples. Nevertheless, when we deal with feature space regions, which is common for imbalanced data analysis tasks, then the distinction between outliers and improperly labeled observations becomes problematic or even impossible.
Designing a \emph{label noise-tolerant} learning classification algorithm is another approach. Usually, works in this area assume a model of label noise distribution and analyze the viability of learning under this model. An example of this approach is presented by Angluin and Laird as a \emph{Class-conditional noise} model (CCN) \cite{Angluin:1998}. Finally, the last approach is designing a \emph{label noise-robust} classifier, which, even in the case when data denoising does not occur, nor any noise is modeled, still produces a model that has a relatively good predictive performance when the learning set is slightly noisy \cite{Frenay:2014}.

\section{MC-CCR: Multi-Class Combined Cleaning and Resampling algorithm}
\label{sec:ccr}

To address the difficulties associated with the classification of noisy and multi-class data, we propose a novel oversampling approach, \emph{Multi-Class Combined Cleaning and Resampling} algorithm (MC-CCR). In the remainder of this section, we begin with a description of the binary variant of the \emph{Combined Cleaning and Resampling} (CCR) and discuss its behavior in the presence of label noise. Afterward we introduce the decomposition strategy used to extend the CCR to the multi-class setting. Finally, we conduct a computation complexity analysis of the proposed algorithm.

\subsection{Binary Combined Cleaning and Resampling}
\label{sec:b-ccr}

The CCR algorithm was initially introduced by Koziarski and Woźniak \cite{Koziarski:2017} in the context of binary classification problems. It was based on two empirical observations: firstly, that data imbalance by itself does not negatively impact the classification performance. Only when combined with other data difficulty factors, such as decomposition of the minority class into rare sub-concepts and overlapping of classes, the data imbalance poses a difficulty for the traditional learning algorithms due to the amplification of the factors mentioned above \cite{stefanowski2016dealing}. Secondly, that when optimizing the classification performance concerning the metrics accounting for the data imbalance, it is often beneficial to sacrifice some of the precision to achieve a better recall of the predictions, possibly to a more significant extent than typical over- or undersampling algorithms. Based on these two observations, an algorithm combining the steps of cleaning the neighborhoods of the minority instances and selectively oversampling the minority class was proposed.

\noindent\textbf{Cleaning the minority neighborhoods.} As a step preceding the oversampling itself, we propose performing a data preprocessing in the form of cleaning the majority observations located in proximity to the minority instances. The aim of such an operation is twofold. First of all, to reduce the problem of class overlap: by designing the regions from which majority observations are being removed, we transform the original dataset intending to simplify it for further classification. Secondly, to skew the classifiers' predictions towards the minority class: while in the case of the imbalanced data such regions, bordering two-class distributions or consisting of overlapping instances, tend to produce predictions biased towards the majority class. By performing clean-up, we either reduce or reverse this trend.

Two key components of such cleaning operation are a mechanism of the designation of regions from which the majority observations are to be removed, and a removal procedure itself. The former, especially when dealing with data affected by label noise, should be able to adapt to the surroundings of any given minority observation, and adjust its behavior depending on whether the observation resembles a mislabeled instance or a legitimate outlier from an underrepresented region, which is likely to occur in the case of imbalanced data with scarce volume. The later should limit the loss of information that could occur due to the removal of a large number of majority observations.

\begin{figure}
\centering
\includegraphics[width=0.75\linewidth]{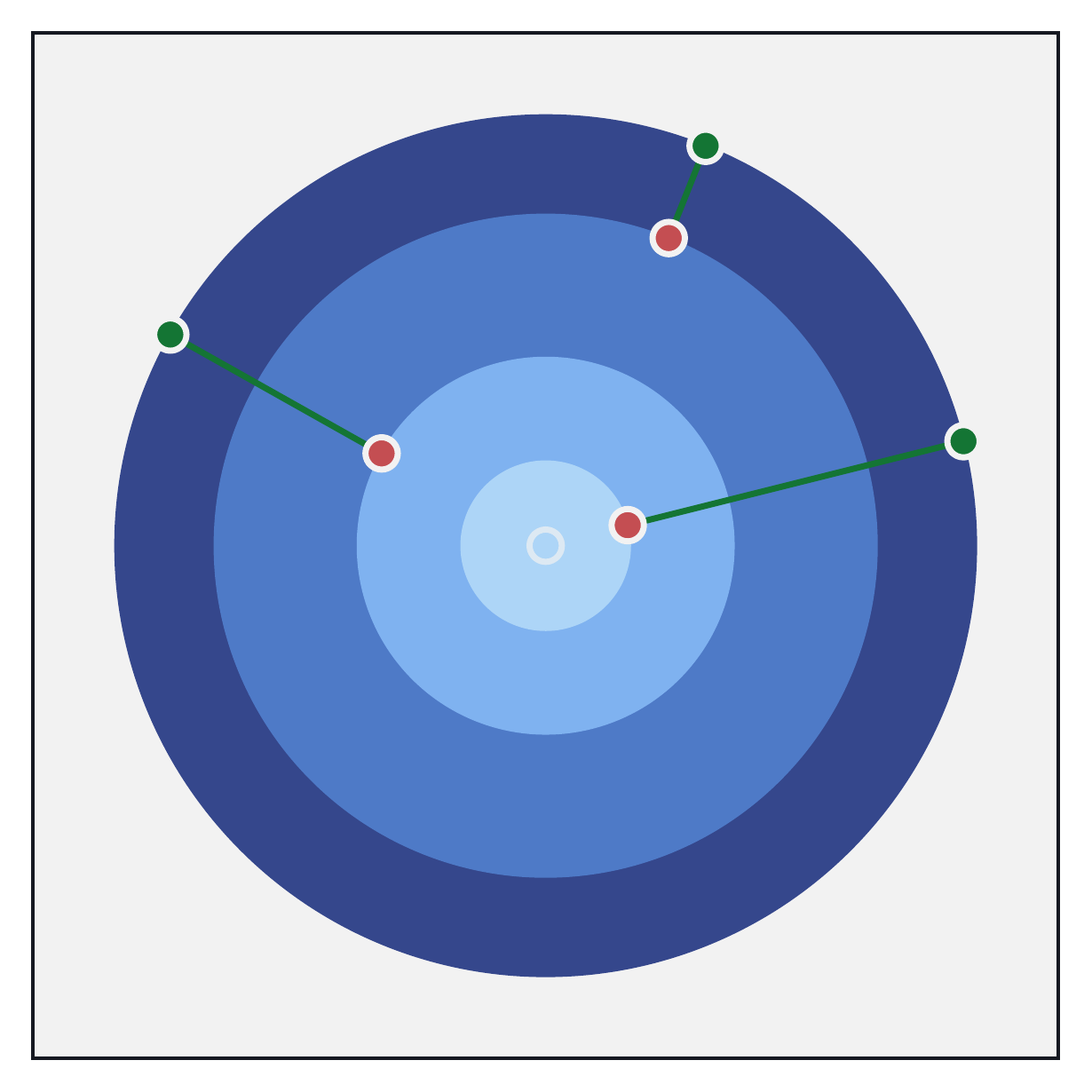}
\caption{An illustration of the sphere creation for an individual minority observation (in the center) surrounded by majority observations (in red). Sphere expends at a normal cost until it reaches a majority observation, at which point the further expansion cost increases (depicted by blue orbits with an increasingly darker color). Finally, after the expansions, the majority observations within the sphere are being pushed outside (in green).}
\label{fig:cleaning}
\end{figure}

To implement such preprocessing in practice, we propose an energy-based approach, in which spherical regions are constructed around every minority observation. Spheres expand using the available energy, a parameter of the algorithm, with the cost increasing for every majority observation encountered during the expansion. More formally, for a given minority observation denoted by $x_i$, current radius of an associated sphere denoted by $r_i$, a function returning the number of majority observations inside a sphere centered around $x_i$ with radius $r$ denoted by $f_n(r)$, a target radius denoted by $r_i'$, and $f_n(r_i') = f_n(r_i) + 1$, we define the energy change caused by the expansion from $r_i$ to $r_i'$ as
\begin{equation}
    \Delta e = - (r_i' - r_i) \cdot f_n(r_i').
\end{equation}
During the sphere expansion procedure, the radius of a given sphere increases up to the point of completely depleting the energy, with the cost increases after each encountered majority observation. Finally, the majority observations inside the sphere are being pushed out to its outskirts. The whole process was illustrated in Figure~\ref{fig:cleaning}.

The proposed cleaning approach meets both of the outlined criteria. First of all, due to the increased expansion cost after each encountered majority observation, it distinguishes the likely mislabeled instances: minority observations surrounded by a large number of majority observations lead to a creation of smaller spheres and, as a result, more constrained cleaning regions. On the other hand, in case of overlapping class distributions, or other words in the presence of a large number of both minority and majority observations, despite the small size of individual spheres, their large volume still leads to large cleaning regions. Secondly, since the majority observations inside the spheres are being translated instead of being completely removed, the information associated with their original positions is to a large extent preserved, and the distortion of class density in specific regions is limited.

\noindent\textbf{Selectively oversampling the minority class.} After the cleaning stage is concluded, new synthetic minority observations are being generated. To further exploit the spheres created during the cleaning procedure, new synthetic instances are being sampled within the previously designed cleaning regions. This not only prevents the synthetic observations from overlapping the majority class distribution but also constraints the oversampling areas for observations displaying the characteristics of mislabeled instances. 

Moreover, in addition to designating the oversampling regions, we propose employing the size of the calculated spheres in the process of weighting the selection of minority observations used as the oversampling origin. Analogous to the ADASYN \cite{he2008adasyn}, we focus on the difficult observations, with difficulty estimated based on the radius of an associated sphere. More formally, for a given minority observation denoted by $x_i$, the radius of an associated sphere denoted by $r_i$, the vector of all calculated radii denoted by $r$, collection of majority observations denoted by $\mathcal{X}_{maj}$, collection of minority observations denoted by $\mathcal{X}_{min}$, and assuming that the oversampling is performed up to the point of achieving balanced class distribution, we define the number of synthetic observations to be generated around $x_i$ as
\begin{equation}
    \label{eq:prop}
    g_i = \lfloor\dfrac{r_i^{-1}}{\sum_{k = 1}^{|\mathcal{X}_{min}|}{r_k^{-1}}} \cdot (|\mathcal{X}_{maj}| - |\mathcal{X}_{min}|)\rfloor.
\end{equation}
Just like in the ADASYN, such weighting aims to reduce the bias introduced by the class imbalance and to shift the classification decision boundary toward the difficult examples adaptively. However, compared to the ADASYN, in the proposed method the relative distance of the observations plays an important role: while in the ADASYN outlier observations, located in a close proximity of neither majority nor minority instances, based on their far-away neighbors could be categorized as difficult, that is not the case under the proposed weighting, where the full sphere expansion would occur.

\begin{figure*}
\centering
        \includegraphics[width=0.24\textwidth]{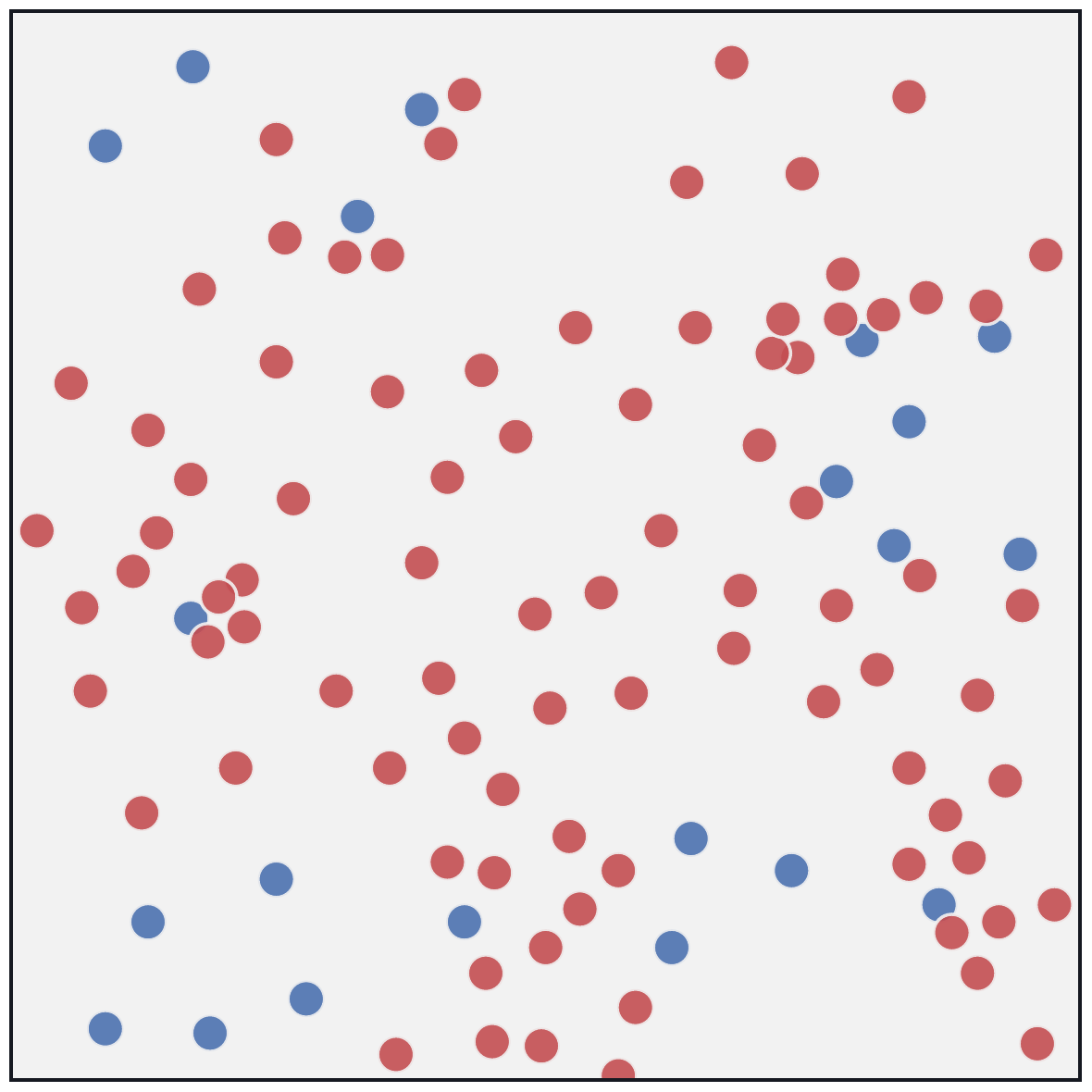}
        \includegraphics[width=0.24\textwidth]{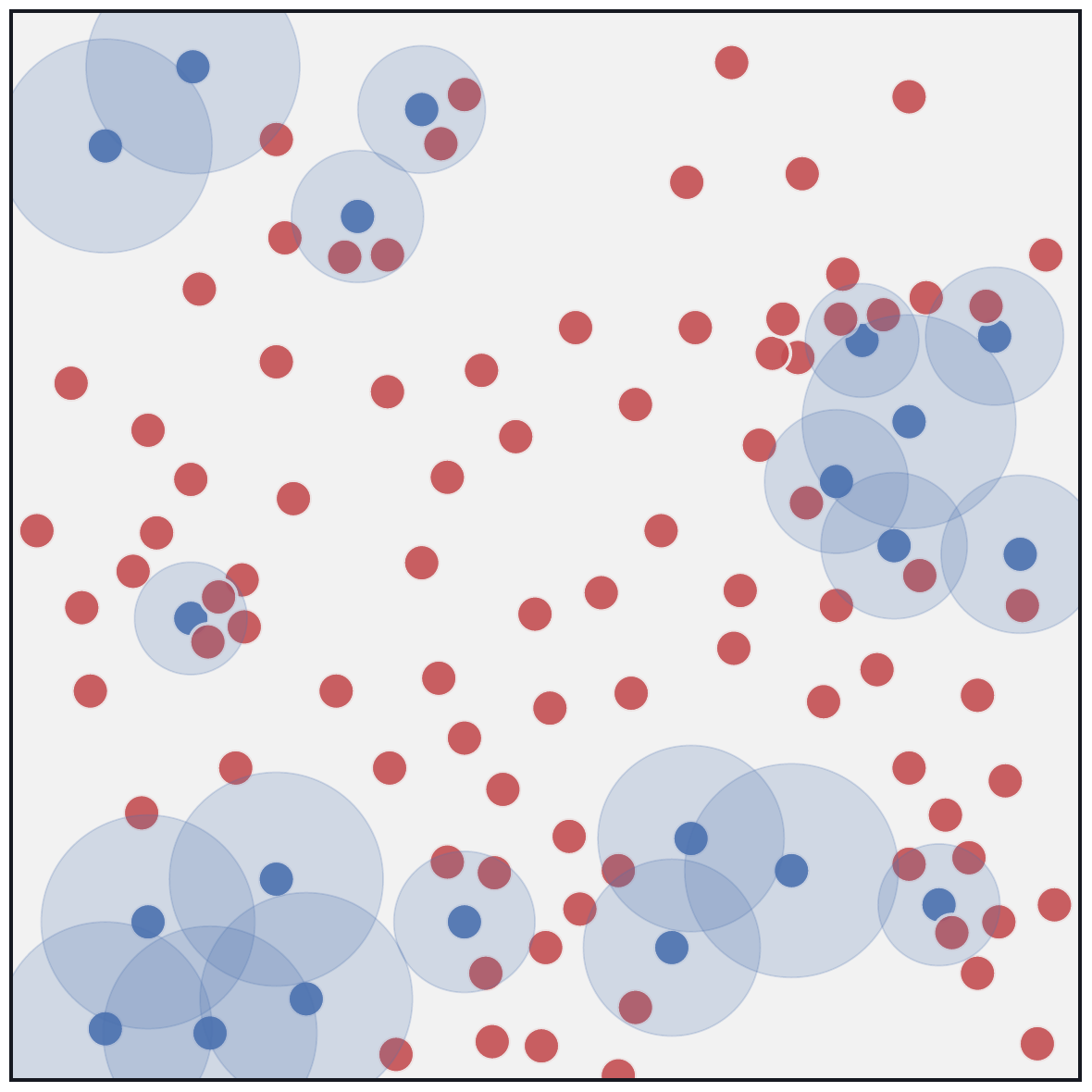}
        \includegraphics[width=0.24\textwidth]{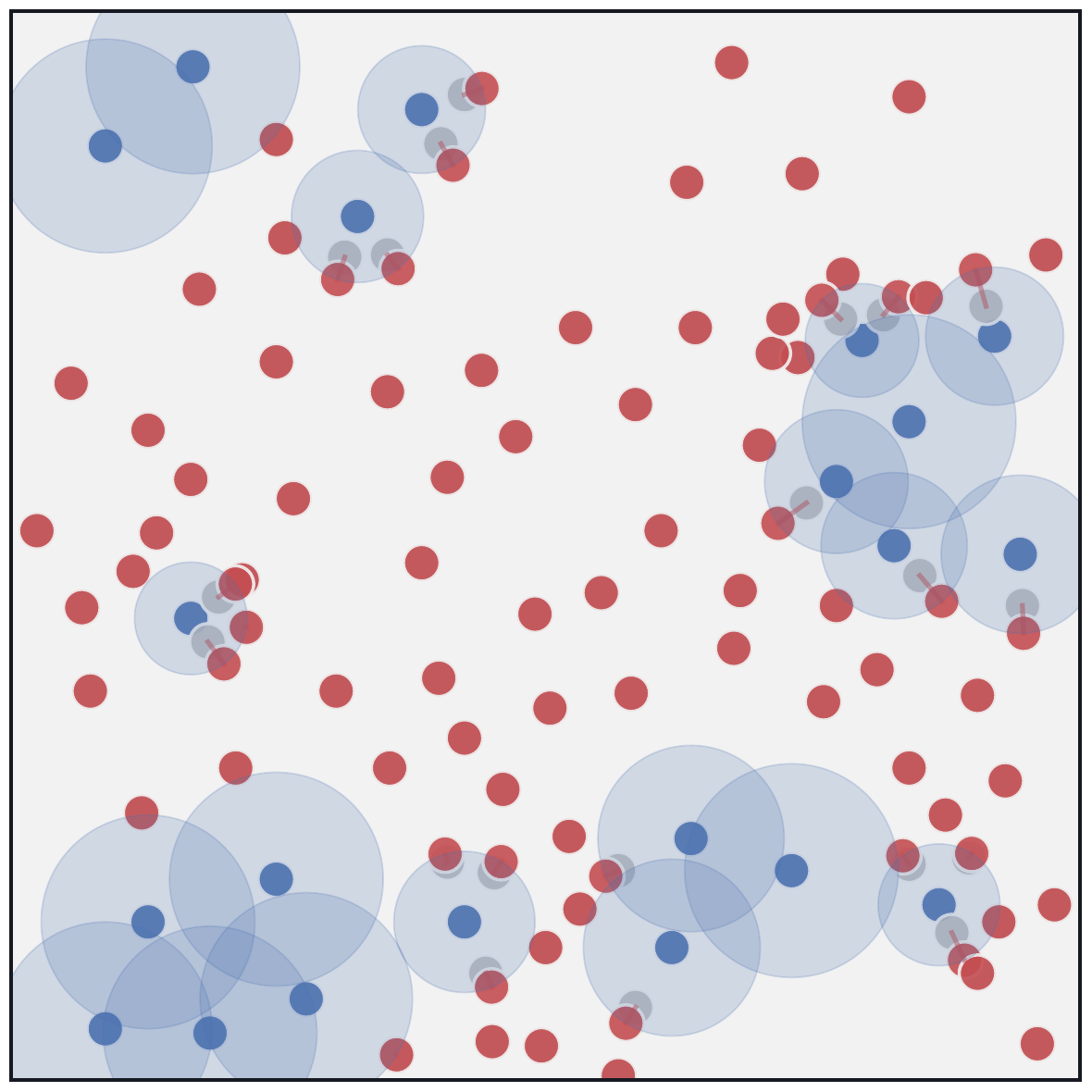}
        \includegraphics[width=0.24\textwidth]{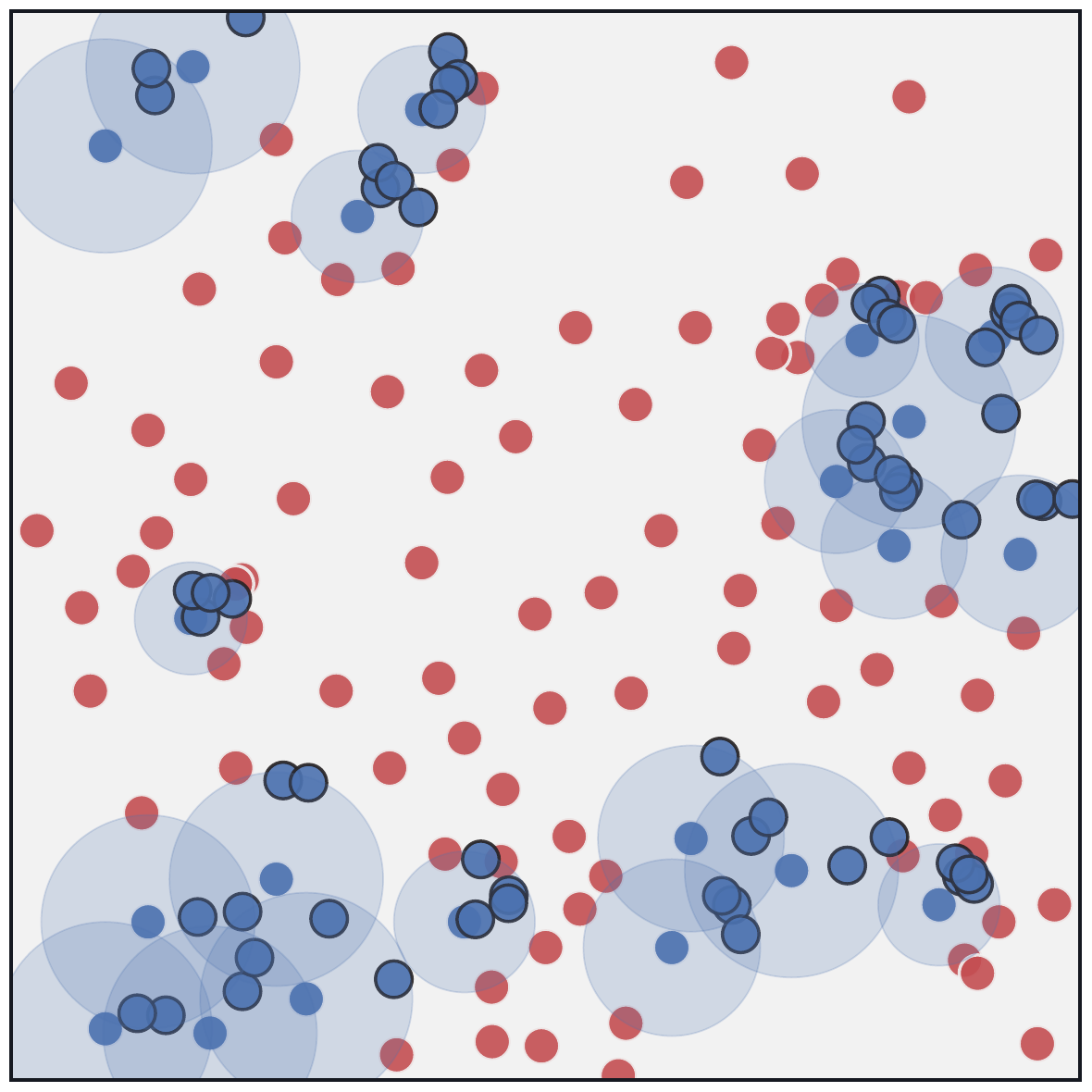}
\caption{An illustration of the algorithms behavior in a binary case. From the left: 1) original dataset, 2) sphere calculation for individual minority objects, with smaller spheres created for the observations surrounded by the majority objects, 3) pushing out the majority objects outside the sphere radius, 4) generating synthetic minority observations inside the spheres, in the number inversely proportional to the sphere radius.}
\label{fig:b-ccr}
\end{figure*}

\noindent\textbf{Combined algorithm.} We present complete pseudocode of the proposed method in Algorithm~\ref{algorithm:b-ccr}. Furthermore, we illustrate the behavior of the algorithm in a binary case in Figure~\ref{fig:b-ccr}. We outline all three major stages of the proposed procedure: forming spheres around the minority observations, clean-up of the majority observations inside the spheres, and adaptive oversampling based on the sphere radii.

\begin{algorithm}[!htb]
	\caption{Binary Combined Cleaning and Resampling}
	\label{algorithm:b-ccr}
	\begin{algorithmic}[1]
		\STATE \textbf{Input:} collections of majority observations $\mathcal{X}_{maj}$ and minority observations $\mathcal{X}_{min}$
		\STATE \textbf{Parameters:} $energy$ budget for expansion of each sphere, $p$-norm used for distance calculation
		\STATE \textbf{Output:} collections of translated majority observations $\mathcal{X}_{maj}'$ and synthetic minority observations $S$
		\STATE
		\STATE \textbf{function} CCR($\mathcal{X}_{maj}$, $\mathcal{X}_{min}$, $energy$, $p$):
		\STATE $S \gets \emptyset$ \COMMENT{synthetic minority observations}
		\STATE $t \gets $ zero matrix of size $|\mathcal{X}_{maj}| \times m$, with $m$ denoting the number of features \COMMENT{translations of majority observations}
		\STATE $r \gets $ zero vector of size $|\mathcal{X}_{min}|$ \COMMENT{radii of spheres associated with the minority observations}
        \FORALL{minority observations $x_i$ in $\mathcal{X}_{min}$}
        \STATE $e$ $\gets$ $energy$ \COMMENT{remaining energy budget}
        \STATE $n_r \gets 0$ \COMMENT{number of majority observations inside the sphere generated around $x_i$}
        \FORALL{majority observations $x_j$ in $\mathcal{X}_{maj}$}
        \STATE $d_j \gets \lVert x_i - x_j \rVert_p$
        \ENDFOR
        \STATE sort $\mathcal{X}_{maj}$ with respect to $d$
        \FORALL{majority observations $x_j$ in $\mathcal{X}_{maj}$}
        \STATE $n_r \gets n_r + 1$
        \STATE $\Delta e \gets - (d_j - r_i) \cdot n_r$
        \IF{$e + \Delta e > 0$}
        \STATE $r_i \gets d_j$
        \STATE $e \gets e + \Delta e$
        \ELSE
        \STATE $r_i \gets r_i + \frac{e}{n_r}$
        \STATE \textbf{break}
        \ENDIF
        \ENDFOR
        \FORALL{majority observations $x_j$ in $\mathcal{X}_{maj}$}
        \IF{$d_j < r_i$}
        \STATE $t_j \gets t_j + \dfrac{r_i - d_j}{d_j} \cdot (x_j - x_i)$
        \ENDIF
        \ENDFOR
        \ENDFOR
        \STATE $\mathcal{X}_{maj}' \gets \mathcal{X}_{maj} + t$
        \FORALL{minority observations $x_i$ in $\mathcal{X}_{min}$}
        \STATE $g_i \gets \lfloor\dfrac{r_i^{-1}}{\sum_{k = 1}^{|\mathcal{X}_{min}|}{r_k^{-1}}} \cdot (|\mathcal{X}_{maj}| - |\mathcal{X}_{min}|)\rfloor$ 
        \FOR{1 \textbf{to} $g_i$}
        \STATE $v \gets$ random point inside a zero-centered sphere with radius $r_i$
		\STATE $S \gets S \cup \{x_i + v\}$
        \ENDFOR
        \ENDFOR
		\STATE \textbf{return} $\mathcal{X}_{maj}'$, $S$
	\end{algorithmic}
\end{algorithm}

\subsection{Multi-Class Combined Cleaning and Resampling}
\label{sec:mc-ccr}

\begin{figure*}
\centering
        \includegraphics[width=0.16\textwidth]{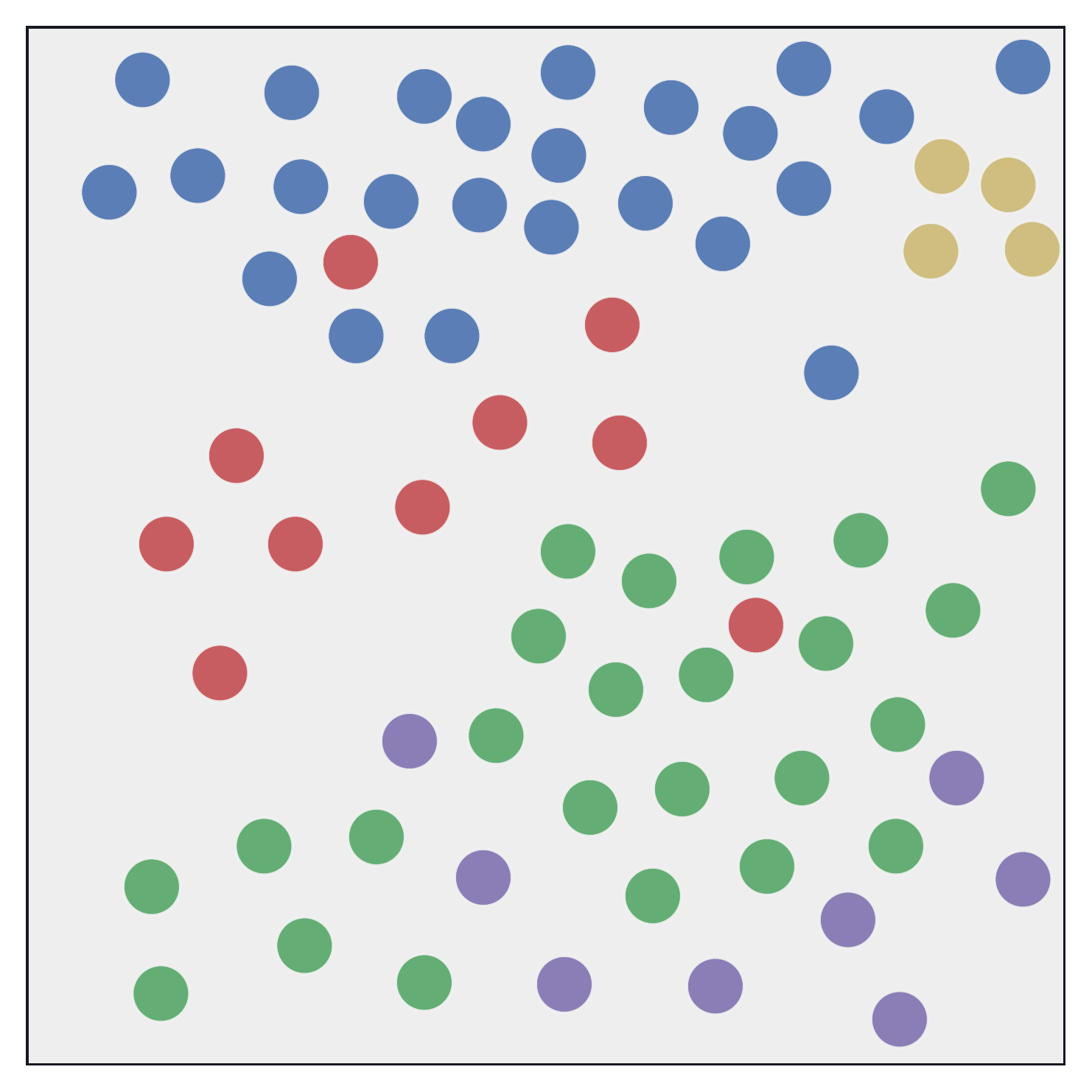}
        \includegraphics[width=0.16\textwidth]{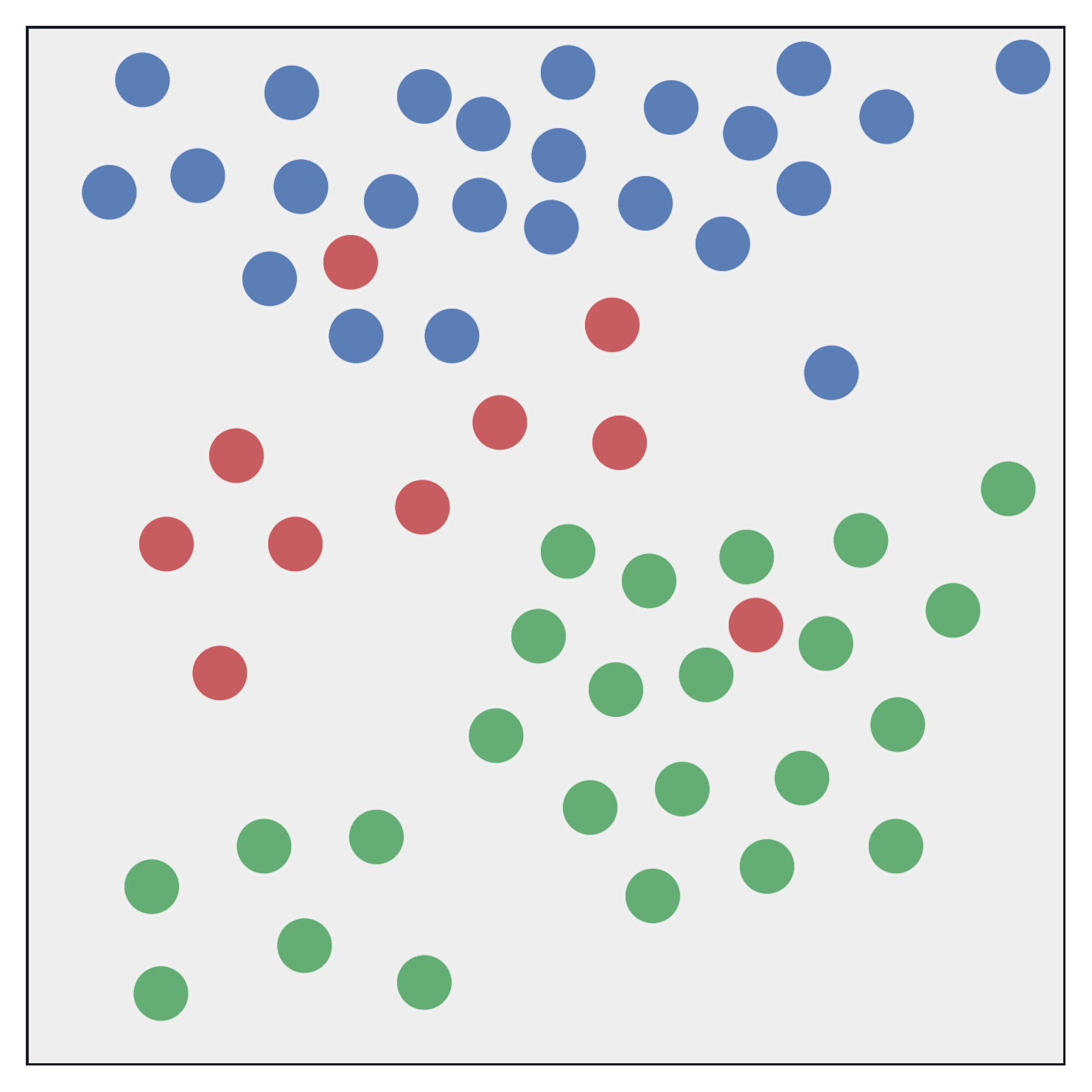}
        \includegraphics[width=0.16\textwidth]{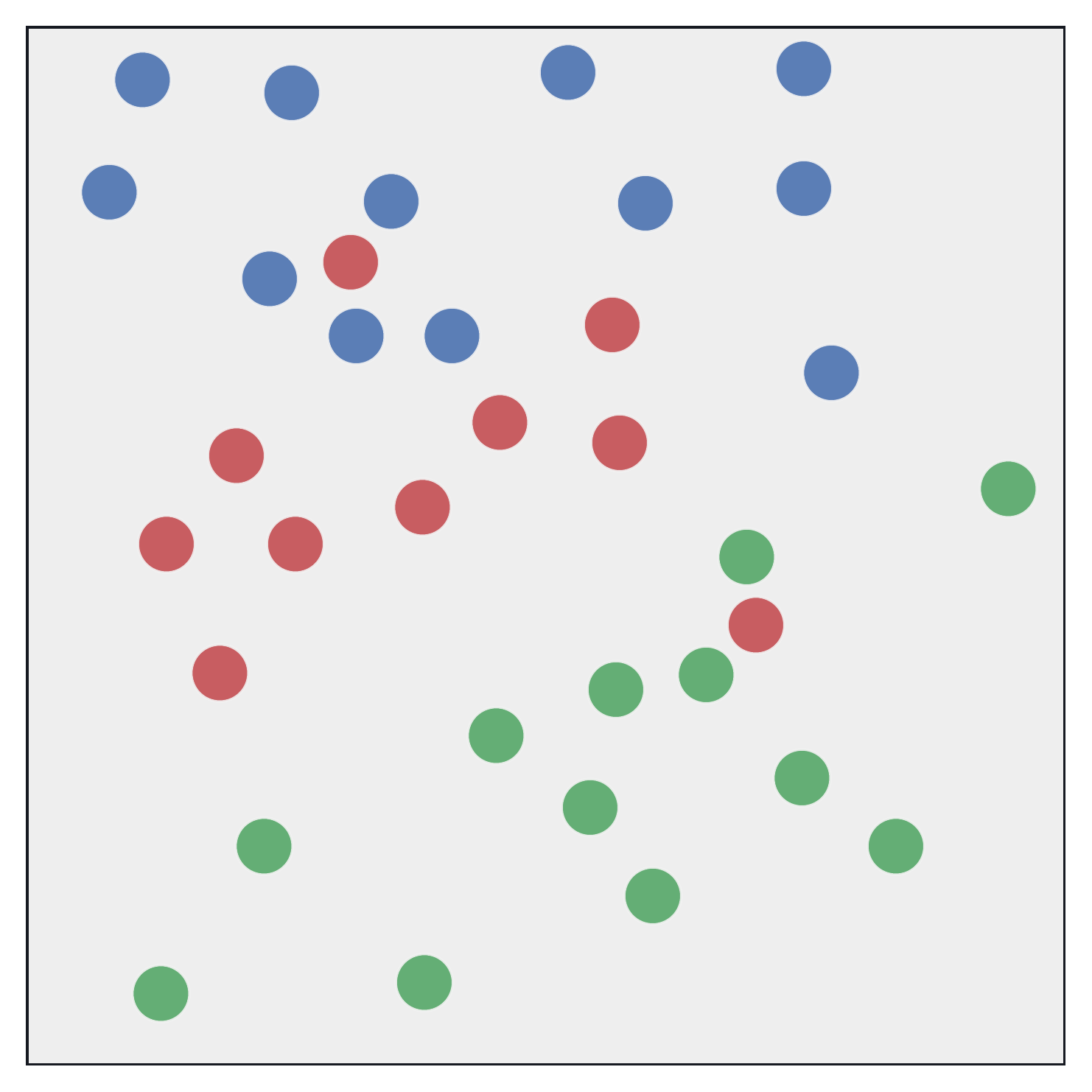}
        \includegraphics[width=0.16\textwidth]{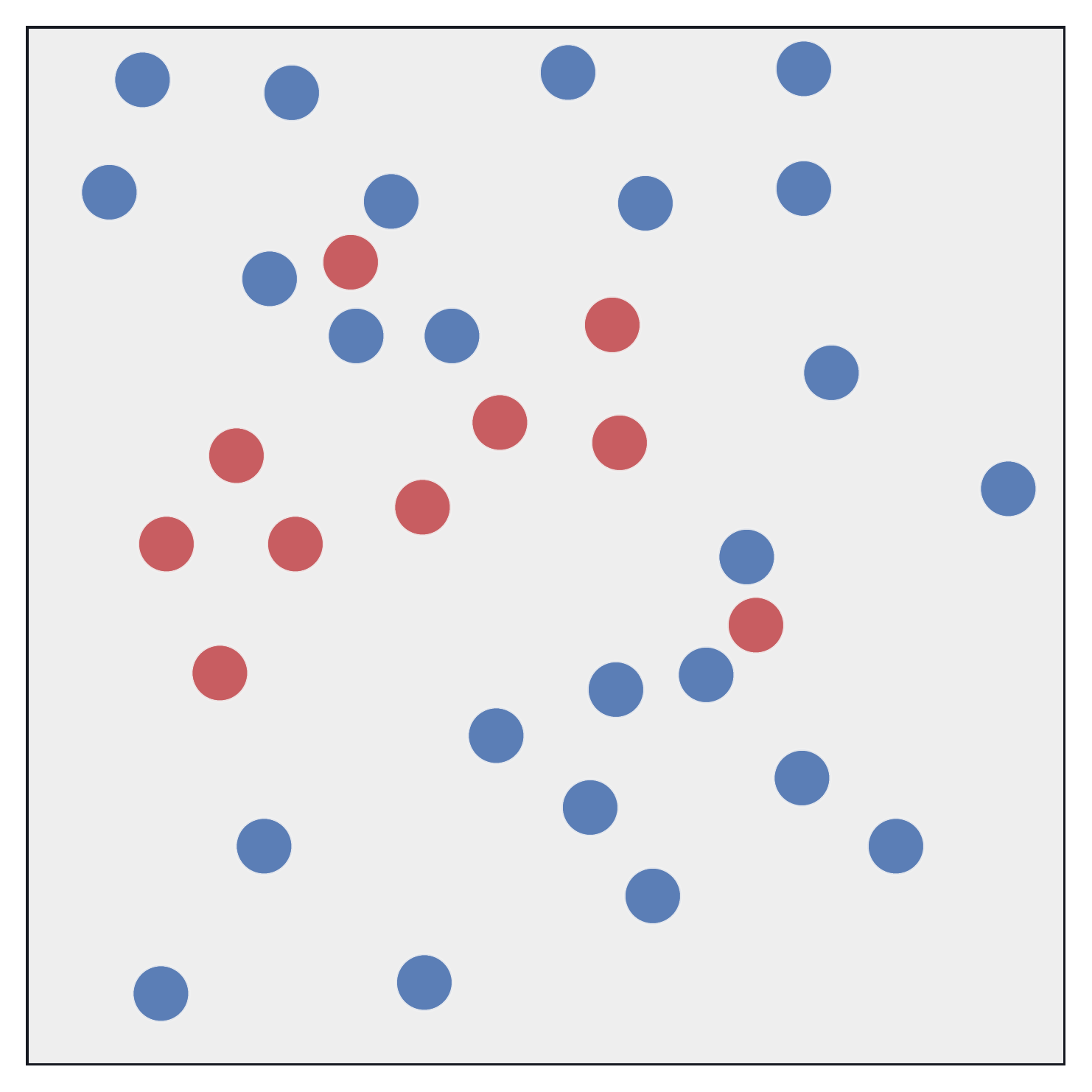}
        \includegraphics[width=0.16\textwidth]{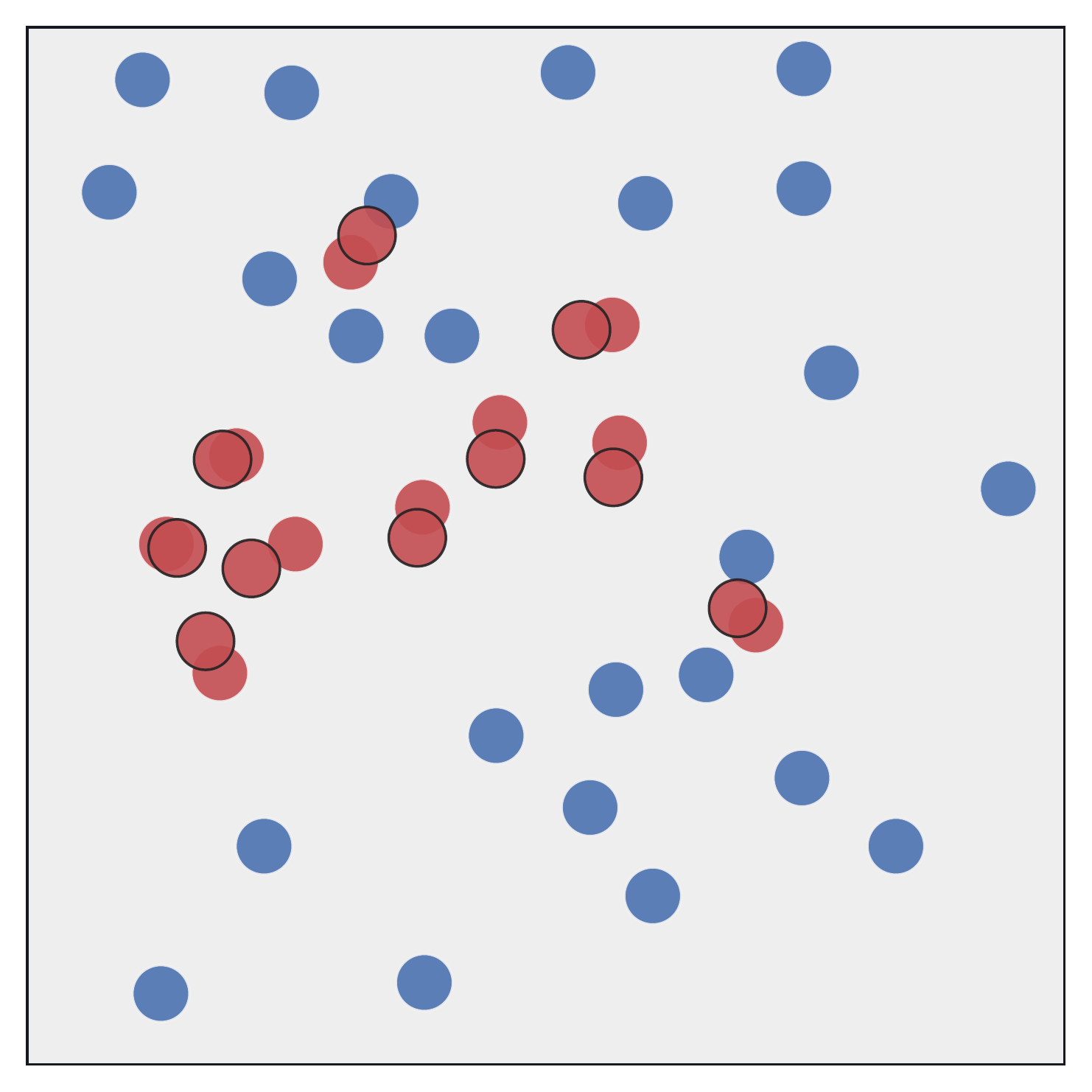}
        \includegraphics[width=0.16\textwidth]{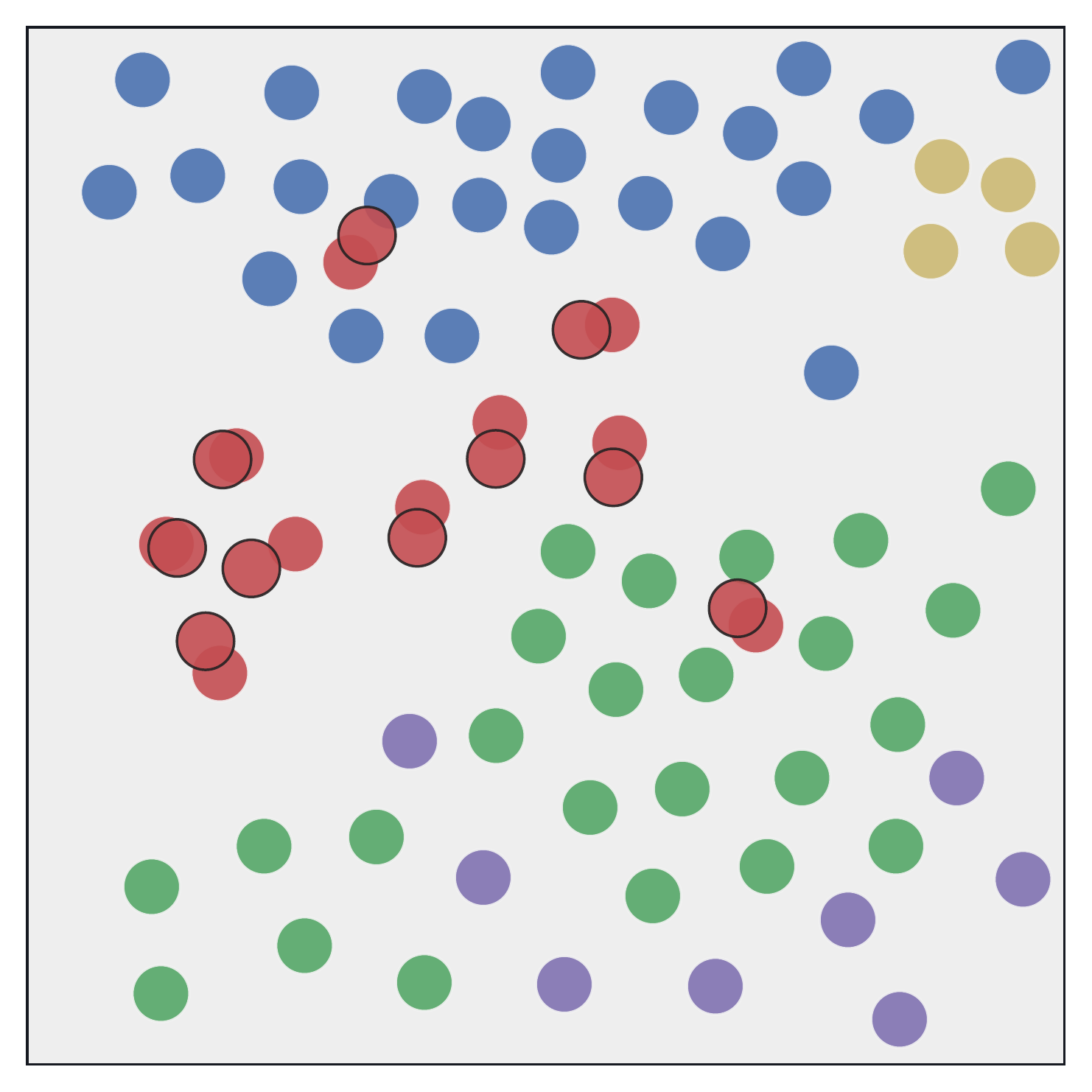}
\caption{An illustration of the multi-class problem decomposition used in MC-CCR. From the left: 1) original dataset, with one of the minority classes (in red) being currently oversampled, 2) classes with smaller number of observations are being temporarily excluded, 3) observations from the remaining majority classes are sampled in an equal proportion, 4) problem is converted to the binary setting by merging all of the majority observations into a single class, 5) cleaning and oversampling with binary CCR is applied, 6) generated synthetic observations are added to the original dataset, and the translations applied during the cleaning stage of binary CCR are preserved.}
\label{fig:mc}
\end{figure*}

To extend the CCR algorithm to a multi-class setting, we use a modified variant of decomposition strategy originally introduced by Krawczyk et al. \cite{krawczyk2019radial}. It is an iterative approach, in which individual classes are being resampled one at a time using a subset of observations from already processed classes. The approach consists of the following steps: first of all, the classes are sorted in the descending order by the number of associated observations. Secondly, for each of the minority classes, we construct a collection of combined majority observations, consisting of a randomly sampled fraction of observations from each of the already considered class. Finally, we perform a preprocessing with the CCR algorithm, using the observations from the currently considered class as a minority, and the combined majority observations as the majority class. Both the generated synthetic minority observations and the applied translations are incorporated into the original data, and the synthetic observations can be used to construct the collection of combined majority observations for later classes. We present complete pseudocode of the proposed method in Algorithm~\ref{algorithm:mc-ccr}. Furthermore, we illustrate the behavior of the algorithm in Figure~\ref{fig:mc}.

\begin{algorithm}[!htb]
	\caption{Multi-Class Combined Cleaning and Resampling}
	\label{algorithm:mc-ccr}
	\begin{algorithmic}[1]
		\STATE \textbf{Input:} collection of observations $\mathcal{X}$, with $\mathcal{X}^{(c)}$ denoting a subcollection of observations belonging to class $c$
		\STATE \textbf{Parameters:} $energy$ budget for expansion of each sphere, $p$-norm used for distance calculation
		\STATE \textbf{Output:} collections of translated and oversampled observations $\mathcal{X}$
		\STATE
		\STATE \textbf{function} MC-CCR($\mathcal{X}$, $energy$, $p$):
		\STATE $C \gets $ collection of all classes, \textbf{sorted by the number of associated observations in a descending order}
		\FOR{$i \gets 1$ \textbf{to} $|C|$}
		\STATE $n_{classes} \gets $ number of classes with higher number of observations than $C_i$
		\IF{$n_{classes} > 0$}
		\STATE $\mathcal{X}_{min} \gets \mathcal{X}^{(C_i)}$
		\STATE $\mathcal{X}_{maj} \gets \emptyset$
		\FOR{$j \gets 1$ \textbf{to} $n_{classes}$}
		\STATE add $\lfloor\frac{|\mathcal{X}^{(C_1)}|}{n_{classes}}\rfloor$ randomly chosen observations from $\mathcal{X}^{(j)}$ to $\mathcal{X}_{maj}$
		\ENDFOR
		\STATE $\mathcal{X}_{maj}'$, $\mathcal{S} \gets $ CCR($\mathcal{X}_{maj}$, $\mathcal{X}_{min}$, $energy$, $p$)
		\STATE $\mathcal{X}^{(C_i)} \gets \mathcal{X}^{(C_i)} \cup \mathcal{S}$
		\STATE substitute observations used to construct $\mathcal{X}_{maj}$ with $\mathcal{X}_{maj}'$
		\ENDIF
		\ENDFOR
		\STATE \textbf{return} $\mathcal{X}$
	\end{algorithmic}
\end{algorithm}

Compared with an alternative strategy of adapting the CCR method to the multi-class task, one-versus-all (OVA) class decomposition, the proposed algorithm has two advantages over them. Firstly, it usually decreases the computational cost, since the collection of combined majority observations was often smaller than the set of all instances in our experiments. Secondly, it assigns equal weight to every class in the collection of combined majority observations since each of them has the same number of examples.
 This would not be the case in the OVA decomposition, in which the classes with a higher number of observations could dominate the rest.

It is also important to note that the proposed approach influences the behavior of underlying resampling with CCR. First of all, because only a subset of observations is used to construct the collection of combined majority observations, the cleaning stage applies translations only on that subset of observations: in other words, the impact of the cleaning step is limited. Secondly, it affects the order of the applied translations: it prioritizes the classes with a lower number of observations, for which the translations are more certain to be preserved, whereas the translations applied during the earlier stages while resampling more populated classes can be negated. While the impact of the former on the classification performance is unclear, we would argue that at least the later is a beneficial behavior, since it further prioritizes least represented classes. Nevertheless, based on our observations, using the proposed class decomposition strategy usually also led to achieving a better performance during the classification than the ordinary OVA.

\begin{figure*}
\centering
    \begin{subfigure}[t]{0.18\textwidth}
    \captionsetup{font=scriptsize,labelfont=scriptsize}
        \centering        
        \includegraphics[width=\textwidth]{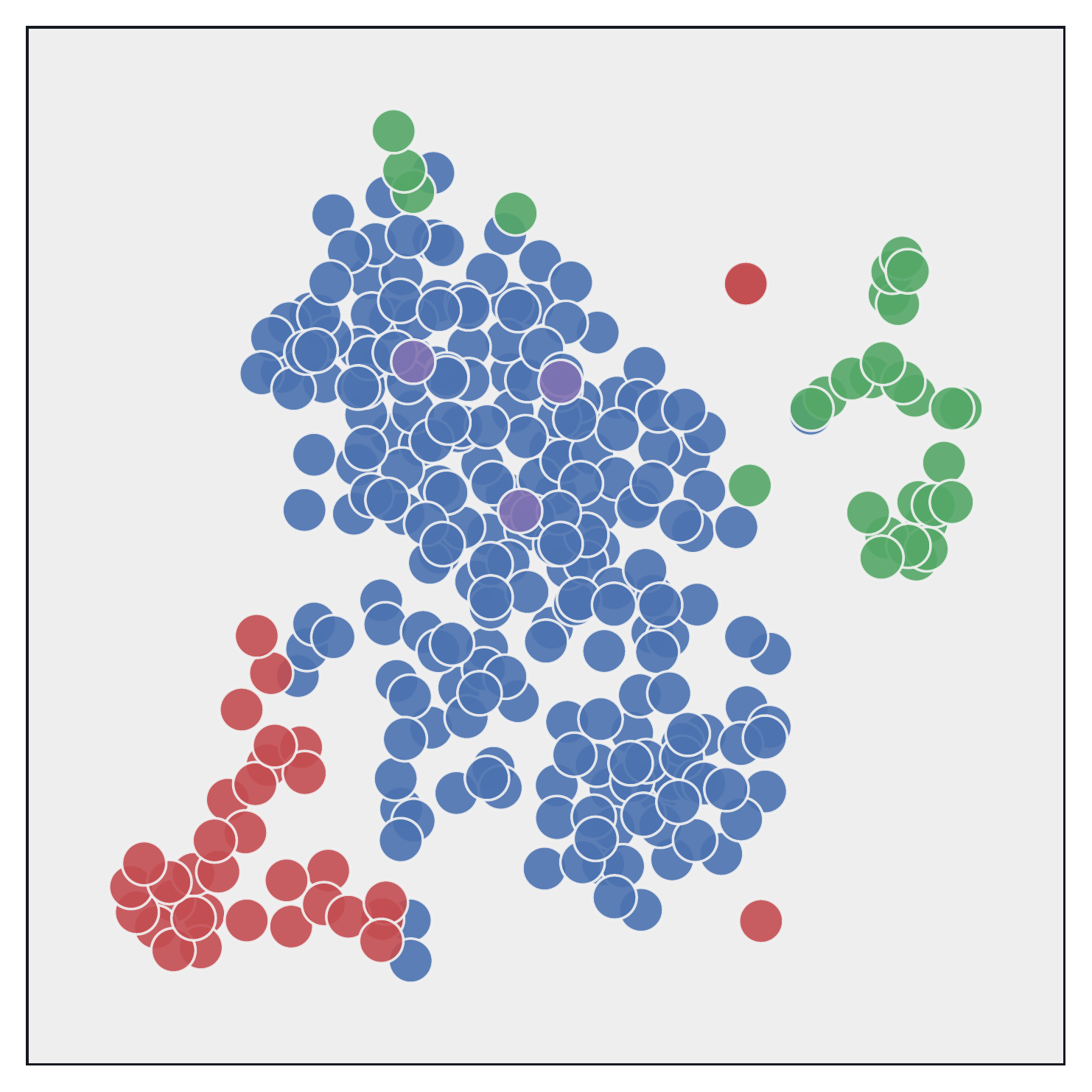}
        \caption{Original dataset}
    \end{subfigure}
    ~ 
    \begin{subfigure}[t]{0.18\textwidth}
    \captionsetup{font=scriptsize,labelfont=scriptsize}
        \centering        
        \includegraphics[width=\textwidth]{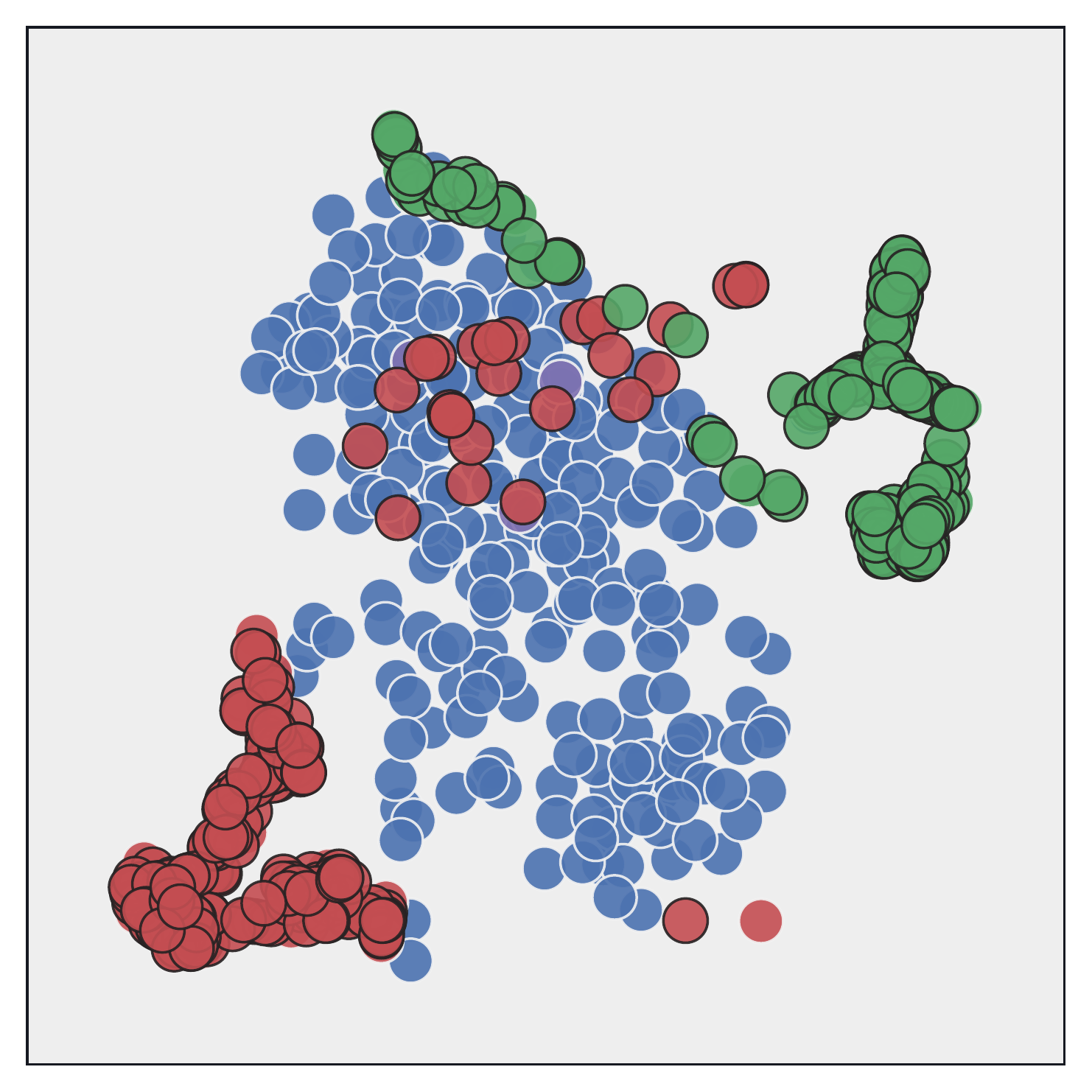}
        \caption{S-SMOTE}
    \end{subfigure}
    ~ 
    \begin{subfigure}[t]{0.18\textwidth}
    \captionsetup{font=scriptsize,labelfont=scriptsize}
        \centering        
        \includegraphics[width=\textwidth]{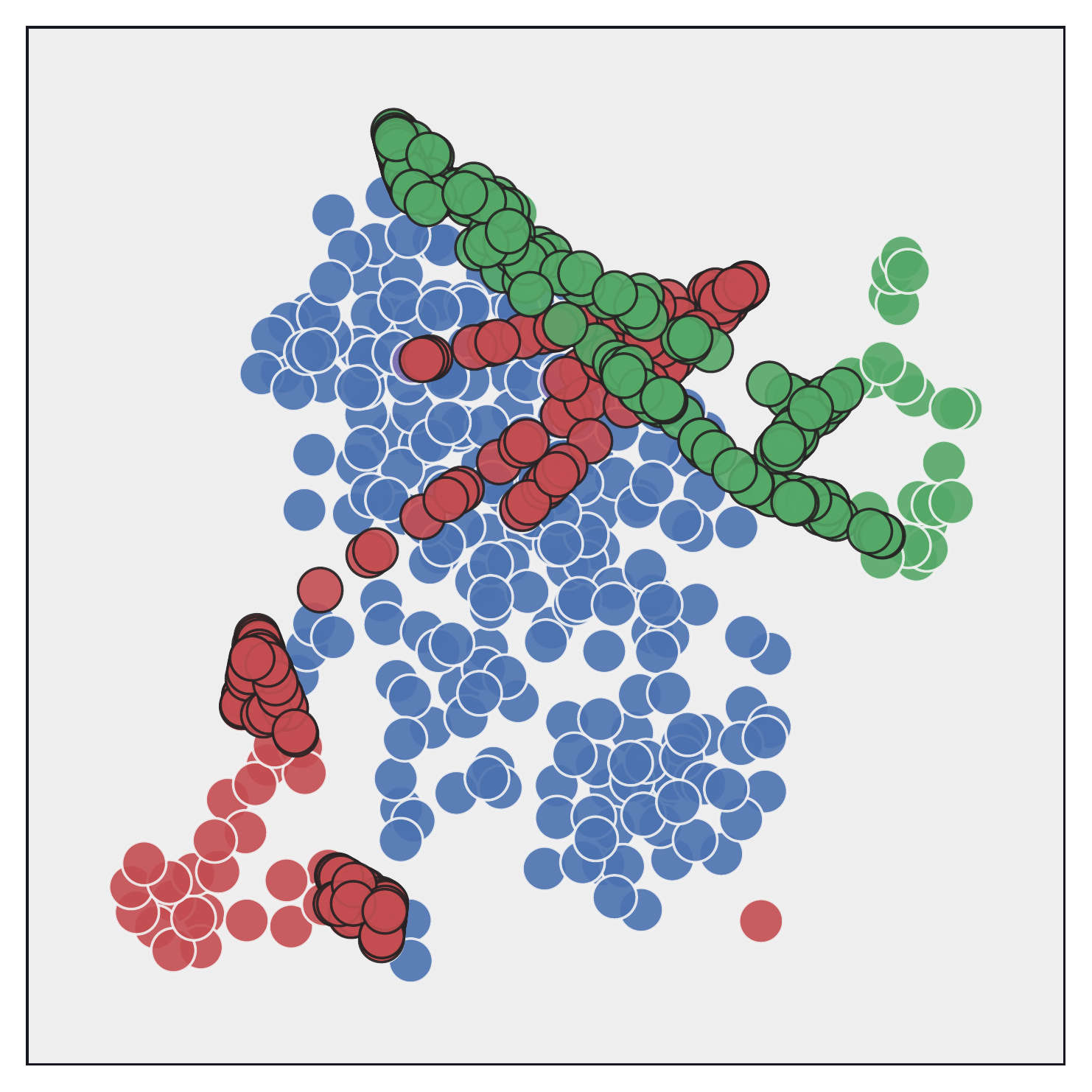}
        \caption{Borderline S-SMOTE}
    \end{subfigure}
    ~ 
    \begin{subfigure}[t]{0.18\textwidth}
    \captionsetup{font=scriptsize,labelfont=scriptsize}
        \centering        
        \includegraphics[width=\textwidth]{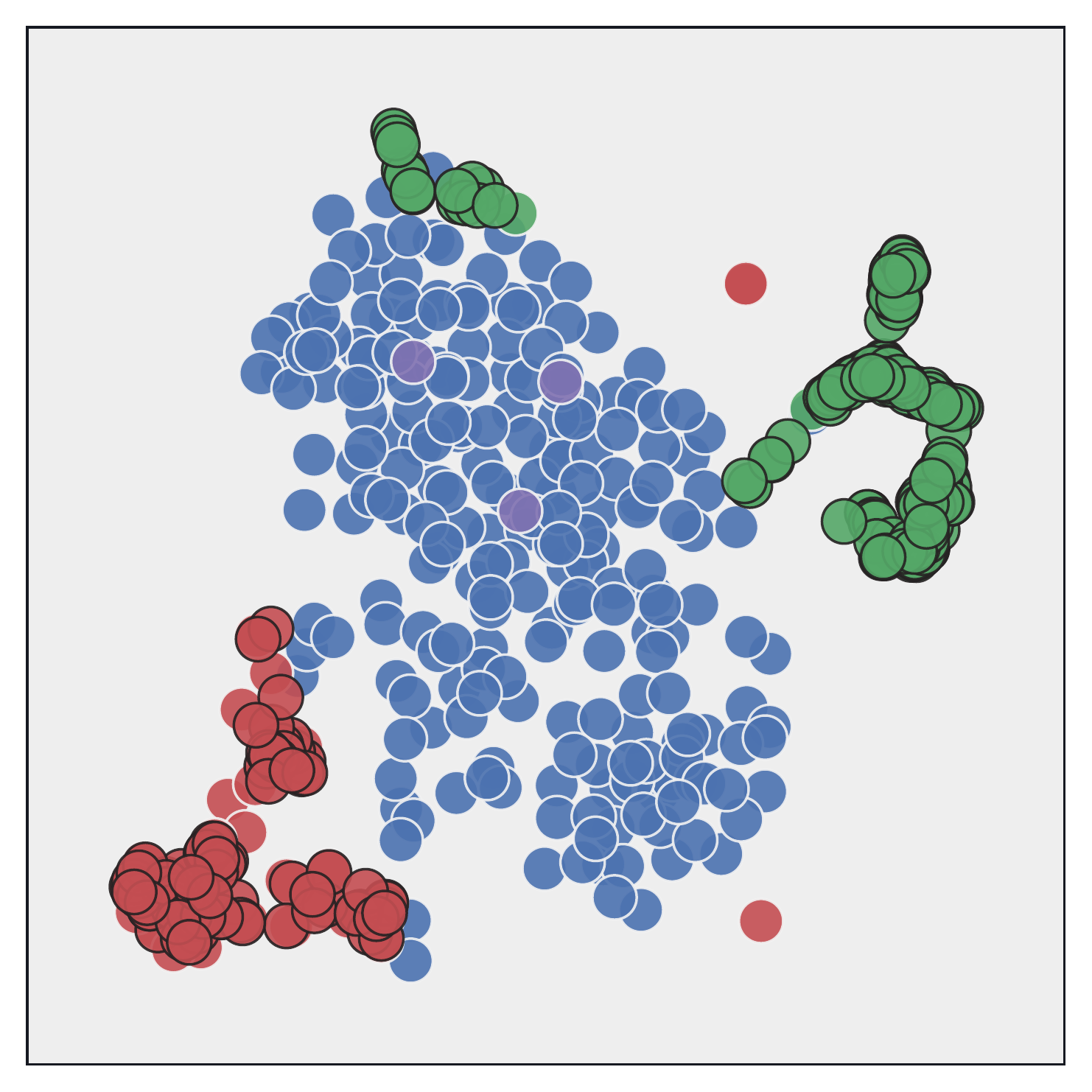}
        \caption{S-SMOTE + ENN}
    \end{subfigure}
    ~ 
    \begin{subfigure}[t]{0.18\textwidth}
    \captionsetup{font=scriptsize,labelfont=scriptsize}
        \centering        
        \includegraphics[width=\textwidth]{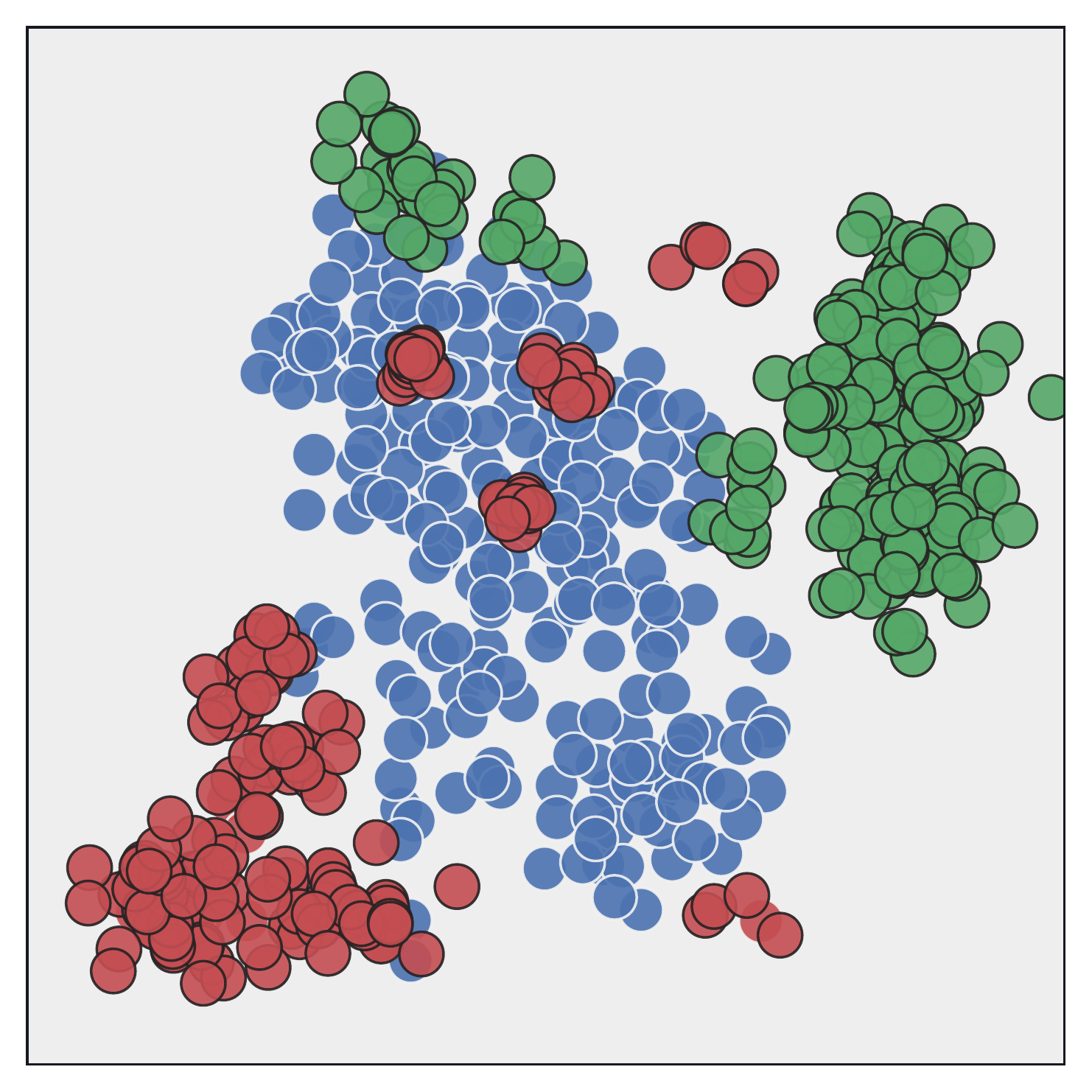}
        \caption{MC-CCR}
    \end{subfigure}
\caption{A comparison of different oversampling algorithms on a multi-class dataset with one majority class (in blue), two minority classes (in green and red), and several majority observations mislabeled as minority (in purple). Both S-SMOTE and Borderline S-SMOTE are susceptible to the presence of label noise and disjoint data distributions. Mechanisms of dealing with outliers, such as postprocessing with ENN, mitigate some of these issues, but at the same time completely exclude underrepresented regions. MC-CCR reduces the negative impact of mislabeled observations by constraining the oversampling regions around them, and at the same time does not ignore outliers not surrounded by majority observations.}
\label{fig:comp}
\end{figure*}

We present a comparison of the proposed MC-CCR algorithm with several SMOTE-based approaches in Figure~\ref{fig:comp}. We use an example of a multi-class dataset with two minority classes, disjoint data distributions and label noise. As can be seen, S-SMOTE is susceptible to the presence of label noise and disjoint data distributions, producing synthetic minority observations overlapping the majority class distribution. Borderline S-SMOTE, while less sensitive to the presence of individual mislabeled observations, remains even more affected by the disjoint data distributions. Mechanisms of dealing with outliers, such as postprocessing with ENN, mitigate both of these issues, but at the same time exclude entirely underrepresented regions, likely to occur in the case of high data imbalance or small total number of observations. MC-CCR reduces the negative impact of mislabeled observations by constraining the oversampling regions around them, and at the same time, does not ignore outliers not surrounded by majority observations.

\subsection{Computational complexity analysis}

Let us define the total number of observations by $n$, the number of majority and minority observations in the binary case by, respectively, $n_{maj}$ and $n_{min}$, the number of features by $m$, and the number of classes in the multi-class setting by $c$. Let us first consider the worst-case complexity of the binary variant of CCR. The algorithm can be divided into three steps: calculating the sphere radii, cleaning the majority observations inside the spheres, and synthesizing new observations. Each one of these steps is applied iteratively to every minority observation. The first step consists of a) calculating a distance vector, which requires $n_{maj}$ distance calculations, each with the complexity equal to $\mathcal{O}(m)$, and a combined complexity equal to $\mathcal{O}(mn_{maj})$, b) sorting said $n_{maj}$-dimensional vector, an operation that has a complexity equal to $\mathcal{O}(n_{maj}\log{n_{maj}})$, and c) calculating the resulting radius, which in the worst-case scenario will never reach the break clause, and will require $n_{maj}$ iterations, each one with scalar operations only, leading to a complexity of $\mathcal{O}(n_{maj})$. Combined, these operations have a complexity equal to $\mathcal{O}(mn_{maj} + n_{maj}\log{n_{maj}} + n_{maj})$ per minority observation, or in other words $\mathcal{O}((mn_{maj} + n_{maj}\log{n_{maj}} + n_{maj})n_{min})$, which can be simplified to $\mathcal{O}((m + \log{n})n^2)$. The second step, cleaning the majority observations inside the spheres, in the worst case, requires $n_{maj}$ operations of calculating and applying the translation vector per minority observation, each with a complexity equal to $\mathcal{O}(m)$, leading to a combined complexity of $\mathcal{O}(mn_{min})$, which can be simplified to $\mathcal{O}(mn)$. The third step, synthesizing new observations, requires $n_{min}$ summations for calculating the the denominator of Equation~\ref{eq:prop}, which has a complexity of $\mathcal{O}(n_{min})$, $n_{min}$ operations of calculating the proportion of generated objects for a given observation, each with a complexity equal to $\mathcal{O}(1)$ (when using the precalculated denominator), and $n_{maj} - n_{min}$ operations of sampling a random observation inside the sphere, each with a complexity equal to $\mathcal{O}(m)$. Combined, the complexity of the step is equal to $\mathcal{O}(n_{min} + n_{min} + m(n_{maj} - n_{min}))$, which can be simplified to $\mathcal{O}(mn)$. As can be seen, the complexity of the algorithm is dominated by the first step and is equal to $\mathcal{O}((m + \log{n})n^2)$. It is also worth noting that in the case of an extreme imbalance, that is when the $n_{min}$ is equal to 1, the complexity of the algorithm is equal to $\mathcal{O}((m + \log{n})n)$, which is the best case. Finally, since the complexity of the binary variant of CCR is not reliant on the number of observations to be generated, and the main computational cost of MC-CCR is associated with $c - 1$ calls to CCR, the worst-case complexity of the MC-CCR algorithm is equal to $\mathcal{O}(c(m + \log{n})n^2)$.

\section{Experimental Study}
\label{sec:exp}
In this section, we will describe the details of a conducted experimental study that can assess the usefulness of MC-CCR. The research questions for this study are: 
\begin{itemize}
\item[RQ1:] What is the best parameter setting for MC-CCR, and how they impact the behavior of the algorithm?
\item[RQ2:] How robust is the MC-CCR to label noise in learning data?
\item[RQ3:] What is the predictive performance of the MC-CCR in comparison to the state-of-art oversampling methods?
\item[RQ4:] How flexible is MC-CCR to be used with the different classifiers?
\end{itemize}

\subsection{Set-up}
\label{sec:set}

\textbf{Datasets.} We based our experiments on 20 multi-class imbalanced datasets from KEEL repository \cite{alcala2011keel}. Their details were presented in Table~\ref{tab:data}. The selection of the datasets was made based on the previous work by S{\'{a}}ez et al. \cite{Saez:2016}, in which it was demonstrated that the chosen datasets possess various challenging characteristics, such as small disjuncts, frequent borderline and noisy instances, and class overlapping.

\begin{table*}[tbp]
\caption{Details of the multi-class imbalanced benchmarks used in the experiments.}
\small
\label{tab:data}
\centering
\begin{tabular}{ l l l l l l}
\toprule
\tabhead{Dataset} & \tabhead{\#Instances} & \tabhead{\#Features} & \tabhead{\#Classes} & \tabhead{IR} & \tabhead{Class distribution} \\ \midrule
Automobile&150&25&6&16.00&3/20/48/46/29/13\\
Balance&625&4&3&5.88&288/49/288\\
Car& 1728 & 6 & 4 & 18.61&65/69/384/1210\\
Cleveland&297&13&5&12.62&164/55/36/35/13\\
Contraceptive&1473&9&3&1.89&629/333/511\\
Dermatology&358&33&6&5.55&111/60/71/48/48/20\\
Ecoli&336&7&8&71.50&143/77/2/2/35/20/5/52\\
Flare&1066&11&6&7.70&331/239/211/147/95/43\\
Glass&214&9&6&8.44&70/76/17/13/9/29\\
Hayes-Roth&160&4&3&2.10&160/65/64/31\\
Led7digit&500&7&10&1.54&45/37/51/57/52/52/47/57/53/49\\
Lymphography&148&18&4&40.50&2/81/61/4\\
New-thyroid&215&5&3&5.00&150/35/30\\
Page-blocks&5472&10&5&175.46&4913/329/28/87/115\\
Thyroid&7200&21&3&40.16&166/368/6666\\
Vehicle& 846 & 18 & 4 & 1.17&199/212/217/218\\
Wine&178&13&3&1.48&59/71/48\\
Winequality-red&1599&11&6&68.10&10/53/681/638/199/18\\
Yeast&1484&8&10&92.60&244/429/463/44/51/163/35/30/20/5\\
Zoo&101&16&7&10.25&41/13/10/20/8/5/4\\
\bottomrule
\end{tabular}
\end{table*}

\noindent\textbf{Reference methods.} Throughout the conducted experiments the proposed method was compared with a selection of state-of-the-art multi-class data oversampling algorithms. Specifically, for comparison we used SMOTE algorithm using round-robin decomposition strategy (SMOTE-all), STATIC-SMOTE (S-SMOTE) \cite{Fernandez-Navarro:2011}, Mahalanobis Distance Oversampling (MBO) \cite{Abdi:2016}, ($k$-NN)-based synthetic minority oversampling algorithm (SMOM) \cite{Zhu:2017}, and SMOTE combined with an Iterative-Partitioning Filter (SMOTE-IPF) \cite{saez2015smote}. Parameters of the reference methods used throughout the experimental study were presented in Table~\ref{tab:par}.

\begin{table}
\begin{center}
\caption{Parameters of the classification and the sampling algorithms used throughout the experimental study.}
\label{tab:par}
\scalebox{0.8}{
\begin{tabular}{ll}
\toprule
\tabhead{Algorithm} & \tabhead{Parameters}\\
\midrule
MLP&training: rprop;\\ 
&iterations $\in [100,200,\cdots,1000]$; \\
&\#hidden neurons = $\frac{\# \text{input} + \# \text{output}}{2}$\\
$k$-NN& nearest neighbors $\in [1,3,\cdots,11]$\\
\midrule
MC-CCR & energy $\in \{0.001, 0.0025, 0.005, 0.01, ..., 100.0\}$; \\
& cleaning strategy: translation; \\
& selection strategy: proportional; \\
& multi-class decomposition method: sampling;\\
& oversampling ratio $\in [50,100,\cdots,500]$\\
SMOTE-all &$k$-nearest neighbors = 5;\\
& oversampling ratio $\in [50,100,\cdots,500]$\\
S-SMOTE \cite{Fernandez-Navarro:2011}&$k$-nearest neighbors = 5;\\
& oversampling ratio $\in [50,100,\cdots,500]$\\
MDO \cite{Abdi:2016} & $K$1  $\in [1,2,\cdots,10]$;\\
&  $K$2 $\in [2,4,\cdots,20]$;\\
& oversampling ratio $\in [50,100,\cdots,500]$\\
SMOM \cite{Zhu:2017}& $K$1  $\in [2,4,\cdots,20]$;\\
&  $K$2 $\in [1,2,\cdots,10]$;\\
& $rTh \in [0.1,0.2,\cdots,1]$;\\
& $rTh \in [1,2,\cdots,10]$;\\
& $w1,w2,r1,r2 \in [0.1,0.2,\cdots,1]$;\\
& $k$-nearest neighbors = 5;\\
& oversampling ratio $\in [50,100,\cdots,500]$\\
SMOTE-IPF \cite{saez2015smote} & $n = 9$; \\
& $n$-nearest neighbors = 5;\\
& $n$ partitions = 9;\\
& $k$ iterations = 3;\\
& $p = 0.01$;\\
& oversampling ratio $\in [50,100,\cdots,500]$\\
\bottomrule
\end{tabular}
}
\end{center}
\end{table}

\noindent\textbf{Classification algorithms.} To ensure the validity of the observed results across different learning methodologies we evaluated the considered oversampling algorithms in combination with four different classification algorithms: decision trees (C5.0 model), neural networks (multi-layer perceptron, MLP), lazy learners ($k$-nearest neighbors, $k$-NN), and probabilistic classifiers (Na\"ive Bayes, NB). The parameters of the classification algorithms used throughout the experimental study were presented in Table~\ref{tab:par}.

\noindent\textbf{Evaluation procedure.} The evaluation of the considered algorithms was conducted using a 10-fold cross validation, with the final performance averaged over 10 experimental runs. Parameter selection was conducted independently for each data partition using 3-fold cross validation on the training data.

\noindent\textbf{Statistical analysis.} To assess the statistical significance of the observed results we used a combined 10-fold cross-validation F-test \cite{Japkowicz:2011} during all of the conducted pairwise comparisons, whereas for the comparisons including multiple methods we used a Friedman ranking test with Shaffer post-hoc analysis \cite{Garcia:2010}. The results of all of the performed tests were reported at a significance level $\alpha = 0.05$.

\noindent\textbf{Reproducibility.} The proposed MC-CCR algorithm was implemented in Python programming language and published as an open-source code at\footnote{https://github.com/michalkoziarski/MC-CCR}.

\subsection{Examination of a validity of the design choices behind MC-CCR}
\label{sec:exp1}

The aim of the first stage of the conducted experimental study was to establish the validity of the design choices behind the MC-CCR algorithm. While intuitively motivated, the individual components of MC-CCR are heuristic in nature, and it is not clear whether they actually lead to a better results. Specifically, three variable parts of MC-CCR that can affect its performance can be distinguished. First of all, the cleaning strategy, or in other words the way MC-CCR handles the majority instances located inside the generated spheres. While the proposed algorithm handles these instances by moving them outside the sphere radius (translation, T), at least two additional approaches can be reasonably argued for: complete removal of the instances located inside the spheres (removal, R), or not conducting any cleaning and ignoring the position of the majority observations with respect to the spheres (ignoring, I). Secondly, the selection strategy, or the approach of assigning greater probability of generating new instances around the minority observations with small associated sphere radius. In the proposed MC-CCR algorithm we use a strategy in which that probability is inversely related to the sphere radius (proportional, P), which corresponds to focusing oversampling on the difficult regions, nearby the borderline and outlier instances. For comparison, we also used a strategy in which the seed instances around which synthetic observations are to be generated are chosen randomly, with no associated weight (random, R). Finally, in the multi-class decomposition we compared two methods of combining several classes into a one combined majority class. First of all, the approach proposed in this paper, that is sampling only the classes with a greater number of observations in an even proportion to generate a combined majority class (sampling, S). Secondly, for a comparison we considered a case in which all of the observations from all of the remaining classes are combined (complete, C).

To experimentally validate the decided upon designed choices we conducted an experiment in which each we compared all of the possible combinations of the outlined parameters. We present the results, averaged across all of the considered datasets, in Table~\ref{tab:param-imp}. As can be seen, the combination of parameters proposed in the form of MC-CCR, that is the combination of cleaning by translation, proportional seed observations selection, and using sampling during the multi-class decomposition, leads to an, on average, best performance for all of the baseline classifiers and performance metrics. In particular, the choice of the cleaning strategy proves to be vital to achieve a satisfactory performance, and conducting no cleaning at all produces significantly worse results.

\begin{table*}[tbp]
\centering
\caption{Impact of MC-CCR parameters on four classification measures and four base classifiers.}
\label{tab:param-imp}
\scalebox{0.65}{
\begin{tabular}{ccccccccccccccccccc}
\toprule
\multicolumn{3}{c}{MC-CCR parameters} & \multicolumn{4}{c}{C5.0} & \multicolumn{4}{c}{MLP} & \multicolumn{4}{c}{k-NN} & \multicolumn{4}{c}{NB}\\
\tabhead{Cleaning}         & \tabhead{Selection}    & \tabhead{Method}  & \tabhead{AvAcc}   & \tabhead{CBA}    & \tabhead{mGM} & \tabhead{CEN}& \tabhead{AvAcc}   & \tabhead{CBA}    & \tabhead{mGM} & \tabhead{CEN}& \tabhead{AvAcc}   & \tabhead{CBA}    & \tabhead{mGM} & \tabhead{CEN}& \tabhead{AvAcc}   & \tabhead{CBA}    & \tabhead{mGM} & \tabhead{CEN}\\
\midrule
T & P & S & \textbf{73.74} & \textbf{75.22} & \textbf{72.22} & \textbf{0.28}& \textbf{71.88} & \textbf{74.82} & \textbf{71.64} & \textbf{0.25}& \textbf{73.01} & \textbf{74.76} & \textbf{71.90} & \textbf{0.27} & \textbf{74.28} & \textbf{75.69} & \textbf{73.64} & \textbf{0.29} \\
T&P&C&72.19&74.76&71.62&0.30&70.81&74.12&70.93&0.27&72.11&73.58&70.19&0.29&72.99&74.18&72.06& 0.31 \\
T&R&S&67.29&70.19&65.19&0.36&65.18&67.02&64.81&0.38&66.01&68.29&65.93&0.37&68.02&70.93&65.99& 0.35 \\
T&R&C&66.02&68.53&64.08&0.38&64.59&66.39&62.88&0.39&65.47&68.11&63.28&0.38&66.38&68.92&64.71&0.37 \\
R&P&S&70.08&70.82&68.15&0.32&66.53&68.49&65.92&0.31&68.91&69.19&68.06&0.32&71.09&71.72&69.69& 0.33 \\
R&P&C&68.69&70.01&67.40&0.34&64.89&65.47&63.94&0.38&65.11&67.09&64.38&0.34&69.02&70.99&68.25& 0.35 \\
R&R&S&65.89&66.52&64.98&0.38&63.19&64.82&62.89&0.40&63.78&65.59&63.71&0.35&68.09&69.17&66.84& 0.36 \\
R&R&C&63.88&64.52&63.28&0.41&61.38&62.70&60.97&0.43&61.99&63.55&61.22&0.42&63.94&64.82&63.77&0.40 \\
I&P&S&61.03&62.35&60.61&0.45&59.62&60.02&59.19&0.48&59.97&60.18&59.70&0.47&61.11&62.49&60.89& 0.44 \\
I&P&C&59.78&60.96&59.33&0.48&56.29&57.56&56.03&0.51&57.17&59.24&56.98&0.49&59.83&61.06&59.82& 0.47 \\
I&R&S&56.95&58.49&56.38&0.53&52.87&54.19&52.46&0.56&53.48&55.07&53.11&0.52&57.01&59.61&56.59& 0.52 \\
I&R&C&55.88&58.02&55.09&0.54&52.10&53.28&51.89&0.58&52.71&54.62&52.28&0.55&56.03&58.72&55.81& 0.53 \\
\bottomrule
\end{tabular}
}
\end{table*}

\subsection{Comparison with the reference methods}
\label{sec:exp2}

\begin{figure*}
\centering
    \begin{subfigure}[t]{0.99\textwidth}
    \captionsetup{font=scriptsize,labelfont=scriptsize}
        \centering        
        \includegraphics[width=0.24\textwidth]{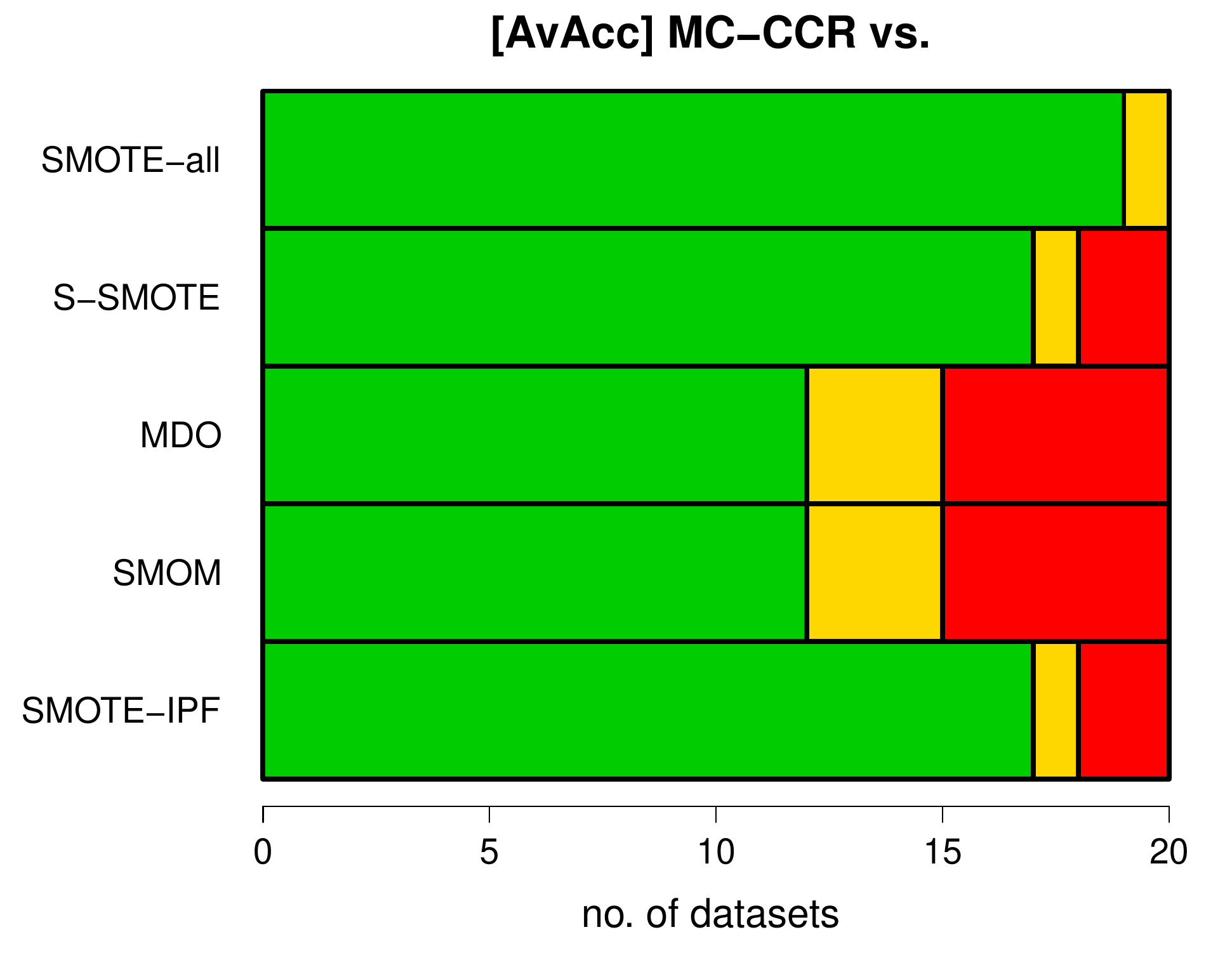}
        \includegraphics[width=0.24\textwidth]{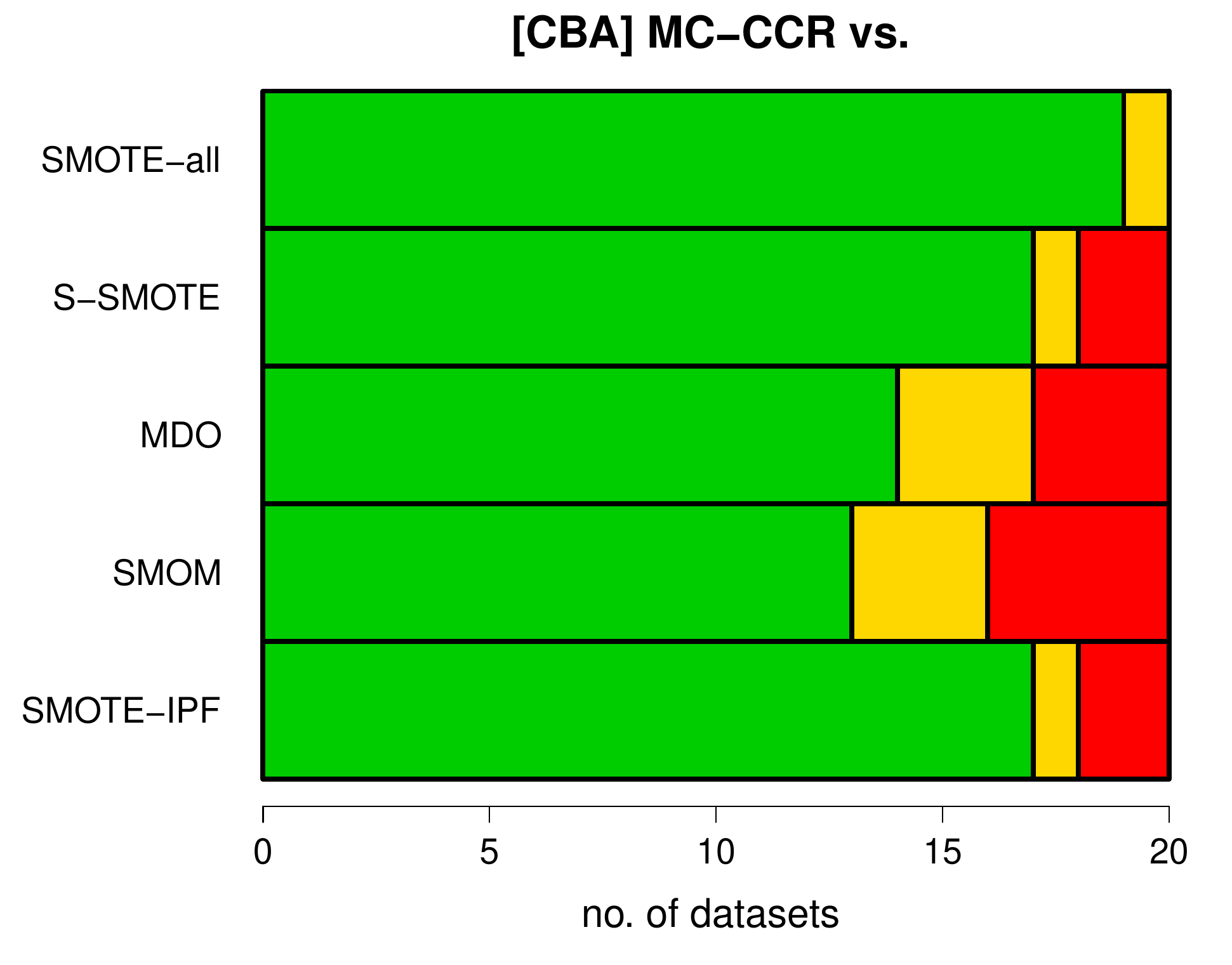}    
        \includegraphics[width=0.24\textwidth]{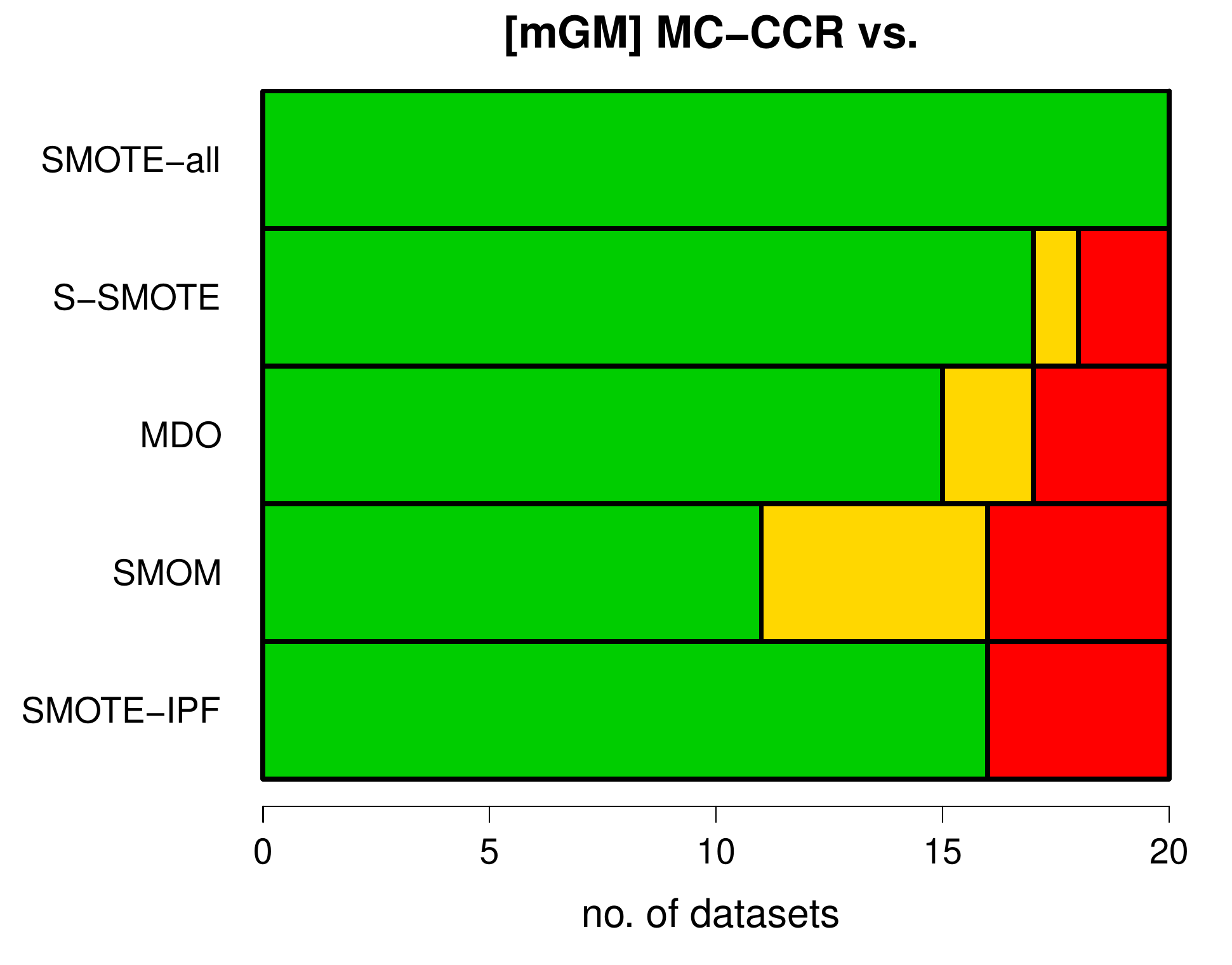}
        \includegraphics[width=0.24\textwidth]{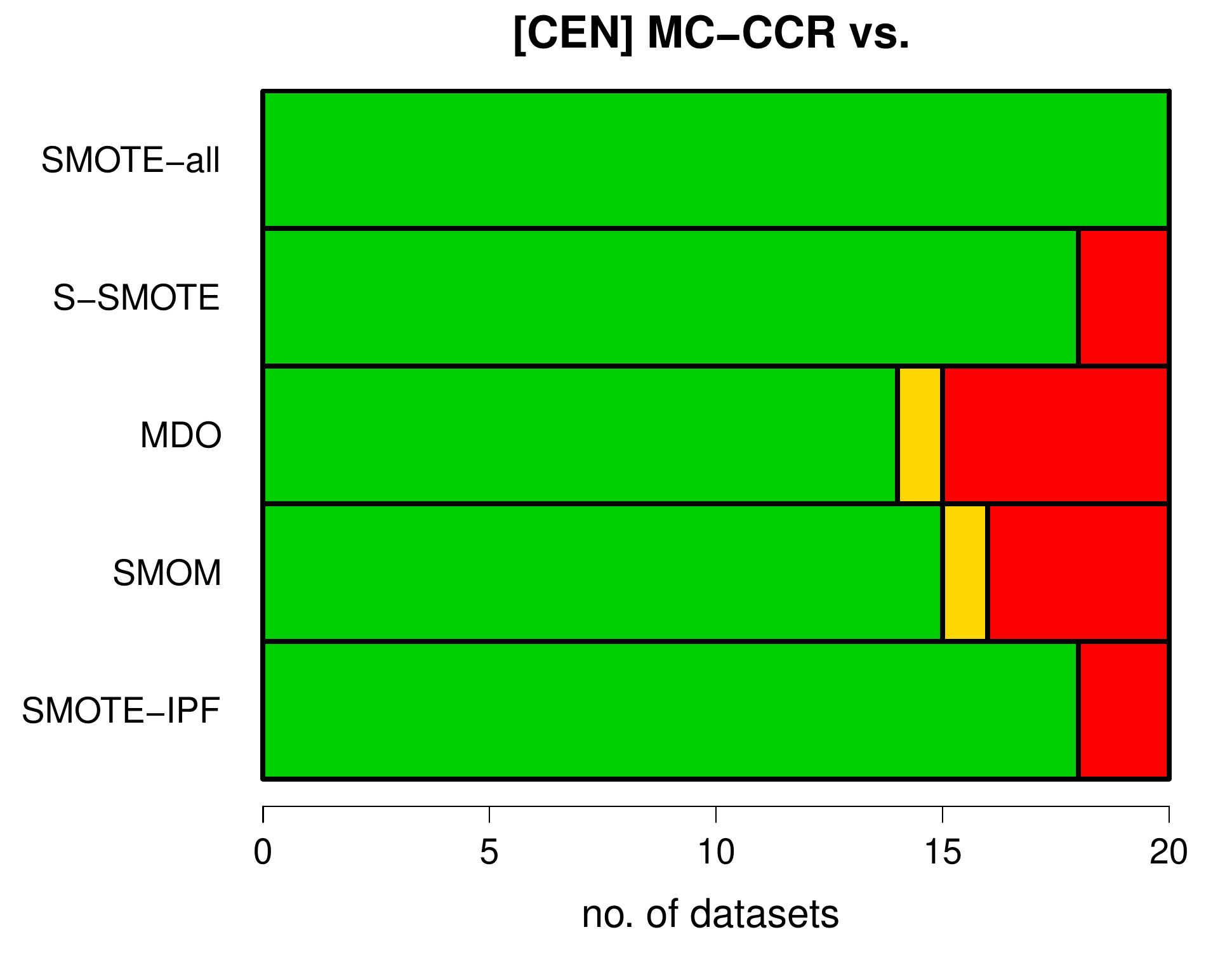}
        \caption{C5.0}
    \end{subfigure}
    ~ 
    \begin{subfigure}[t]{0.99\textwidth}
    \captionsetup{font=scriptsize,labelfont=scriptsize}
        \centering        
        \includegraphics[width=0.24\textwidth]{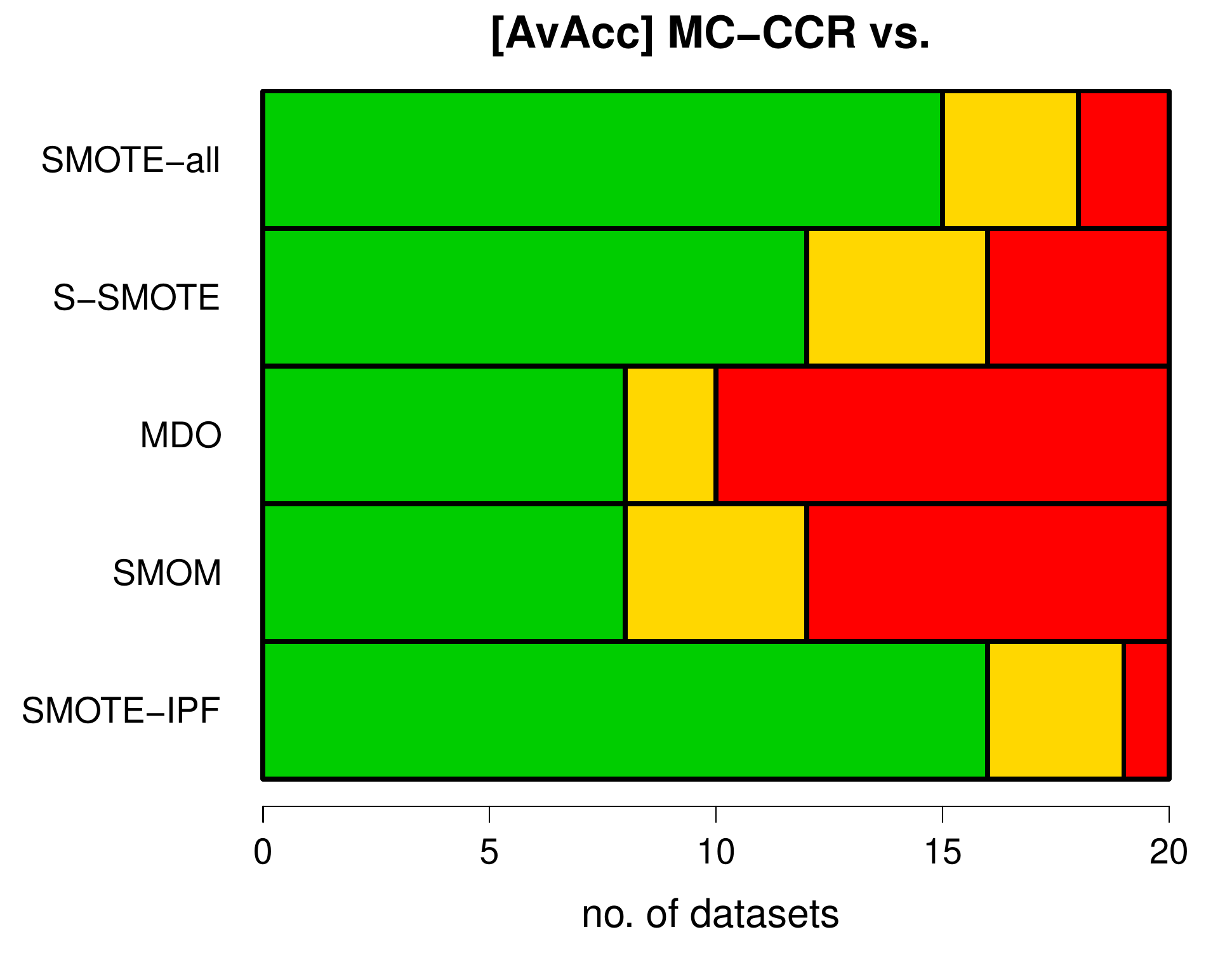}
        \includegraphics[width=0.24\textwidth]{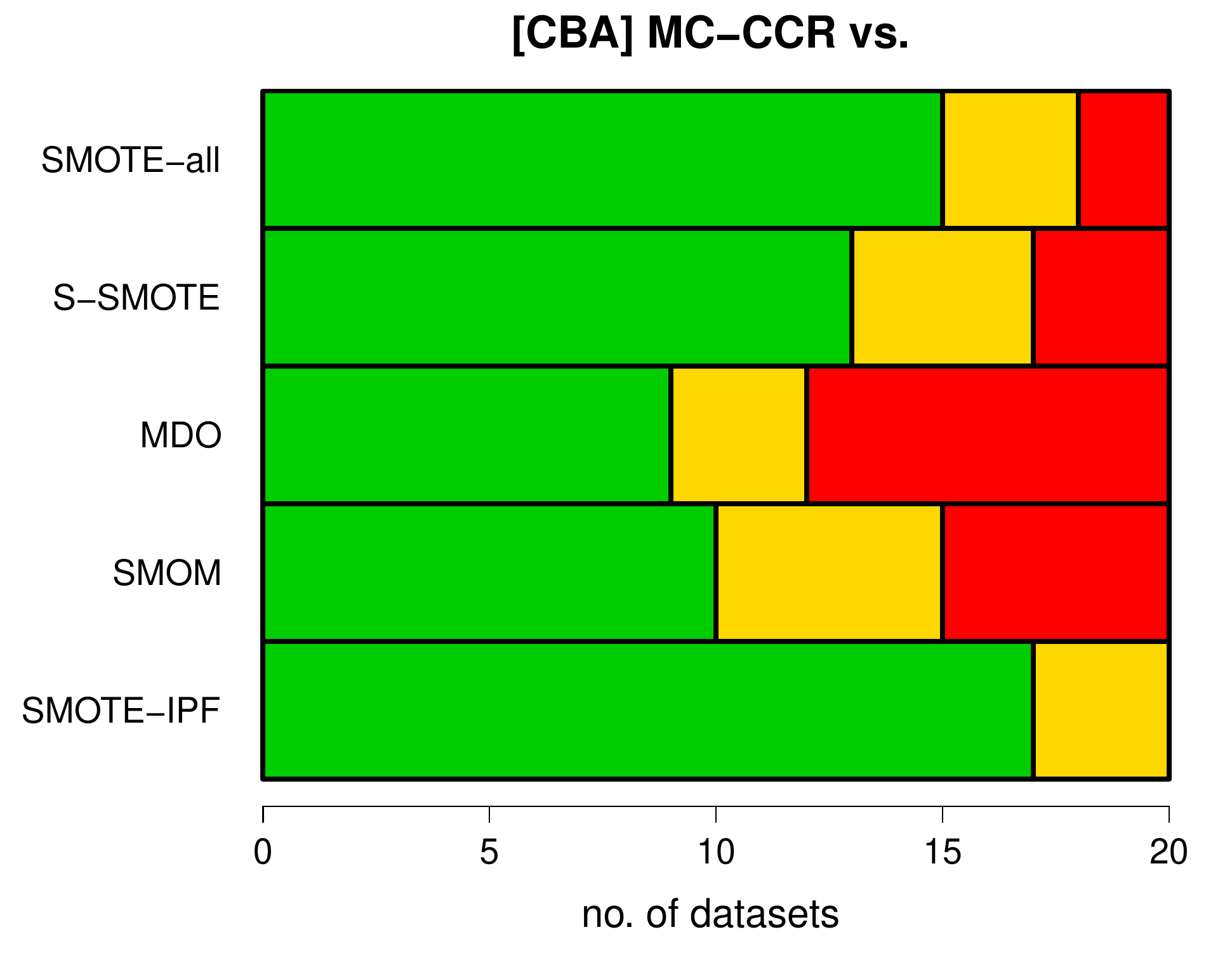}    
        \includegraphics[width=0.24\textwidth]{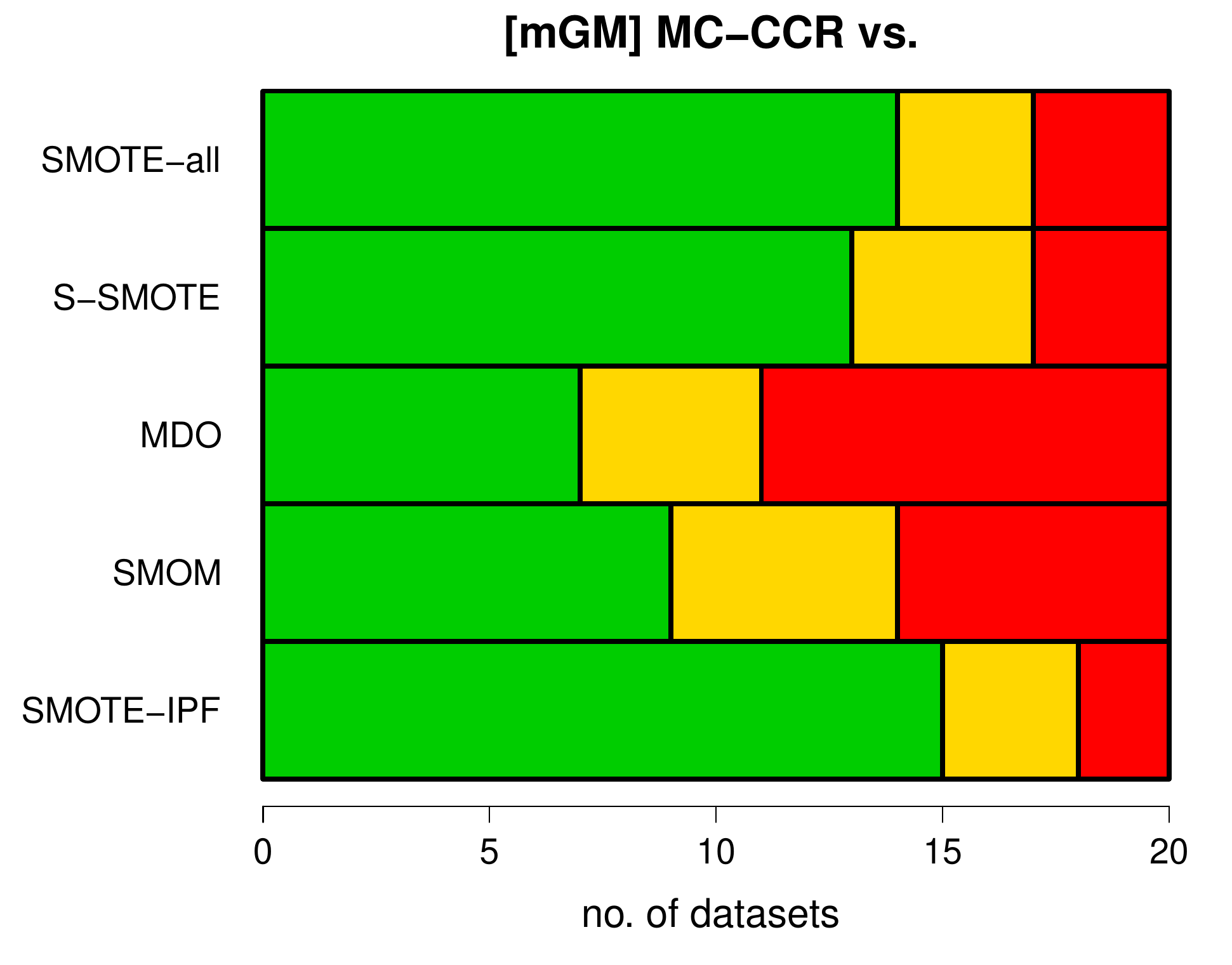}
        \includegraphics[width=0.24\textwidth]{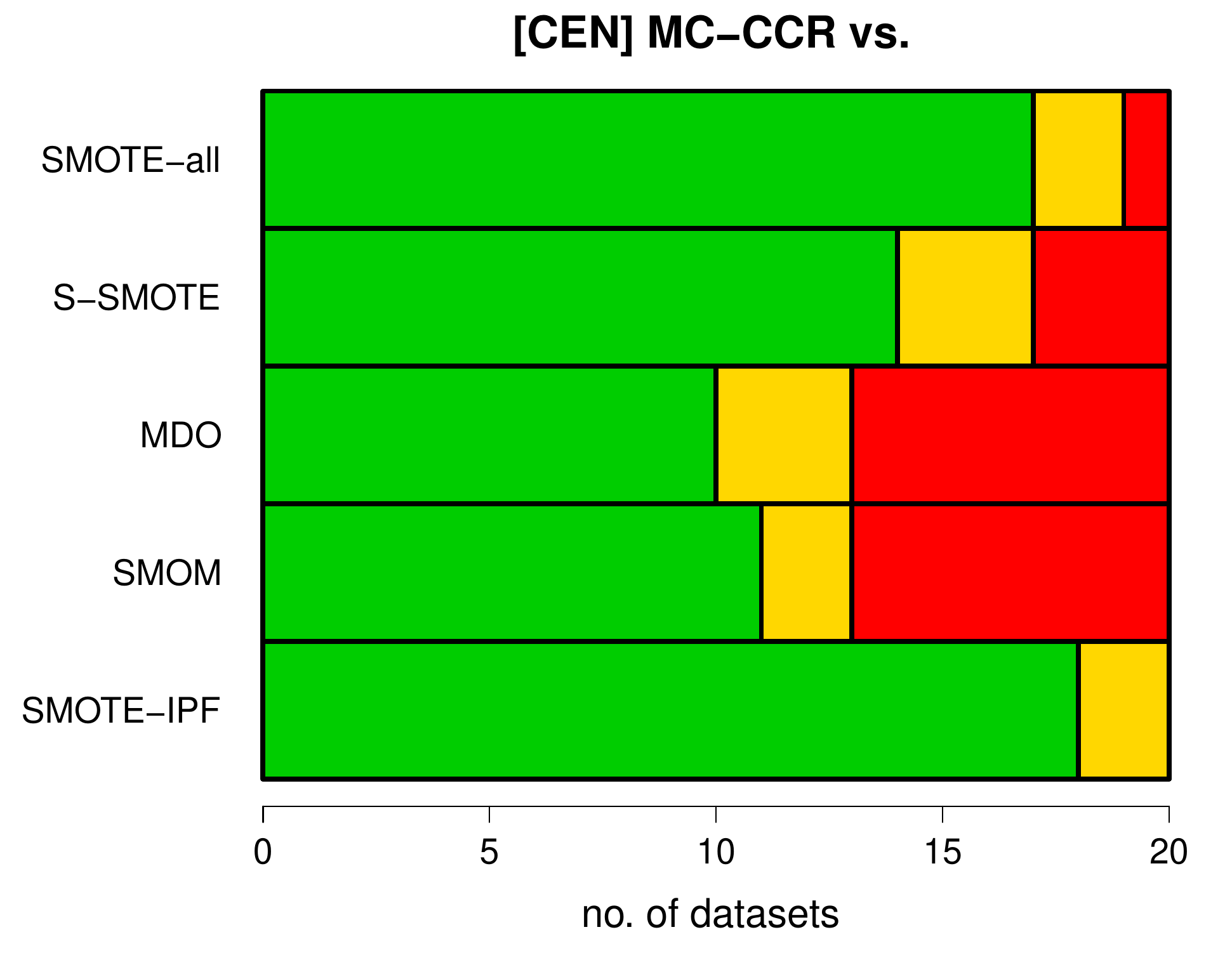}
        \caption{MLP}
    \end{subfigure}
        ~ 
    \begin{subfigure}[t]{0.99\textwidth}
    \captionsetup{font=scriptsize,labelfont=scriptsize}
        \centering        
        \includegraphics[width=0.24\textwidth]{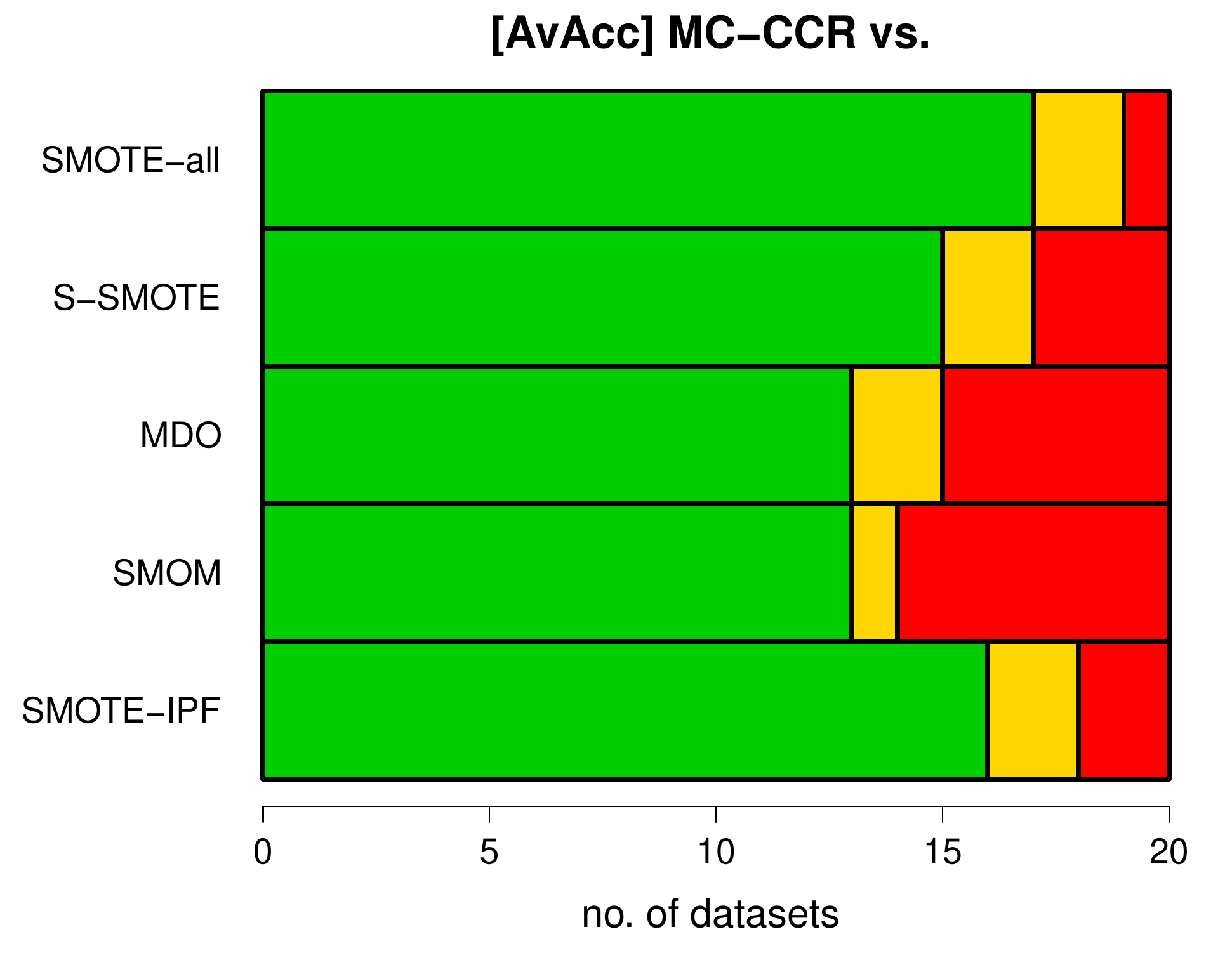}
        \includegraphics[width=0.24\textwidth]{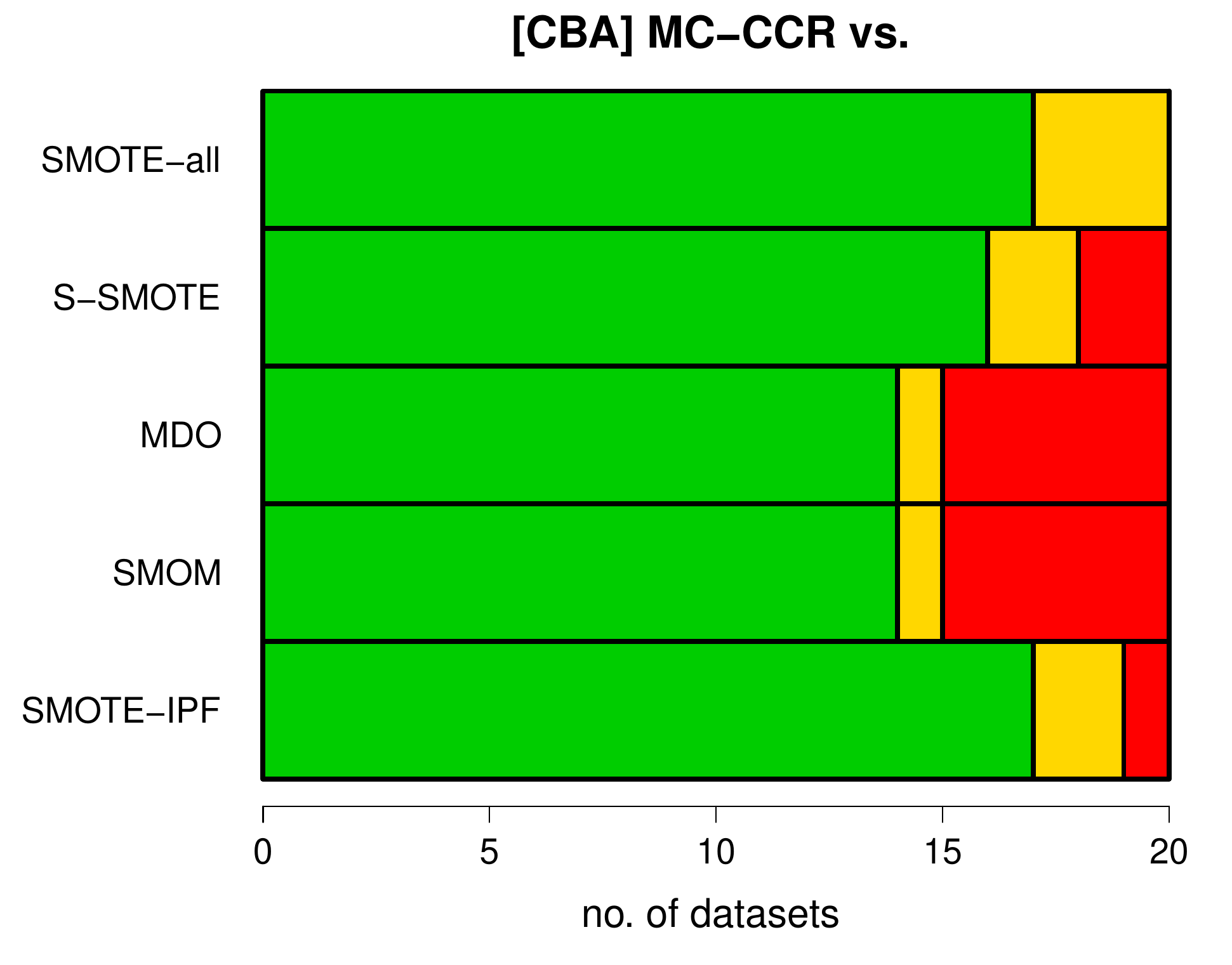}    
        \includegraphics[width=0.24\textwidth]{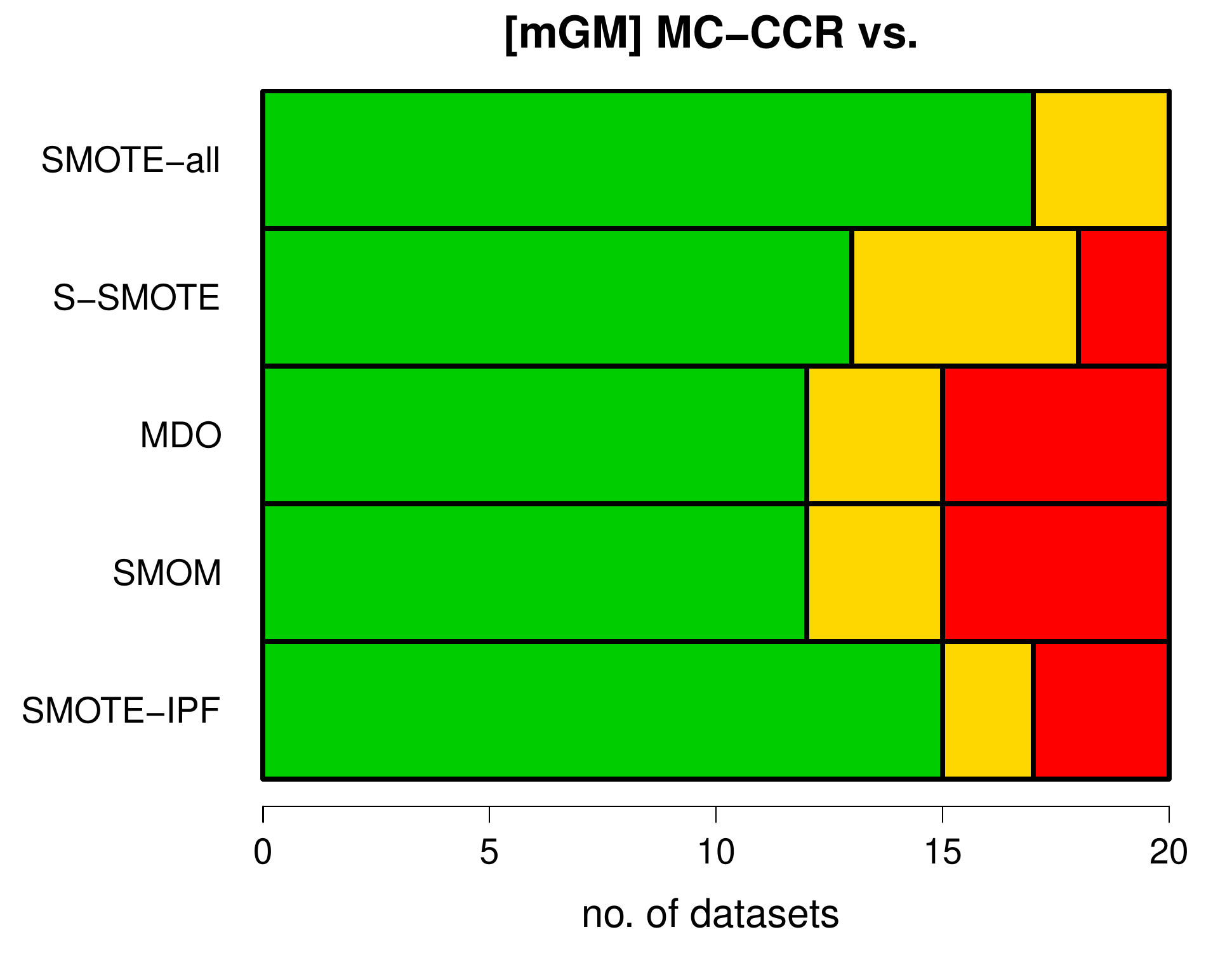}
        \includegraphics[width=0.24\textwidth]{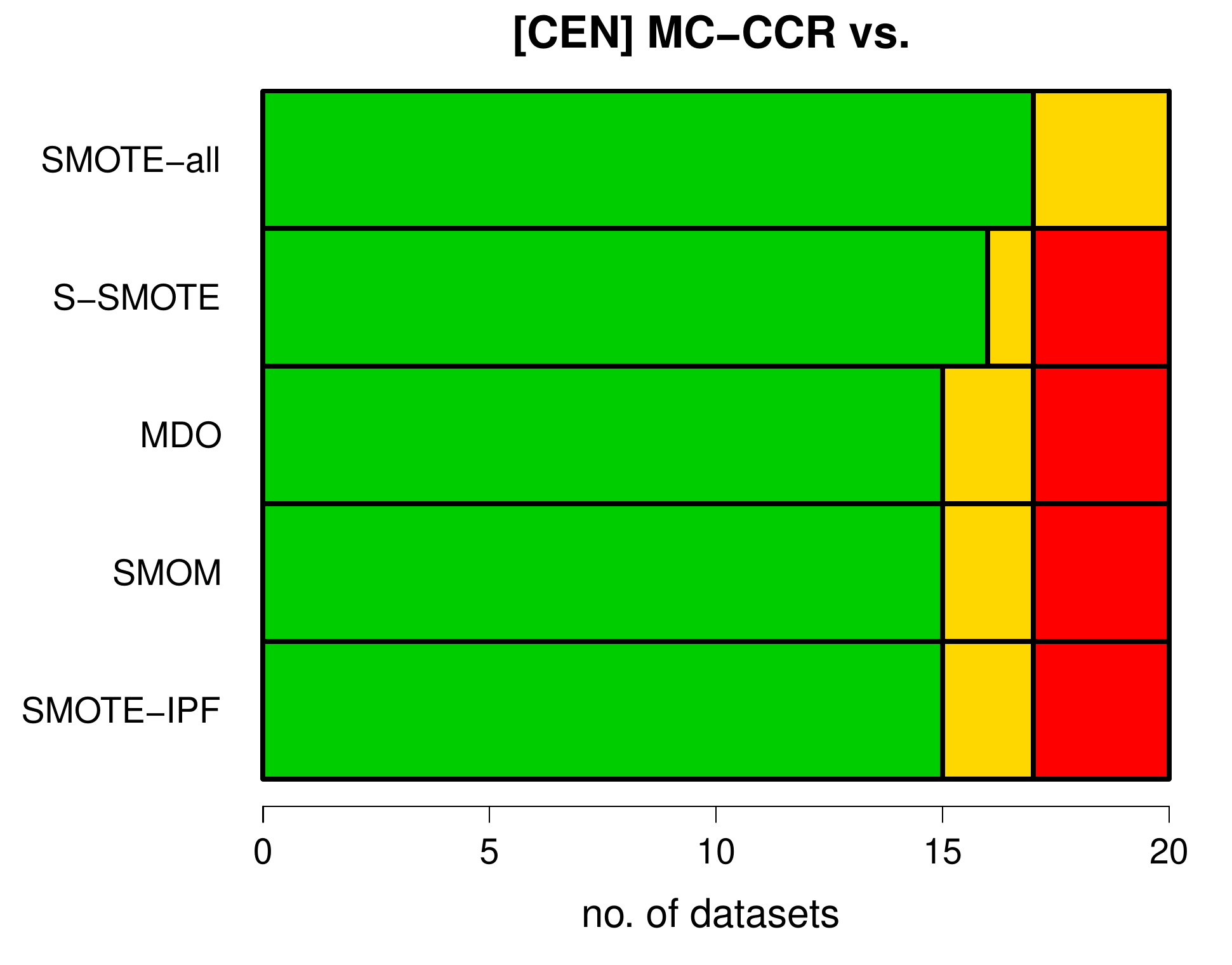}
        \caption{k-NN}
    \end{subfigure}
        ~ 
    \begin{subfigure}[t]{0.99\textwidth}
    \captionsetup{font=scriptsize,labelfont=scriptsize}
        \centering        
        \includegraphics[width=0.24\textwidth]{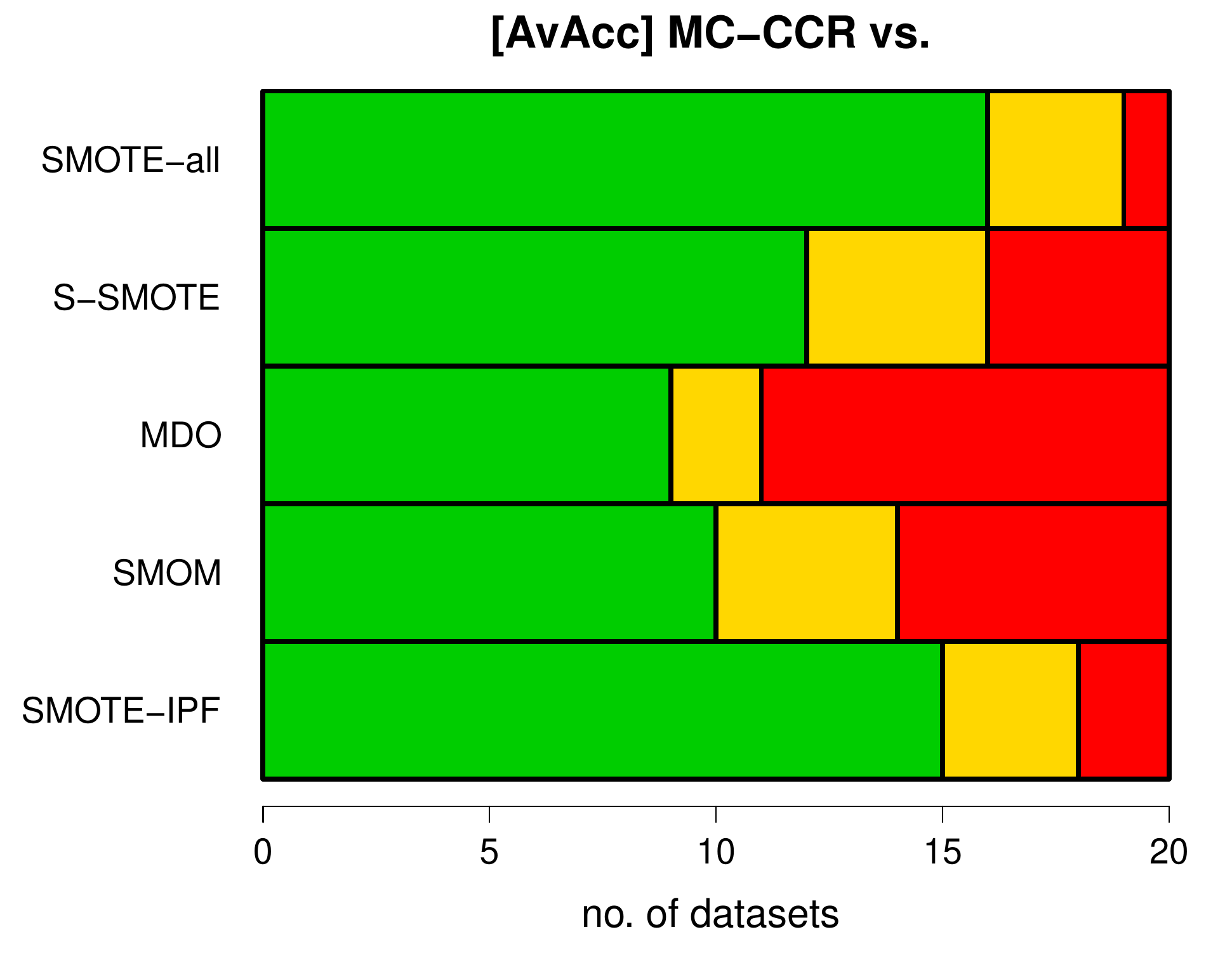}
        \includegraphics[width=0.24\textwidth]{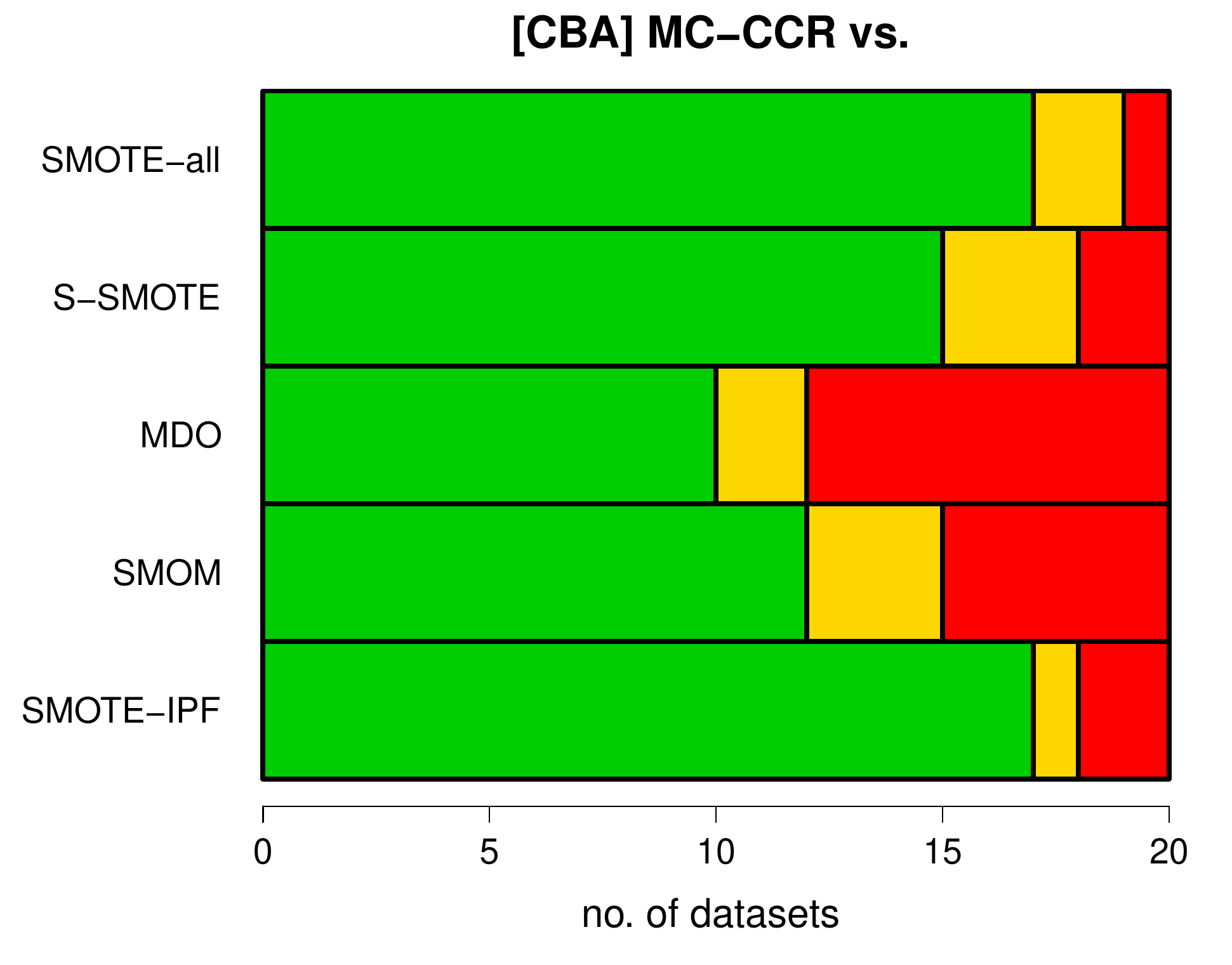}    
        \includegraphics[width=0.24\textwidth]{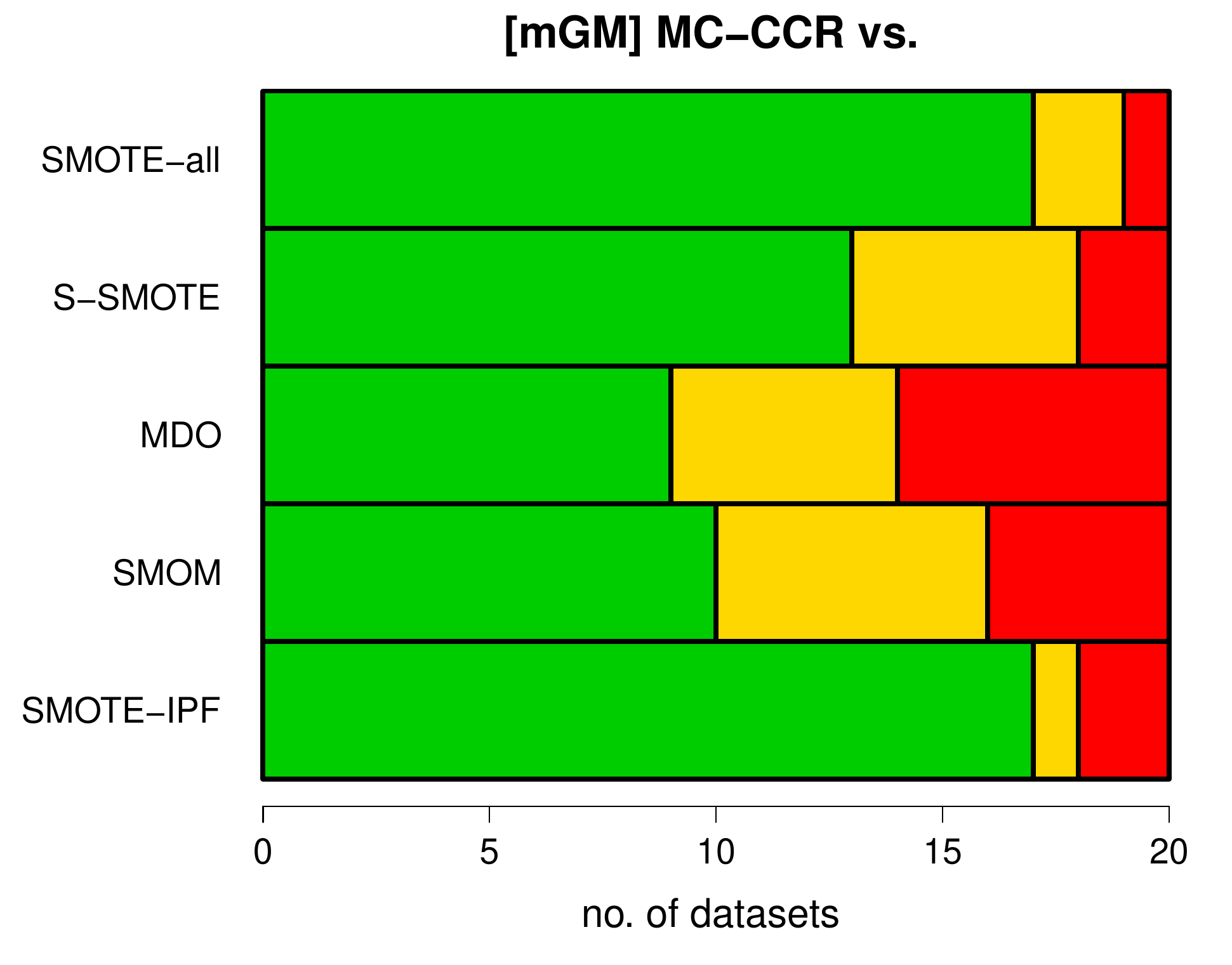}
        \includegraphics[width=0.24\textwidth]{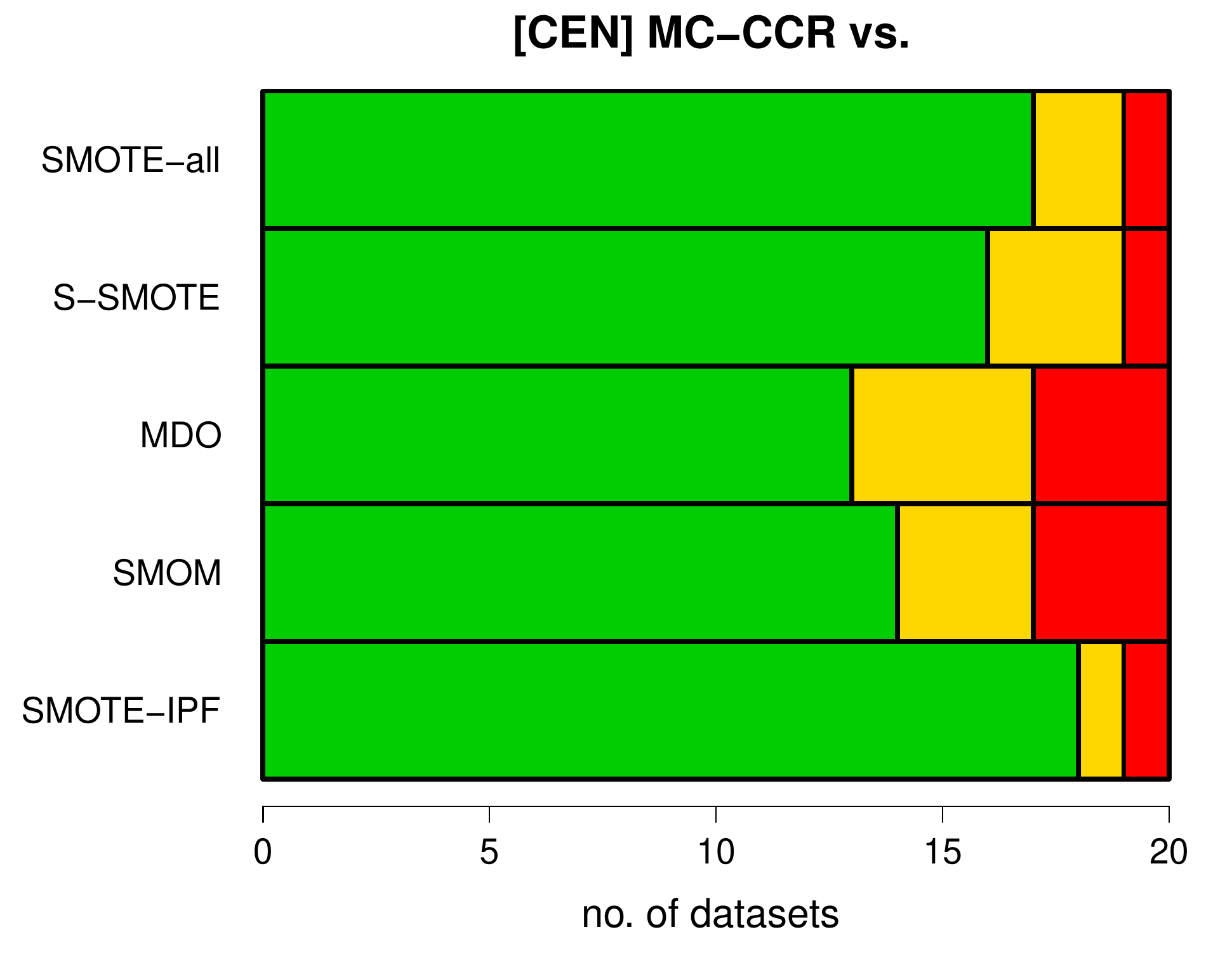}
        \caption{NB}
    \end{subfigure}
\caption{Comparison of MC-CCR with reference methods for four tested base classifiers with respect to the number of datasets on which MC-CCR was statistically significantly better (green), similar (yellow), or worse (red) using combined 10-fold CV F-test over 20 datasets.}
\label{fig:wins}
\end{figure*}

In the second stage of the conducted experimental study, we compared the proposed MC-CCR algorithm with the reference oversampling strategies to evaluate its relative usefulness. Detailed results on per-dataset basis for C5.0 classifier were presented in Tables~\ref{tab:avac50}--\ref{tab:cenc50}. In Figure~\ref{fig:wins}, we present the results of a win-loss-tie analysis, in which we compare the number of datasets on which MC-CCR achieved statistically significantly better, equal or worse performance than the individual methods on a pairwise basis, for all of the considered classifiers. Finally, in Table~\ref{tab:sh1} we present the $p$-values of comparison between all of the considered methods. As can be seen, in all of the cases, MC-CCR tended to outperform the oversampling reference strategies, which manifested in the highest average ranks concerning all of the performance metrics for C5.0 and a majority of wins in a per-dataset pairwise comparison of the methods.
Furthermore, the observed improvement in performance was statistically significant in comparison to most of the reference methods. In particular, when combined with the C5.0 and $k$-NN classifier, the proposed MC-CCR algorithm achieved a statistically significantly better performance than all of the reference methods. It is also worth noting that in the remainder of the cases, even if statistically significant differences at the significance level $\alpha = 0.05$ were not observed, the $p$-values remained small, indicating important differences.

\begin{table}[tbp]
\centering
\caption{Results according to AvAcc [\%] metric for MC-CCR and reference sampling methods with C5.0 as base classifier.}
\label{tab:avac50}
\scalebox{0.56}{
\begin{tabular}{lcccccc}
\toprule
\tabhead{Dataset}         & \tabhead{MC-CCR}    & \tabhead{SMOTE-all}  & \tabhead{S-SMOTE}   & \tabhead{MDO}    & \tabhead{SMOM} & \tabhead{SMOTE-IPF}\\
\midrule
Automobile      & 76.98   & \textbf{80.12}       & 73.53     & 78.13  & 79.04  & 75.32  \\
Balance         & \textbf{82.87}   & 55.06       & 55.01     & 57.70  & 59.52  &  54.26  \\
Car             & \textbf{97.12}   & 89.84       & 90.13     & 93.36  & 95.18  &  90.96  \\
Cleveland       & \textbf{37.88}   & 28.92       & 27.18     & 28.92  & 28.01  & 24.98  \\
Contraceptive   & 53.18   & 50.63       & 46.92     & 53.27  & \textbf{55.09}  & 52.88    \\
Dermatology     & 94.29   & 95.72       & 96.10     & 97.48  & \textbf{99.31}   & 92.18 \\
Ecoli           & \textbf{74.07}   & 64.68       & 67.54     & 61.16  & 61.16  & 60.43 \\
Flare           & 68.92& \textbf{71.86}   & 71.52       & 68.72     & 70.64  & 68.55     \\
Hayes-roth      & \textbf{92.11}   & 86.45       & 88.04     & 87.33  & 90.06  &   89.74  \\
Led7digit       & 70.48   & 72.39       & 72.55     & 75.03  &\textbf{75.94}  & 71.35    \\
Lymphography    & \textbf{79.60}   & 73.02       & 62.67     & 76.54  & 74.72  & 74.20 \\
Newthyroid      & \textbf{96.18}   & 94.70       & 93.48     & 92.06  & 90.24  &  93.05 \\
Pageblocks      & \textbf{83.71}   & 75.83       & 75.25     & 78.47  & 77.56  & 74.20 \\
Thyroid         & 80.52   & 80.02       & \textbf{85.34}     & 79.14  & 80.96  &  78.91   \\
Vehicle         & 72.71   & 73.49       & \textbf{73.71}     & 70.85  & 70.85  & 71.02  \\
Wine            & 95.28   & \textbf{92.53}       & 90.80     & 93.41  & 93.41  &  90.16\\
Winequality-red & \textbf{46.93}   & 37.41       & 35.79     & 40.05  & 42.78  & 36.28  \\
Yeast           & \textbf{58.39}   & 51.03       & 52.42     & 54.55  & 56.37  & 53.77\\
Zoo             & \textbf{85.92}   & 82.61       & 68.69     & 79.09  & 79.09  & 67.30  \\
\midrule
Avg. rank             &1.95 &5.55 & 3.10& 3.00 & 2.85 & 4.55 \\
\bottomrule
\end{tabular}
}
\end{table}

\begin{table}[tbp]
\centering
\caption{Results according to CBA [\%] metric for MC-CCR and reference sampling methods with C5.0 as base classifier.}
\label{tab:cbac50}
\scalebox{0.56}{
\begin{tabular}{lcccccc}
\toprule
\tabhead{Dataset}         & \tabhead{MC-CCR}    & \tabhead{SMOTE-all}  & \tabhead{S-SMOTE}   & \tabhead{MDO}    & \tabhead{SMOM} & \tabhead{SMOTE-IPF}\\
\midrule
Automobile&\textbf{77.93} &54.47&71.79&75.11&73.35& 70.84\\
Balance   &\textbf{64.92} &45.79&55.88&57.70&60.34& 57.29\\
Car &95.87&85.25&89.26&95.73& \textbf{98.37} & 87.51 \\
Cleveland & \textbf{33.91} &24.09&25.44&27.34&27.34& 21.99\\
Contraceptive   &50.01&41.63&44.31&51.69& \textbf{52.97} & 39.99\\
Dermatology     & \textbf{96.19} &86.30&94.36&95.64&93.88& 88.92\\
Ecoli     & 68.33 &56.07&\textbf{68.41}&63.53&61.77& 54.72\\
Flare     & 61.85&58.59&\textbf{61.92}&59.51&61.27& 60.03\\
Glass     & \textbf{69.89} &60.63&66.11&70.64&69.76& 61.02\\
Hayes-roth& \textbf{90.03} &77.48&86.30&84.96&87.62& 75.52\\
Led7digit       &\textbf{84.07}&63.94&70.81&78.19& 79.95 & 70.98\\
Lymphography    & \textbf{80.66} &43.21&59.19&75.75&78.39& 70.83\\
Newthyroid      &90.14&85.04&90.00& \textbf{95.22} &93.46& 89.69\\
Pageblocks&\textbf{84.63} &78.60&71.77&80.84&79.96& 76.35\\
Thyroid         &\textbf{81.99}&80.58&81.86&76.77&75.01& 70.46\\
Vehicle   &71.74 &62.18&70.23&\textbf{72.43}&\textbf{72.43}& 71.49\\
Wine      & \textbf{92.89} &84.30&91.67&91.83&90.95& 87.99\\
Winequality-red & 40.37 &24.76&34.92&\textbf{41.63}&39.87& 40.01\\
Yeast     & \textbf{57.83}   &46.44&48.94&53.76&55.52& 49.03\\
Zoo       & \textbf{81.52}   &60.83&66.08&76.72&76.72& 79.05\\
\midrule
Avg. rank             &1.65 & 5.75 & 3.65&3.15 &2.75 & 4.05  \\
\bottomrule
\end{tabular}
}
\end{table}

\begin{table}[tbp]
\centering
\caption{Results according to mGM [\%] metric for MC-CCR and reference sampling methods with C5.0 as base classifier.}
\label{tab:mgmc50}
\scalebox{0.56}{
\begin{tabular}{lcccccc}
\toprule
\tabhead{Dataset}         & \tabhead{MC-CCR}    & \tabhead{SMOTE-all}  & \tabhead{S-SMOTE}   & \tabhead{MDO}    & \tabhead{SMOM} & \tabhead{SMOTE-IPF}\\
\midrule
Automobile&75.68&51.86&74.4&75.11& \textbf{78.75}& 73.28 \\
Balance   &\textbf{62.22}&40.57&52.4&55.92&58.65 & 55.69 \\
Car       & 95.03 &81.16&91.00&\textbf{95.73}&94.82 & 90.85 \\
Cleveland & \textbf{30.98}&18.00&24.57&26.45&24.63 & 23.39 \\
Contraceptive   &49.72&38.15&43.44&51.69&\textbf{52.60} & 50.07 \\
Dermatology &\textbf{94.88}&79.34&95.23&92.08&92.08 & 93.02 \\
Ecoli     & \textbf{70.98} &51.72&67.54&63.53&63.53 & 62.88 \\
Flare     & \textbf{64.71} &55.98&61.92&58.62&60.44 & 57.29 \\
Glass     & \textbf{71.06} &55.41&62.63&67.97&67.97 & 66.77 \\
Hayes-roth& \textbf{85.16}&70.52&84.56&83.18&86.82 & 84.20 \\
Led7digit &\textbf{80.44}&57.85&70.81& 78.19 &77.28 &75.48 \\
Lymphography&\textbf{77.36} &40.60&58.32&73.08&73.99 & 71.86 \\
Newthyroid& 92.17 &80.69&86.52&92.55&\textbf{93.46} & 92.55 \\
Pageblocks& \textbf{82.55} &71.64&74.38&79.06&77.24 & 76.72 \\
Thyroid   &81.48&73.62&\textbf{81.86}&75.88&77.70 & 75.10  \\
Vehicle   &70.35&59.57&71.97&\textbf{70.65}&\textbf{70.65} & 69.37 \\
Wine      &92.87&80.82& \textbf{93.41}  &90.05&89.14 & 90.08\\
Winequality-red & \textbf{46.66} &17.80&36.66&40.74&42.56 & 38.42 \\
Yeast     & \textbf{56.91} &43.83&52.42&51.09&52.91 & 50.03 \\
Zoo       & \textbf{84.29} &55.61&63.47&75.83&77.65 & 78.55 \\
\midrule
Avg. rank             & 1.80& 5.45& 3.25 &2.90 &2.70 & 4.9  \\
\bottomrule
\end{tabular}
}
\end{table}

\begin{table}[tbp]
\centering
\caption{Results according to CEN metric for MC-CCR and reference sampling methods with C5.0 as base classifier.}
\label{tab:cenc50}
\scalebox{0.56}{
\begin{tabular}{lcccccc}
\toprule
\tabhead{Dataset}         & \tabhead{MC-CCR}    & \tabhead{SMOTE-all}  & \tabhead{S-SMOTE}   & \tabhead{MDO}    & \tabhead{SMOM} & \tabhead{SMOTE-IPF}\\
\midrule
Automobile&\textbf{0.25}  &0.56&0.29&0.29&0.30 & 0.29 \\
Balance   &\textbf{0.41}   &0.64&0.49&0.48&0.46 & 0.49 \\
Car&0.09&0.26&0.11&0.09& \textbf{0.04}  & 0.10 \\
Cleveland       &\textbf{0.62}          &0.87&0.76&0.75&0.76 & 0.71 \\
Contraceptive   &0.51&0.69&0.59&0.49& \textbf{0.44}  &  0.48\\
Dermatology     & \textbf{0.06}     &0.27&\textbf{0.06}&0.13&0.14 & 0.21 \\
Ecoli     &\textbf{0.32}    &0.53&0.34&0.41&0.40 & 0.37 \\
Flare     &\textbf{0.31}  &0.51&0.39&0.46&0.41 & 0.49 \\
Glass           &0.38&0.53&0.38&\textbf{0.34}&\textbf{0.34} & 0.37 \\
Hayes-roth      &0.16&0.34&0.19&0.19& \textbf{0.14}  & 0.35 \\
Led7digit &0.15     &\textbf{0.20}&0.32&0.24&0.21 & 0.29 \\
Lymphography    & \textbf{0.26}  &0.68&0.43&0.31&0.34 & 0. 40 \\
Newthyroid&\textbf{0.08}  &0.28&0.14&0.08&0.12 & 0.15 \\
Pageblocks& \textbf{0.13} &0.34&0.29&0.24&0.19 & 0.26 \\
Thyroid   & \textbf{0.17}  &0.34&0.22&0.27&0.29 & 0.29 \\
Vehicle   & 0.35 &0.47&0.30 & 0.34&\textbf{0.33} &  0.37\\
Wine            & \textbf{0.11} &0.25&0.10&0.15&0.18 & 0.20 \\
Winequality-red &\textbf{0.49}   &0.86&0.67&0.64&0.59 & 0.56 \\
Yeast     & \textbf{0.45} &0.62&0.51&0.54&0.47 & 0.53\\
Zoo       &\textbf{0.19}    &0.48&0.38&0.27&0.27 & 0.22 \\

\midrule
Avg. rank             & 1.25& 5.70 & 3.70& 3.05& 2.95& 4.35 \\
\bottomrule
\end{tabular}
}
\end{table}

\begin{table}[!ht]
\small
 \centering
 \caption{Shaffer's tests for comparison between MC-CCR and reference oversampling methods with respect to each metric and base classifier. We report obtained $p$-values. Symbol '$>$' stands for situation in which MC-CCR is statistically superior and '$=$' for when there are no significant differences.}
 \label{tab:sh1}
 \scalebox{0.8}{
 \begin{tabular}{lcccc}
 \toprule
 \tabhead{Hypothesis} &  \tabhead{AvACC} &  \tabhead{CBA} &  \tabhead{mGM} &  \tabhead{CEN}\\
\midrule
C5.0 & & & & \\
vs. SMOTE-all & $>$ (0.0000) & $>$ (0.0004) & $>$ (0.0001) & $>$ (0.0007)\\
vs. S-SMOTE & $>$ (0.0326) & $>$ (0.0105) & $>$ (0.0194) & $>$ (0.0111)\\
vs. MDO & $>$ (0.0407) & $>$ (0.0396) & $>$ (0.0482) & $>$ (0.0433)\\
vs. SMOM & $>$ (0.0471) & $>$ (0.0412) & $>$ (0.0408) & $>$ (0.0399)\\
vs. SMOTE-IPF & $>$ (0.0301) & $>$ (0.0173) & $>$ (0.0188) & $>$ (0.0136)\\
\midrule
MLP & & & & \\
vs. SMOTE-all & $>$ (0.0018) & $>$ (0.0049) & $>$ (0.0039) & $>$ (0.0027)\\
vs. S-SMOTE & $>$ (0.0277) & $>$ (0.0261) & $>$ (0.0302) & $>$ (0.0166)\\
vs. MDO & $=$ (0.1307) & $=$ (0.0852) & $=$ (0.1003) & $=$ (0.0599)\\
vs. SMOM & $=$ (0.1419) & $=$ (0.1001) & $=$ (0.1188) & $=$ (0.0627)\\
vs. SMOTE-IPF & $>$ (0.0318) & $>$ (0.0251) & $>$ (0.0292) & $>$ (0.0199)\\
\midrule
k-NN & & & & \\
vs. SMOTE-all & $>$ (0.0000) & $>$ (0.007) & $>$ (0.0012) & $>$ (0.0008)\\
vs. S-SMOTE & $>$ (0.0385) & $>$ (0.0111) & $>$ (0.0358) & $>$ (0.0122)\\
vs. MDO & $>$ (0.0372) & $>$ (0.0316) & $>$ (0.0386) & $>$ (0.0355)\\
vs. SMOM & $>$ (0.0407) & $>$ (0.0394) & $>$ (0.0388) & $>$ (0.0401)\\
vs. SMOTE-IPF & $>$ (0.0174) & $>$ (0.0116) & $>$ (0.0158) & $>$ (0.0099)\\
\midrule
NB & & & & \\
vs. SMOTE-all & $>$ (0.0000) & $>$ (0.0001) & $>$ (0.0002) & $>$ (0.0001)\\
vs. S-SMOTE & $>$ (0.0162) & $>$ (0.0105) & $>$ (0.0122) & $>$ (0.0088)\\
vs. MDO & $=$ (0.0866) & $=$ (0.0681) & $=$ (0.0791) & $>$ (0.0372)\\
vs. SMOM & $=$ (0.1283) & $=$ (0.0599) & $=$ (0.0629) & $>$ (0.0498)\\
vs. SMOTE-IPF & $>$ (0.0126) & $>$ (0.0093) & $>$ (0.0127) & $>$ (0.0088)\\
\bottomrule
 \end{tabular}}
\end{table}

\subsection{Evaluation of the impact of class label noise}
\label{sec:exp3}
Finally, we evaluated how the presence of the label noise affects the predictive performance of MC-CCR compared to the state-of-the-art algorithms.
To input the noise, we decided to use random label noise imputation, i.e., according to a given noise level and the uniform distribution, we choose a subset of training examples and replace their labels to randomly chosen remaining ones. 
In our experimental study, we limited ourselves to the noise levels in $\{0.0, 0.05, 0.1, 0.15, 0.20, 0.25\}$.
 
\begin{figure*}
\centering
        \includegraphics[width=0.24\textwidth]{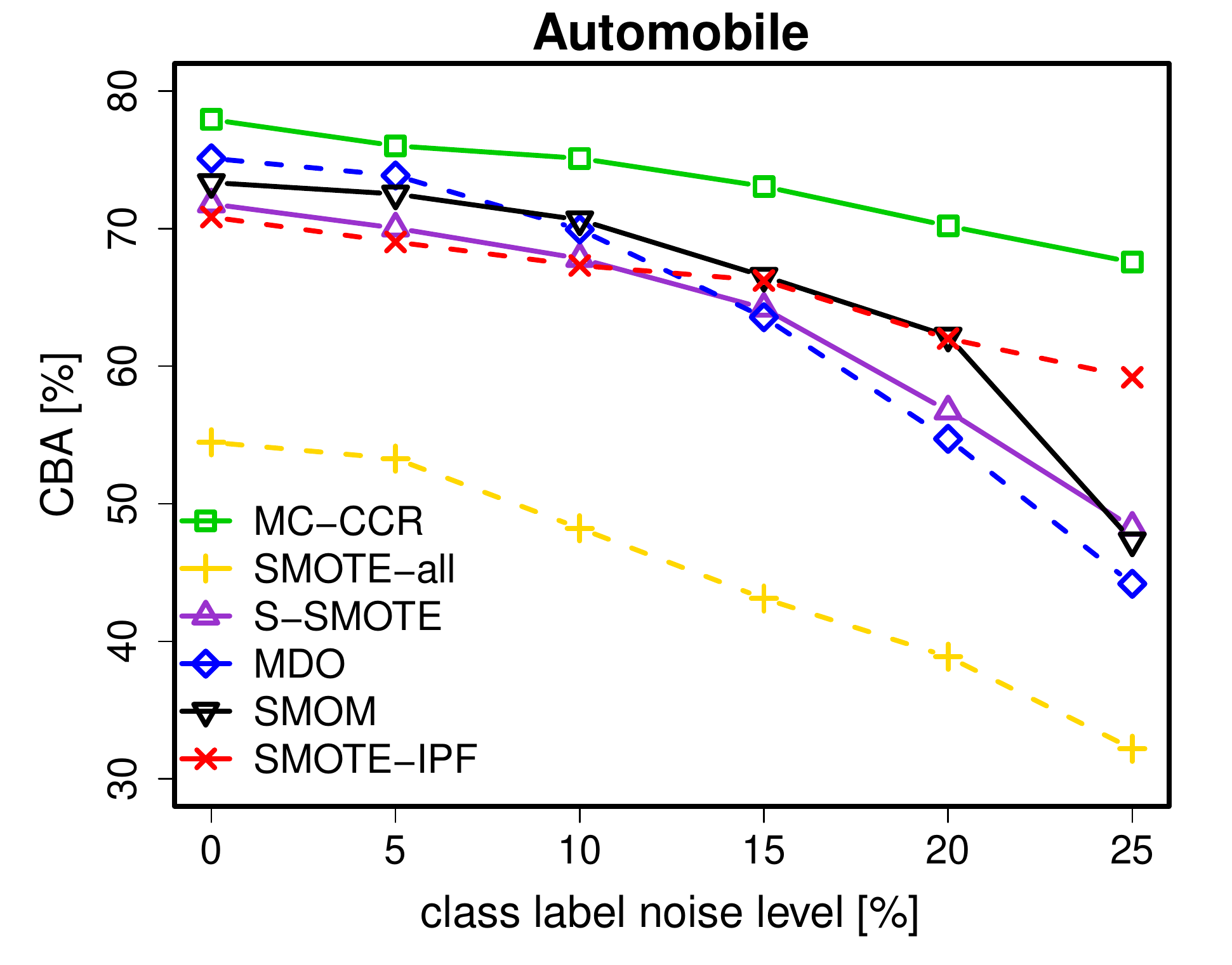}
        \includegraphics[width=0.24\textwidth]{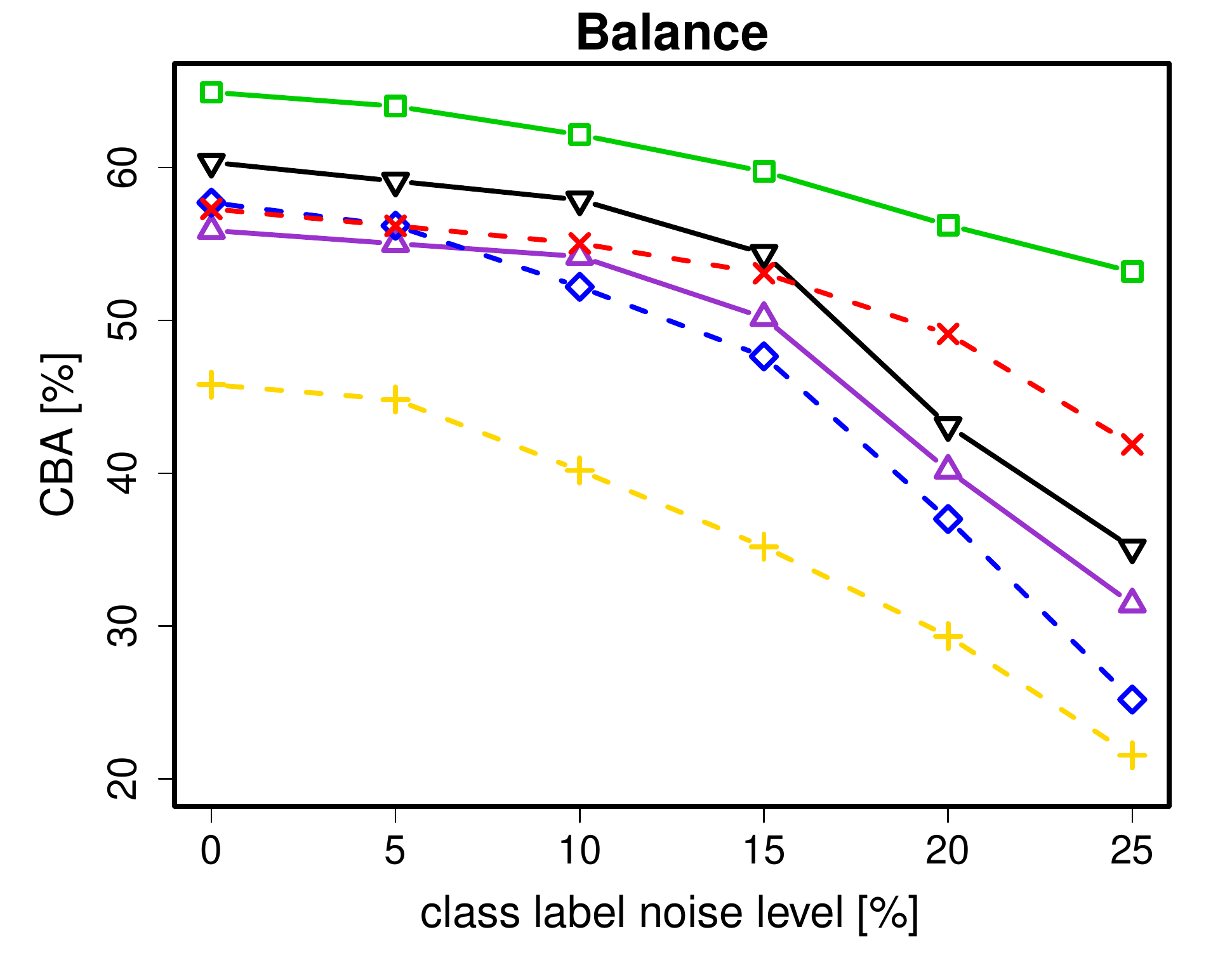}
        \includegraphics[width=0.24\textwidth]{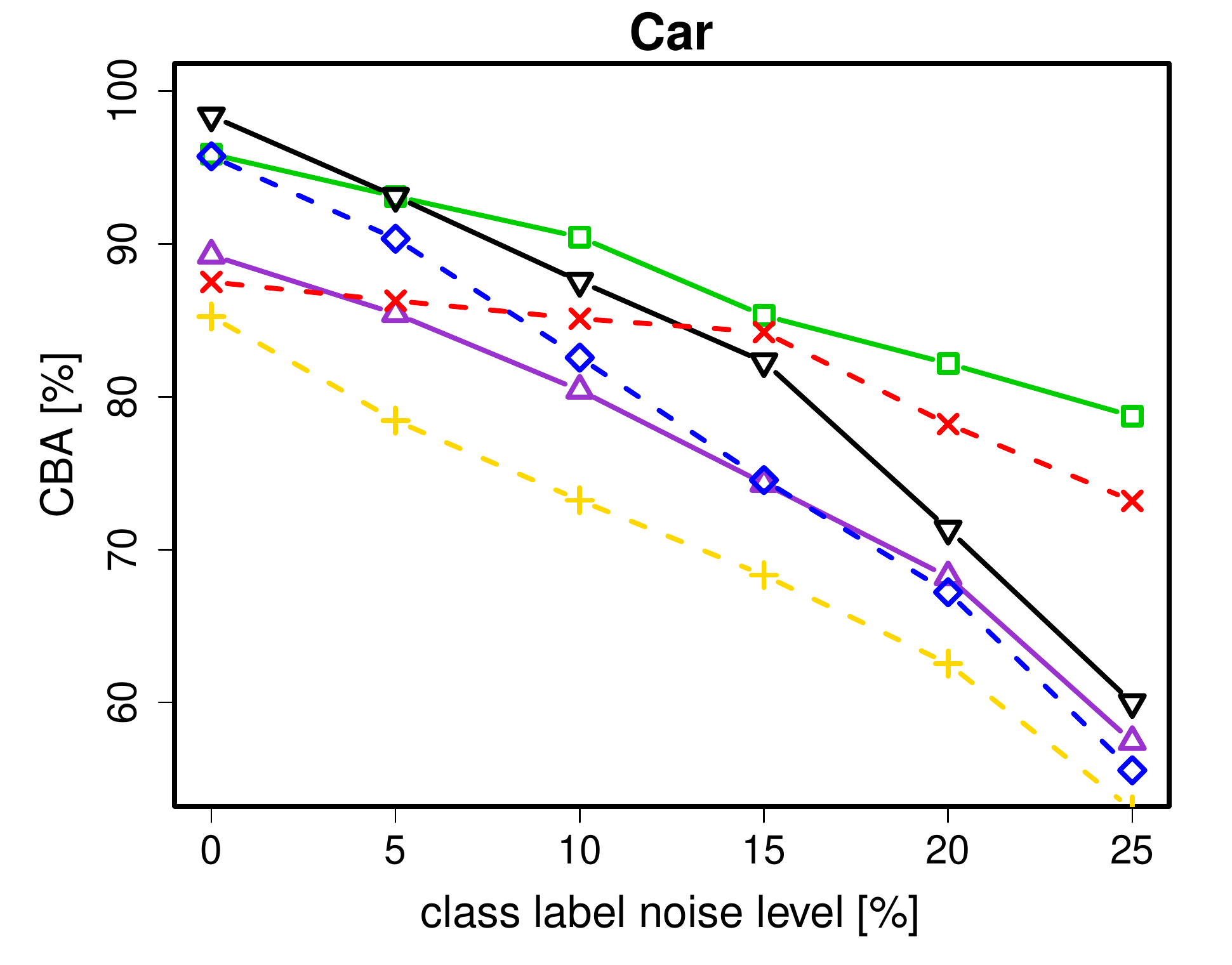}
        \includegraphics[width=0.24\textwidth]{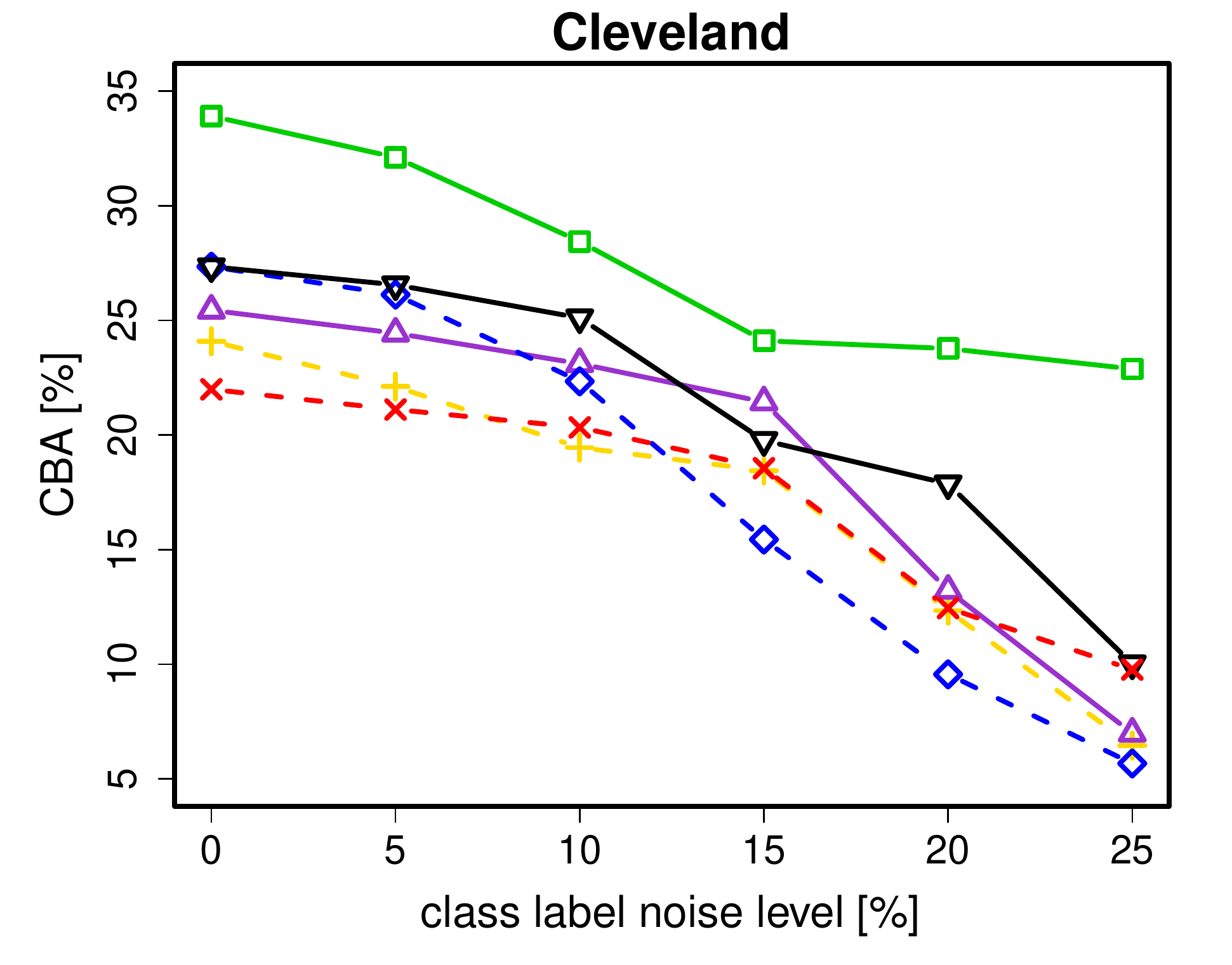}
        \includegraphics[width=0.24\textwidth]{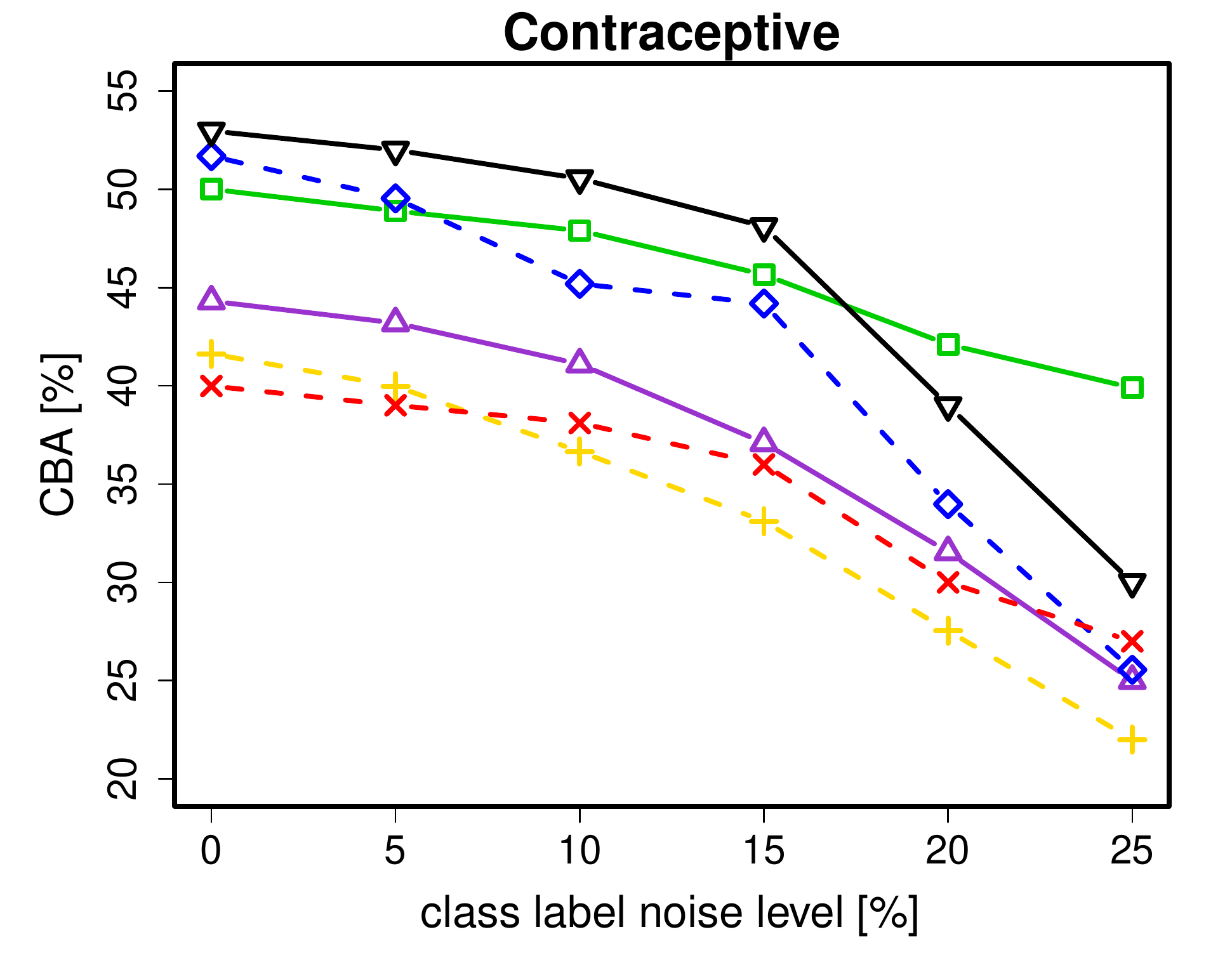}
        \includegraphics[width=0.24\textwidth]{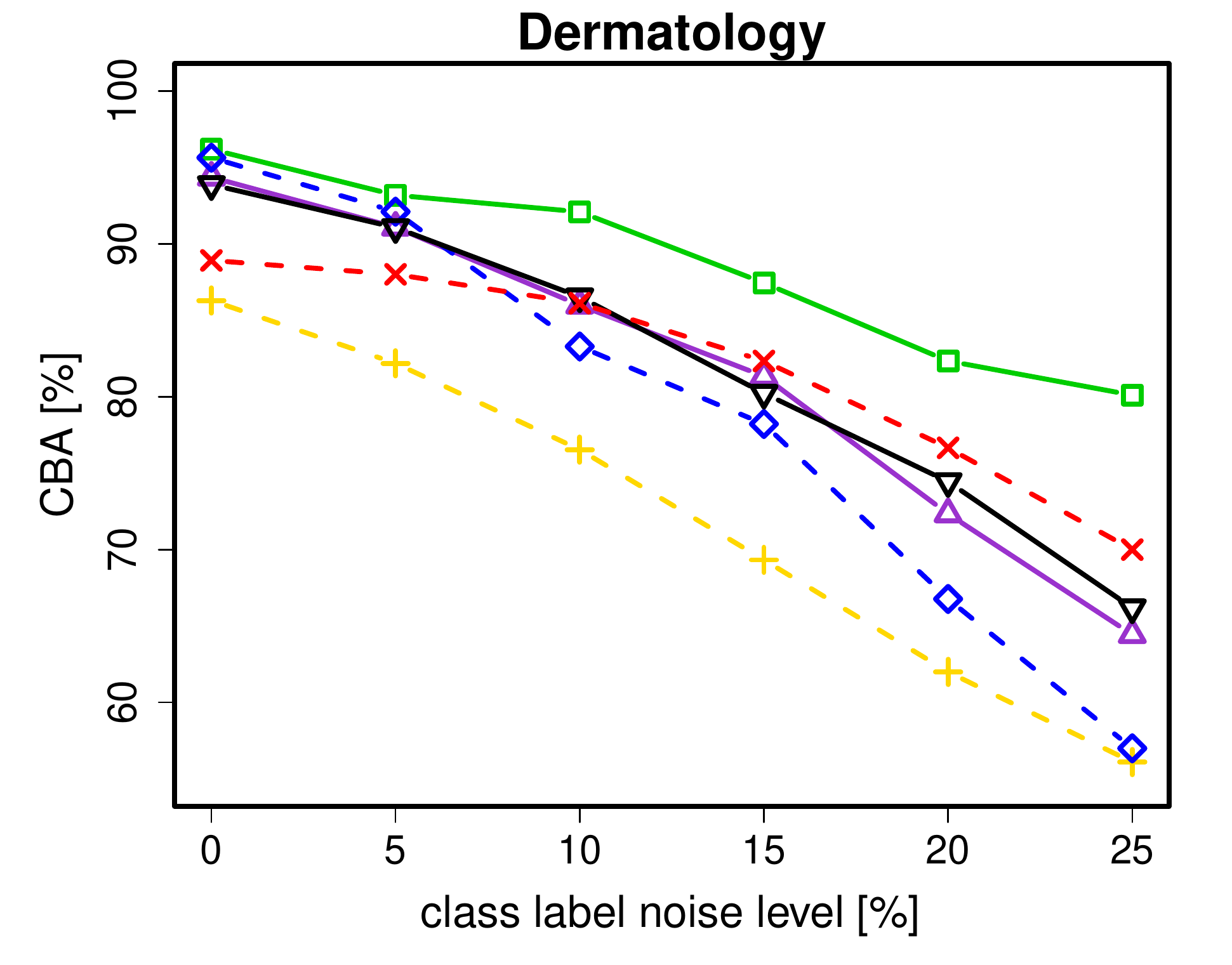}
        \includegraphics[width=0.24\textwidth]{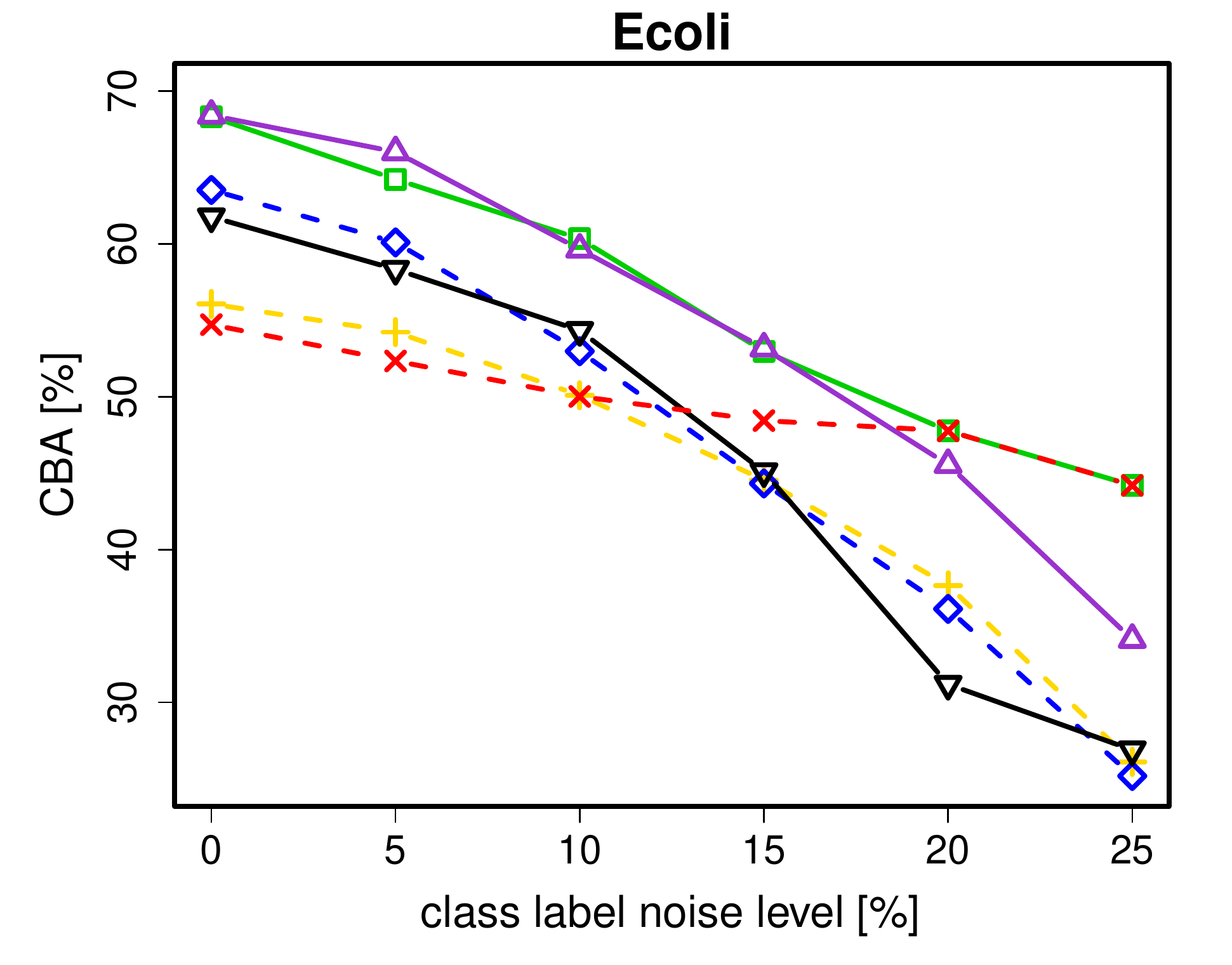}
        \includegraphics[width=0.24\textwidth]{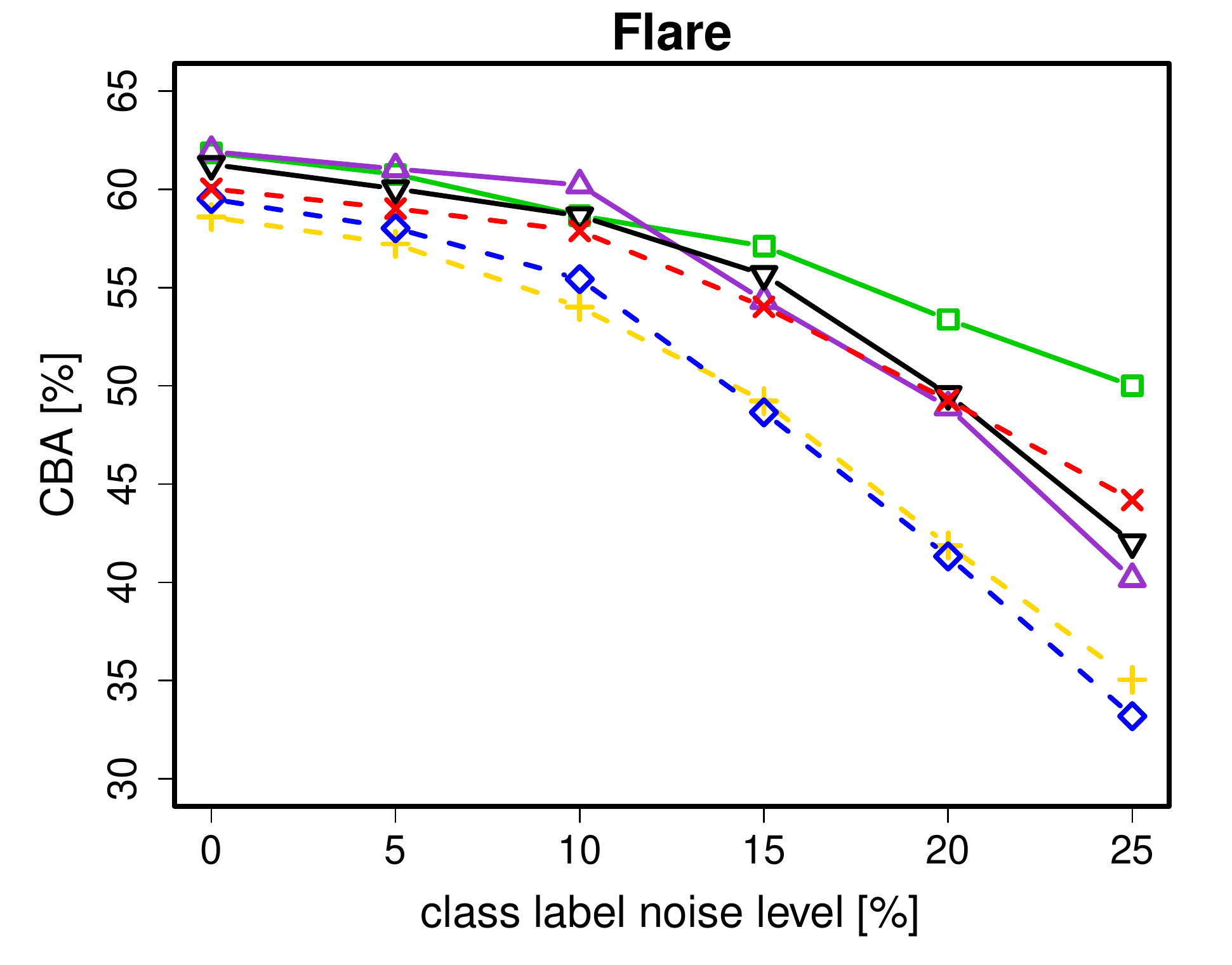}
        \includegraphics[width=0.24\textwidth]{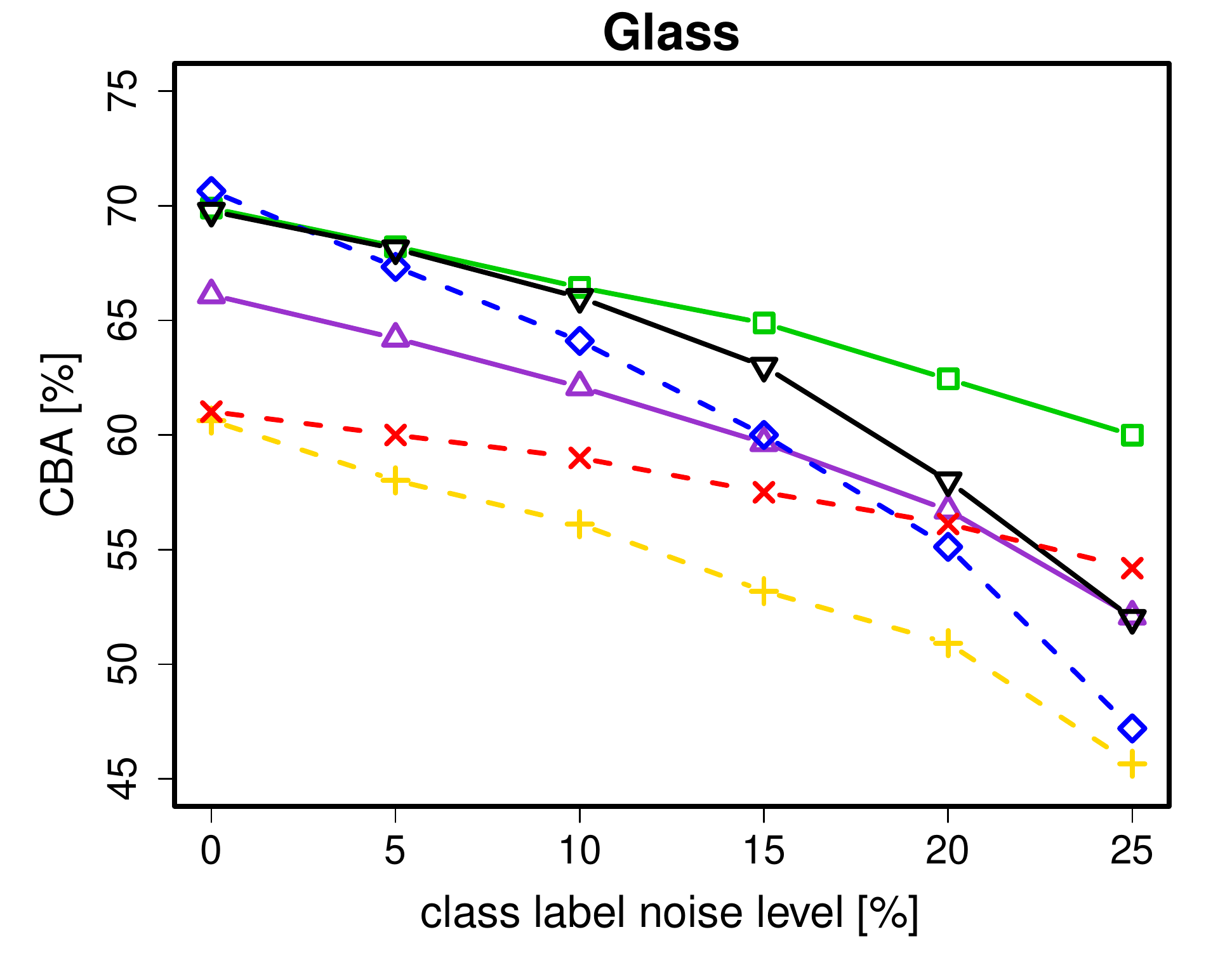}
        \includegraphics[width=0.24\textwidth]{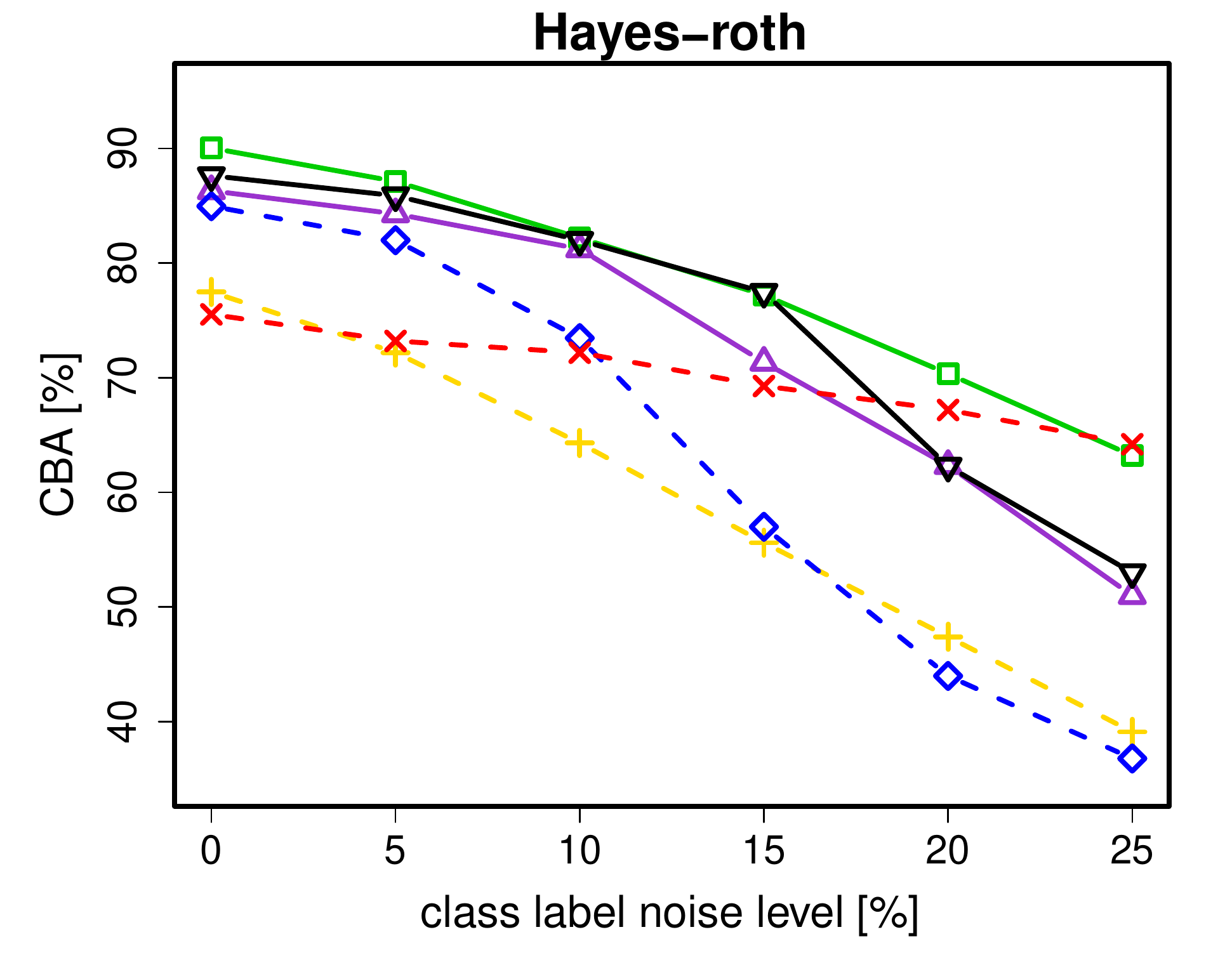}
        \includegraphics[width=0.24\textwidth]{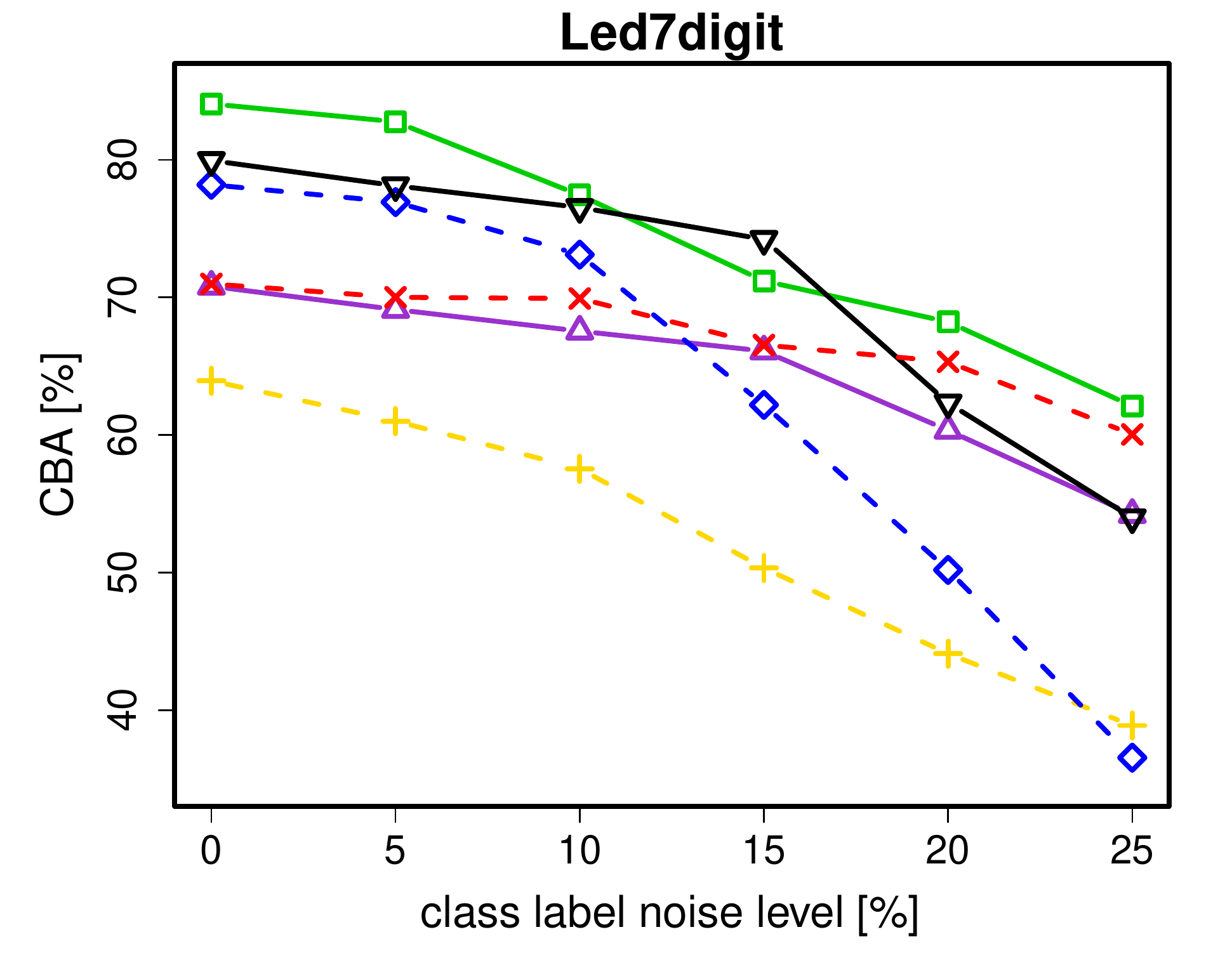}
        \includegraphics[width=0.24\textwidth]{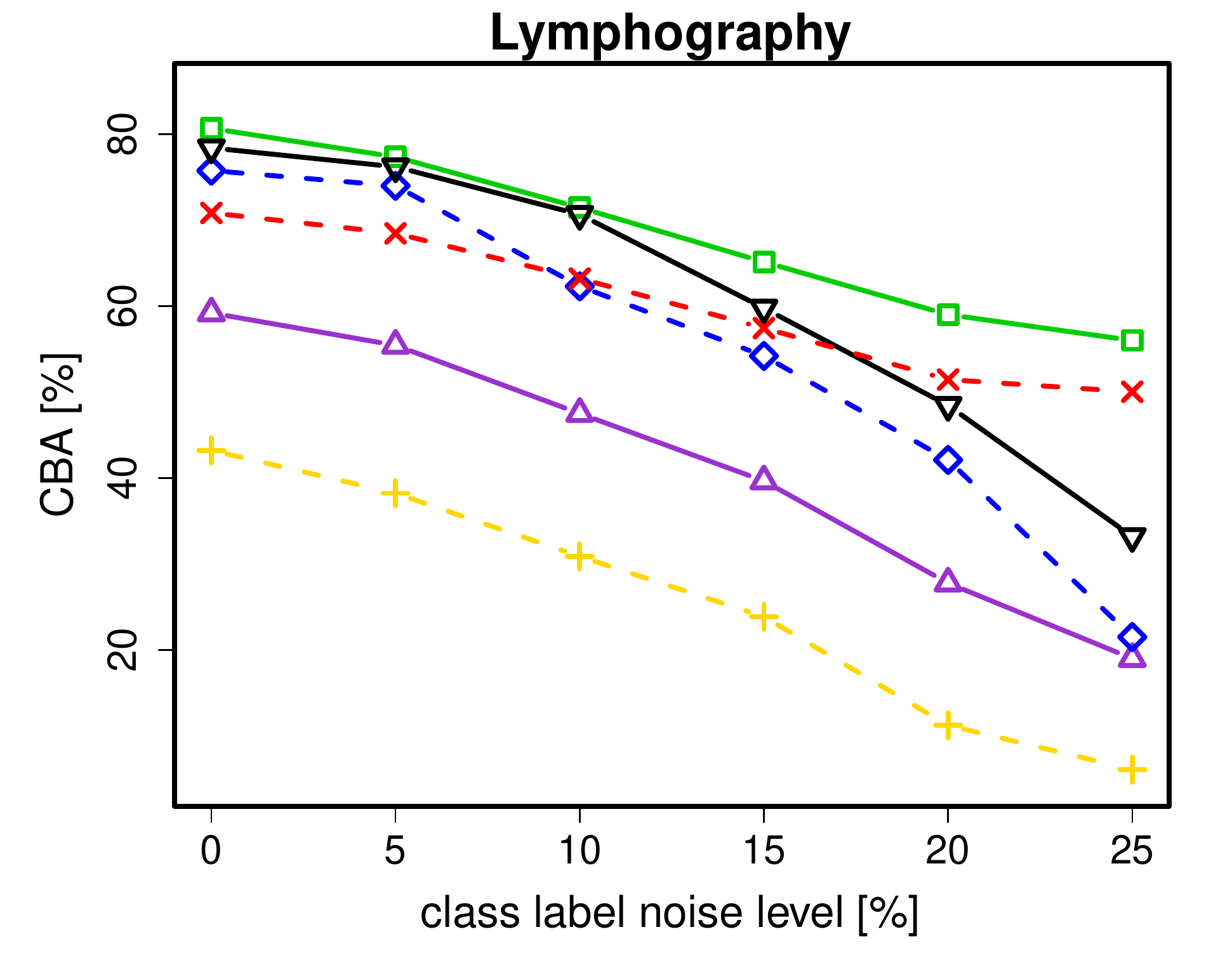}
        \includegraphics[width=0.24\textwidth]{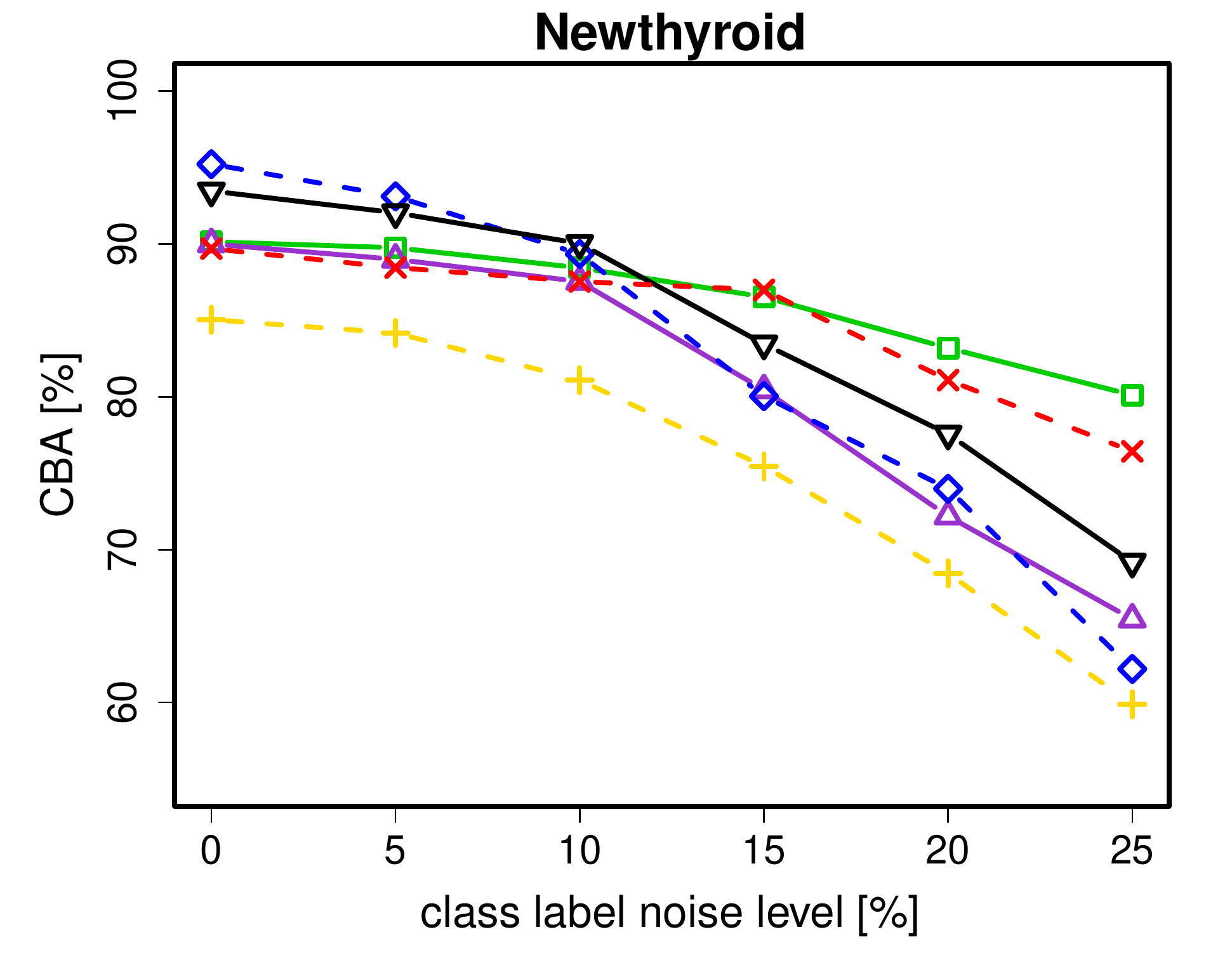}
        \includegraphics[width=0.24\textwidth]{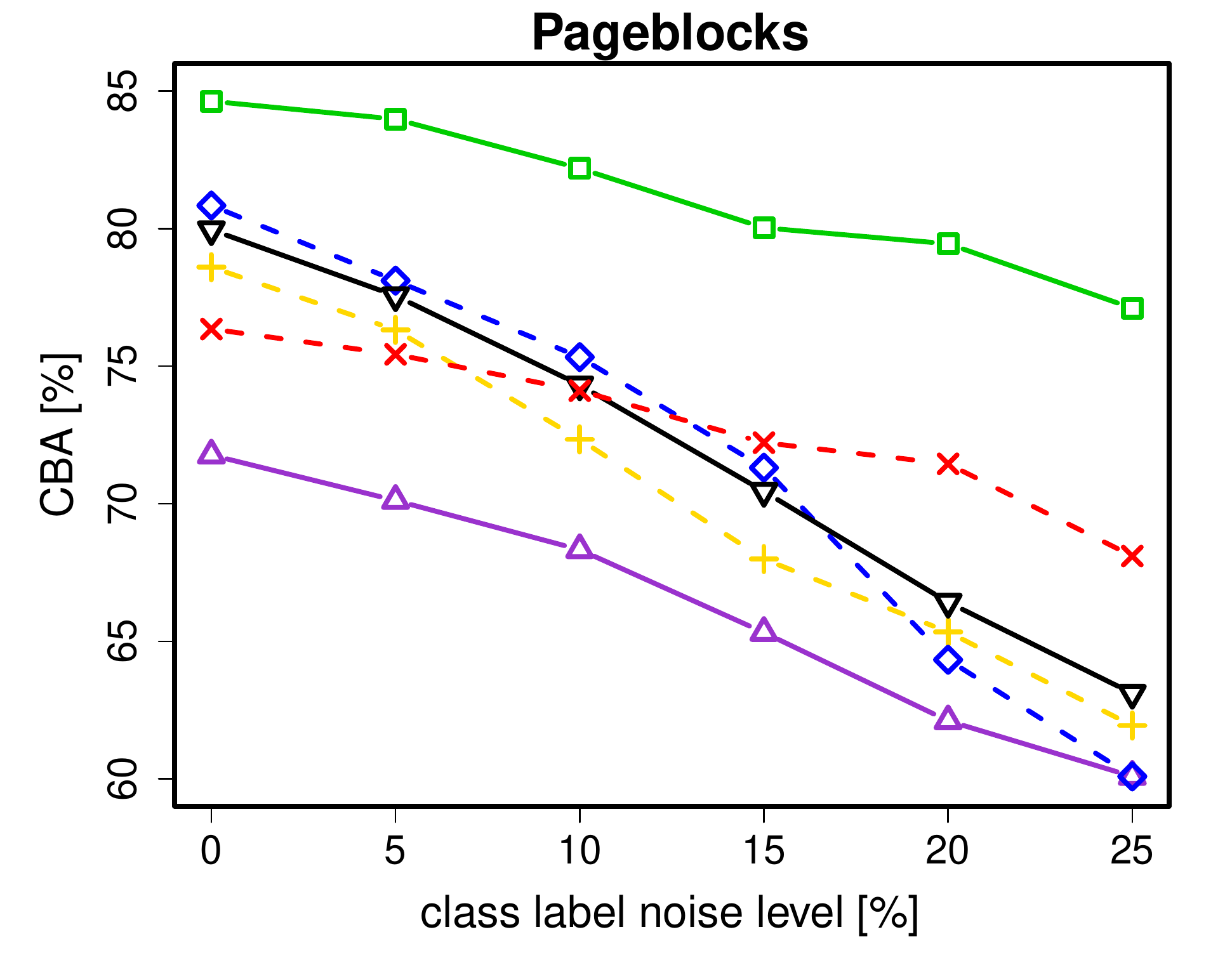}
        \includegraphics[width=0.24\textwidth]{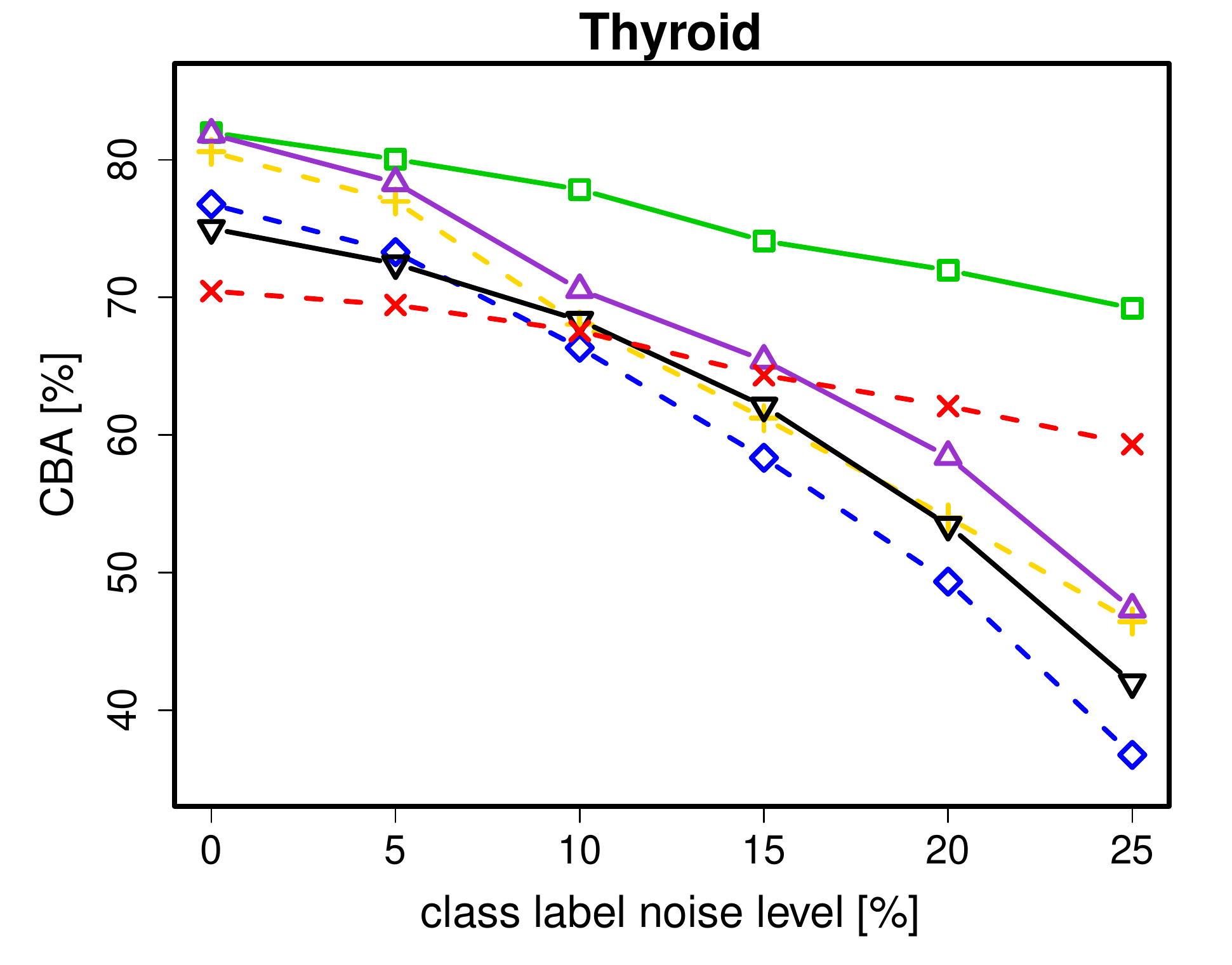}
        \includegraphics[width=0.24\textwidth]{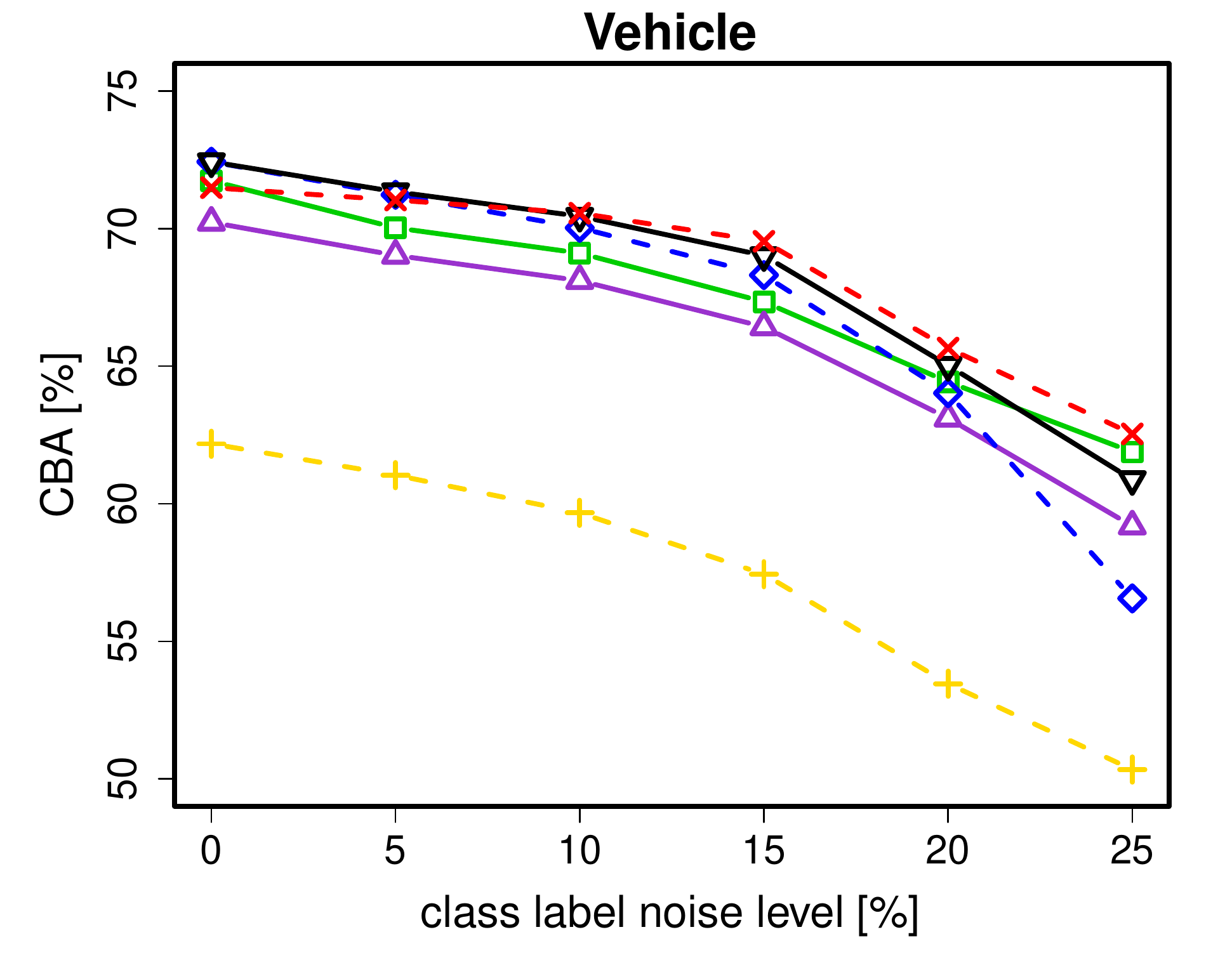}
        \includegraphics[width=0.24\textwidth]{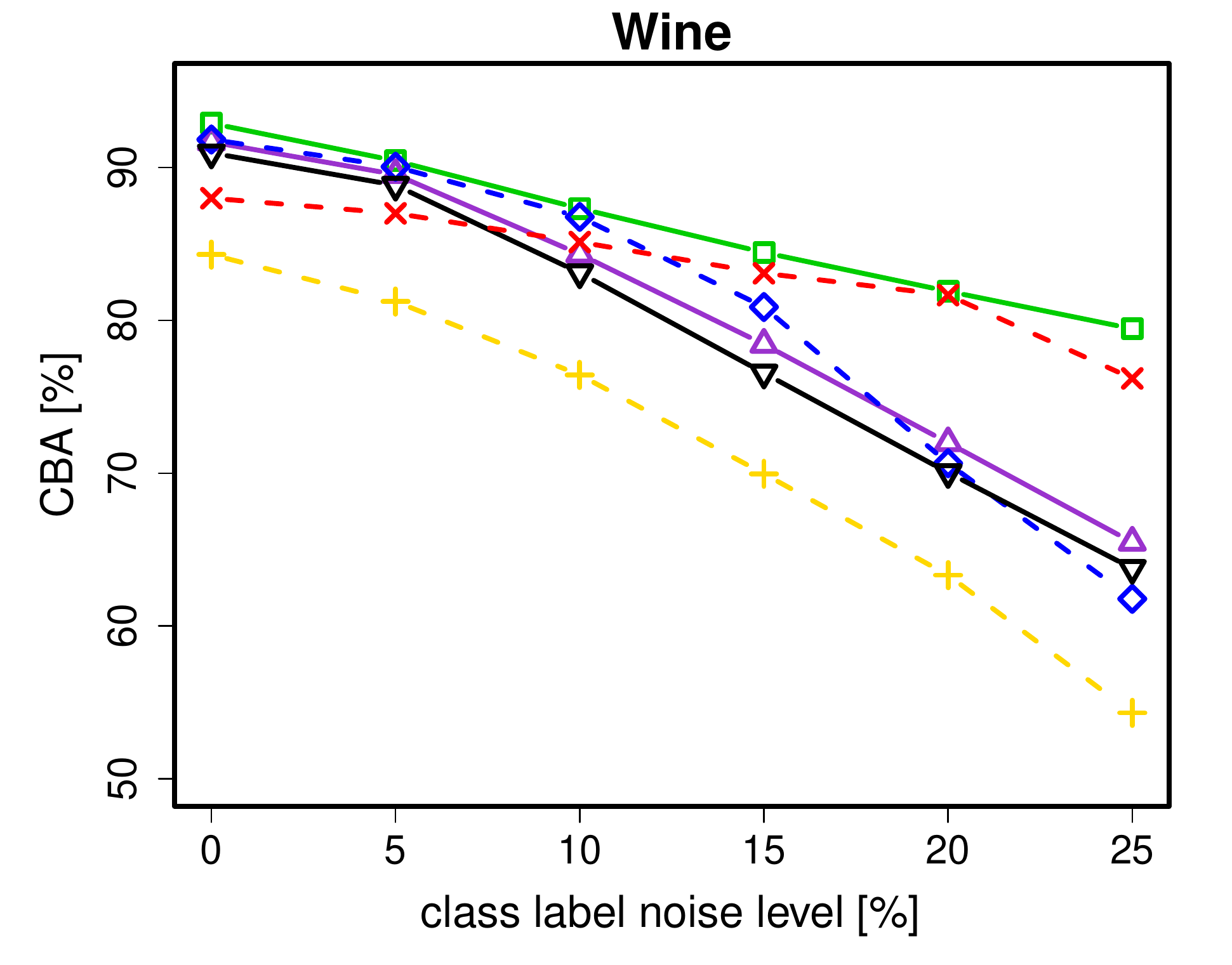}
        \includegraphics[width=0.24\textwidth]{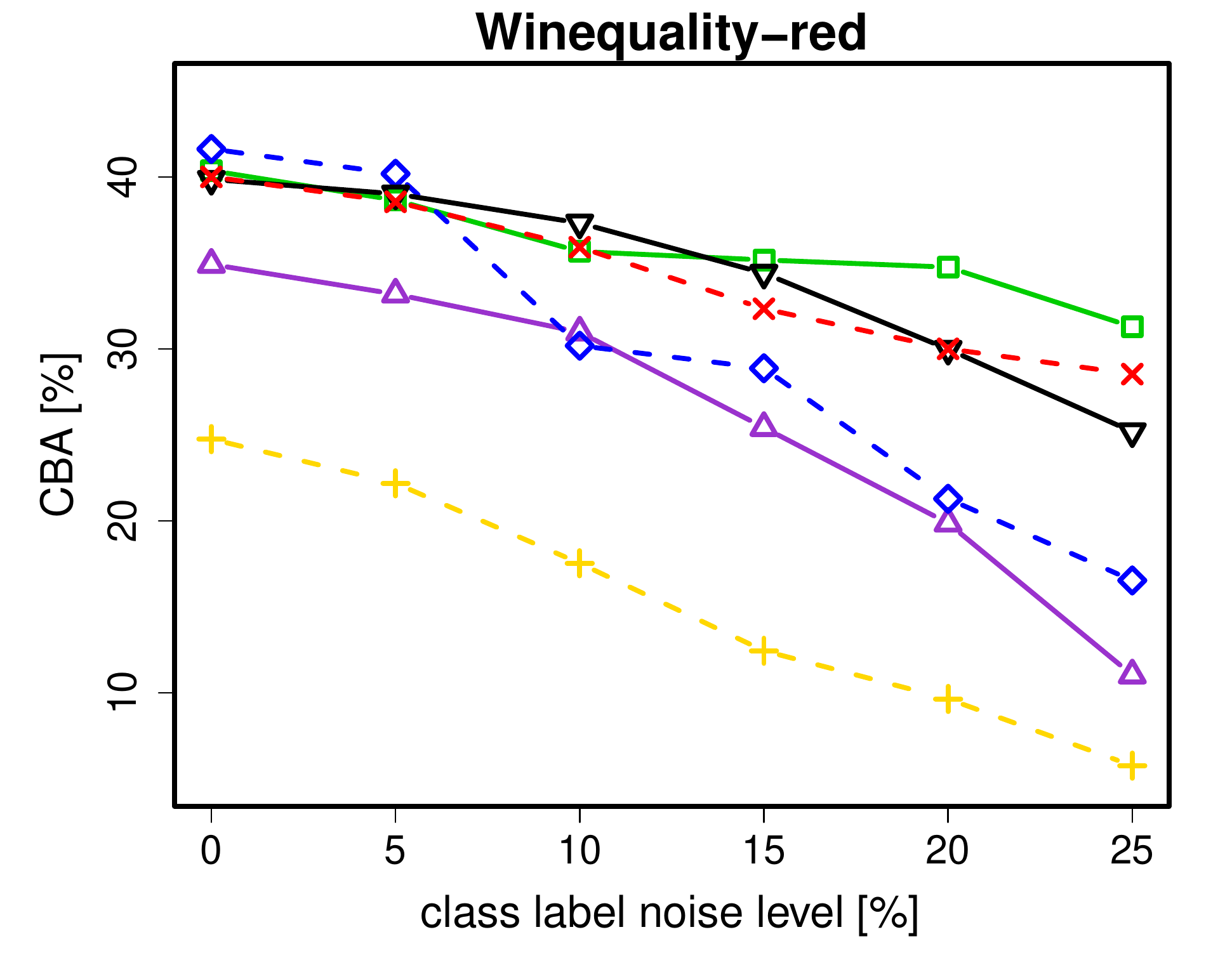}
        \includegraphics[width=0.24\textwidth]{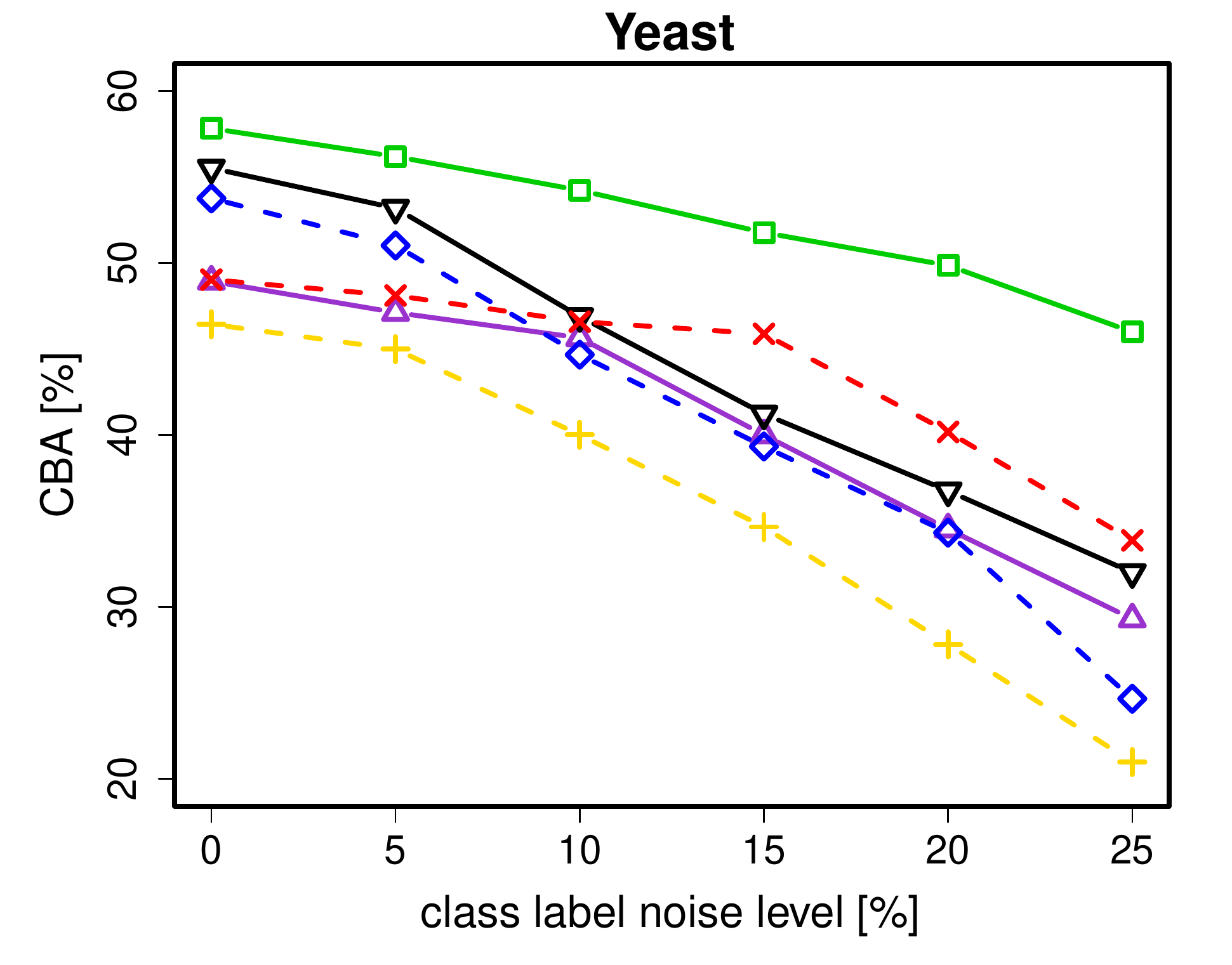}
        \includegraphics[width=0.24\textwidth]{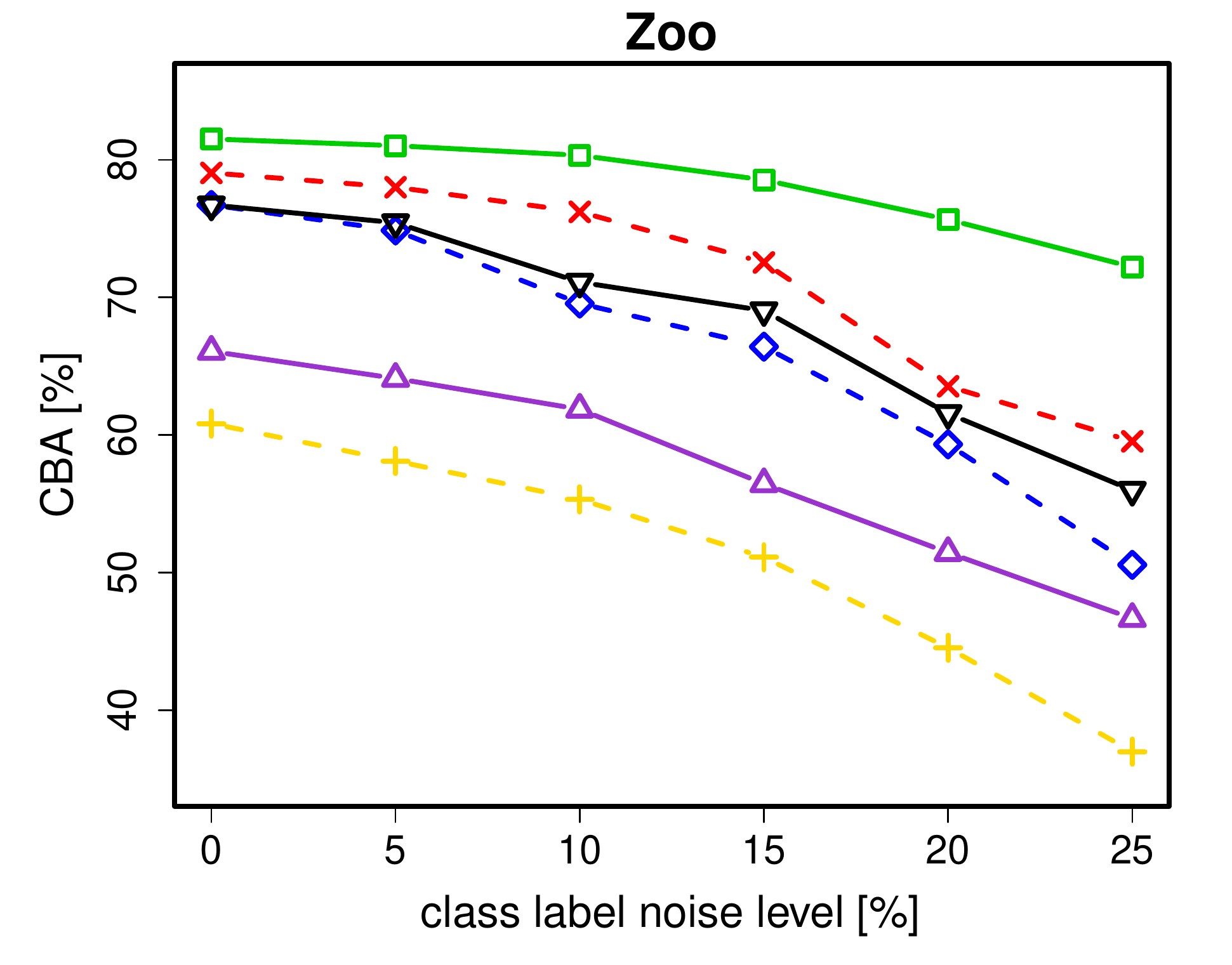}
\caption{Influence of varying class label noise levies on MC-CCR and reference sampling algorithms according to CBA [\%] and C5.0 as a base classifier.}
\label{fig:noise}
\end{figure*}
\begin{figure*}
\centering
    \begin{subfigure}[t]{0.99\textwidth}
    \captionsetup{font=scriptsize,labelfont=scriptsize}
        \centering        
        \includegraphics[width=0.24\textwidth]{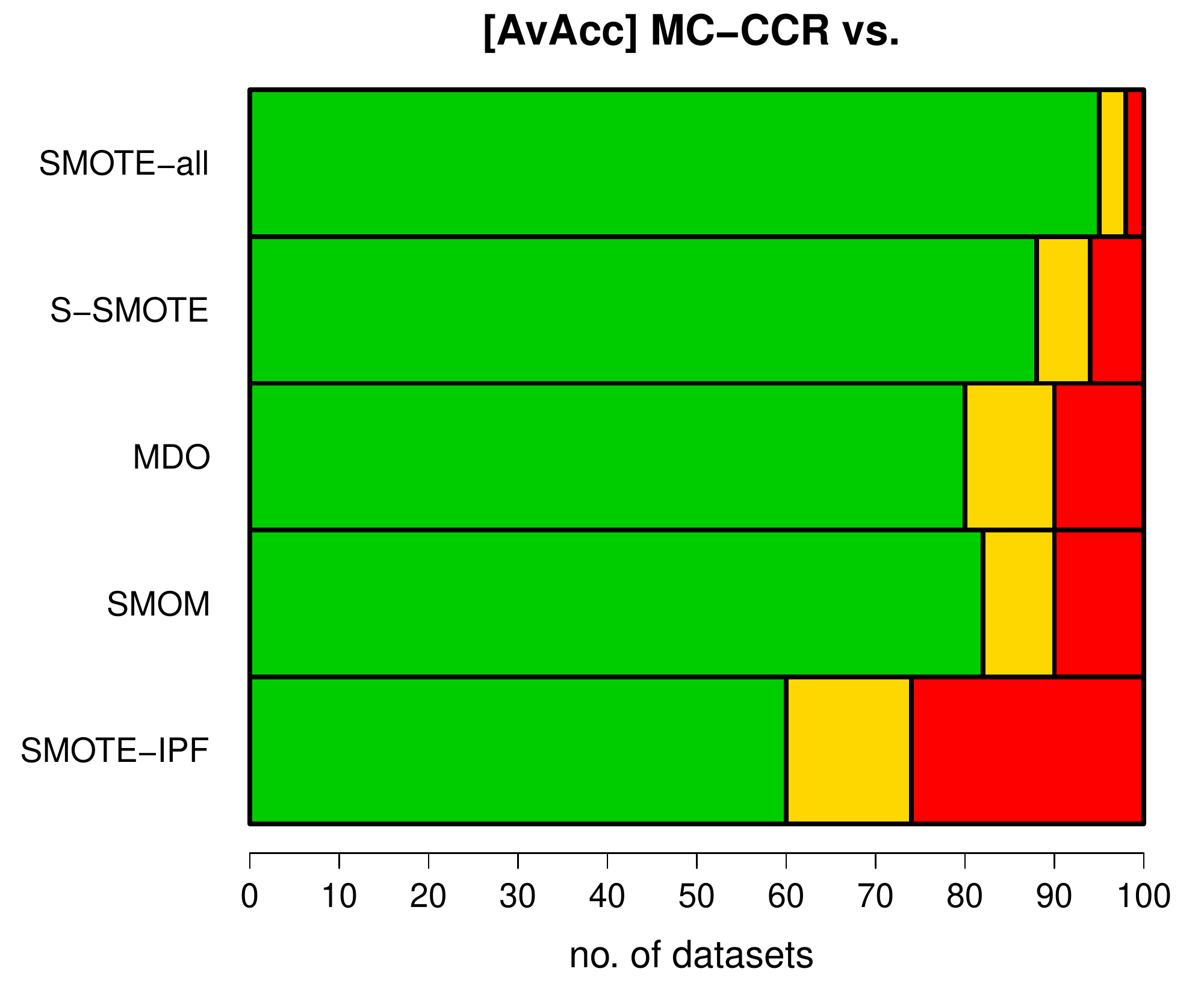}
        \includegraphics[width=0.24\textwidth]{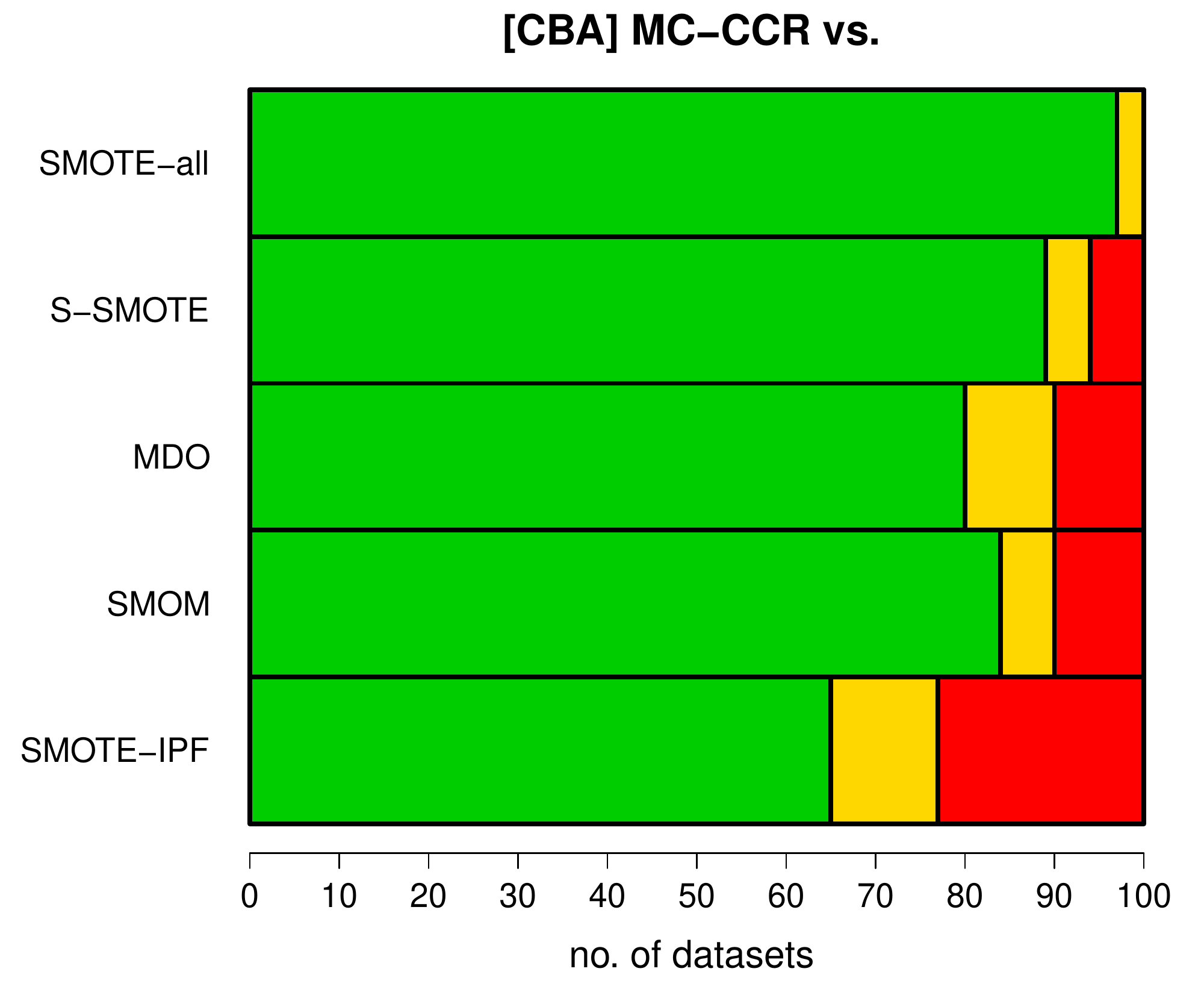}    
        \includegraphics[width=0.24\textwidth]{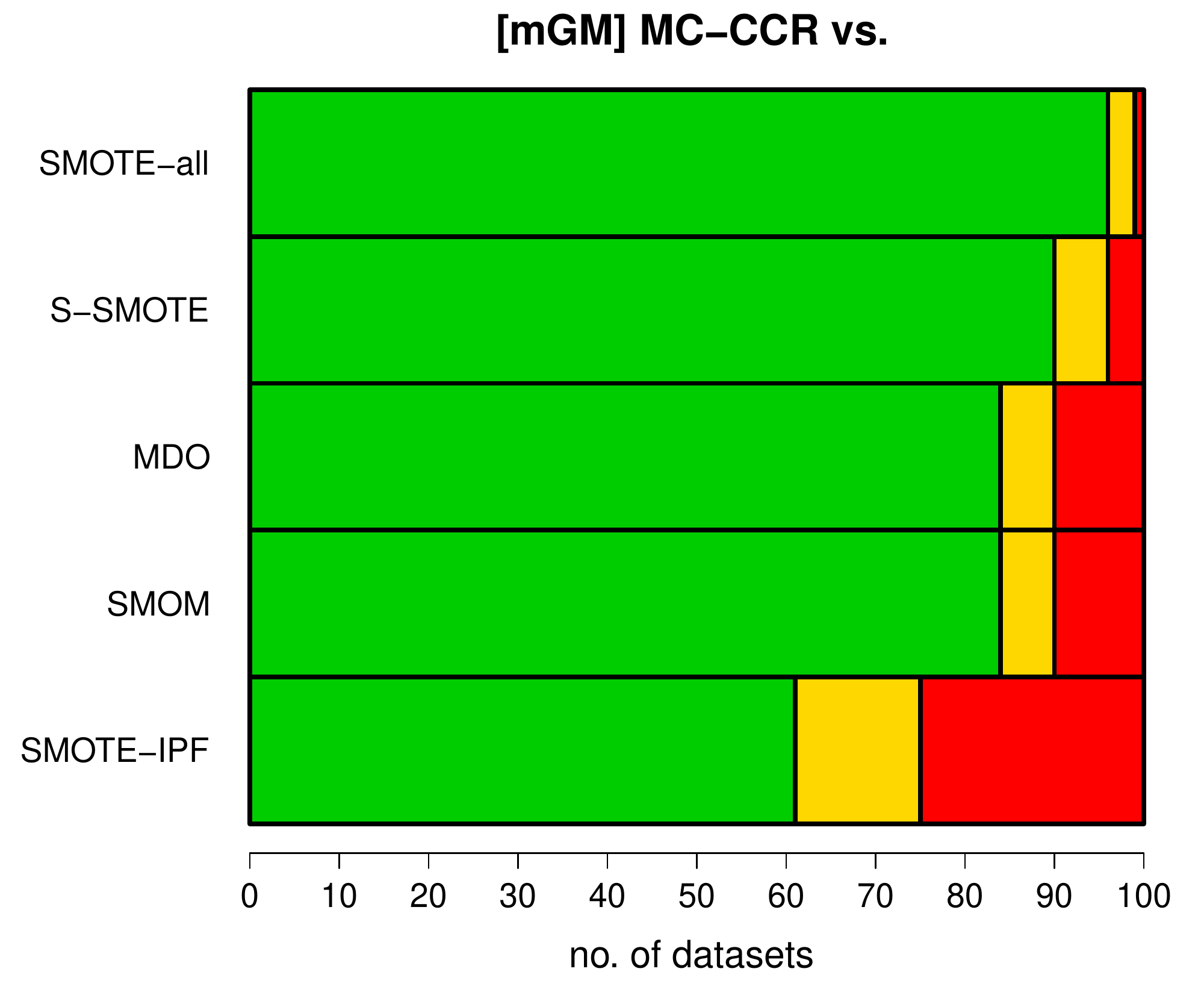}
        \includegraphics[width=0.24\textwidth]{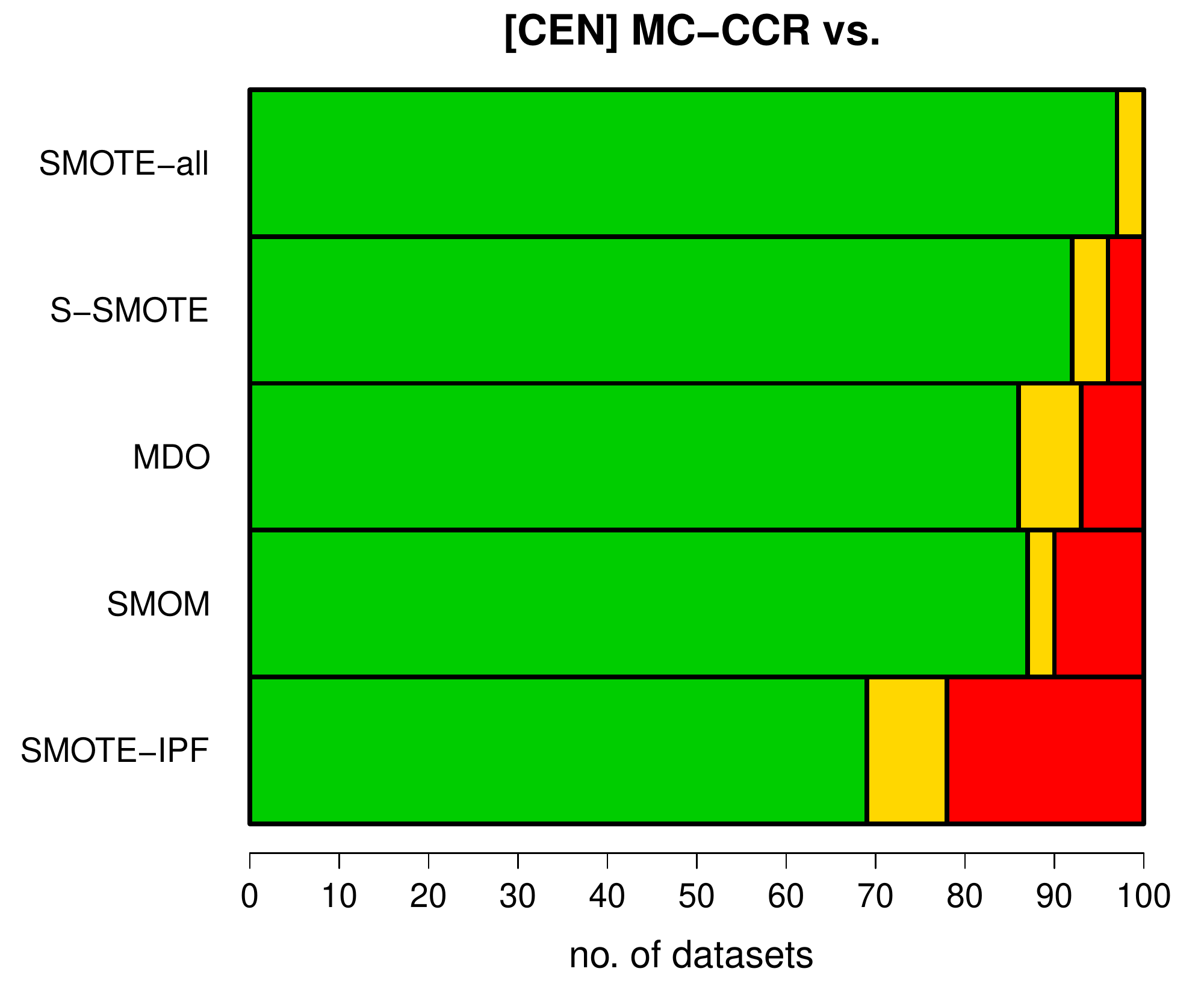}
        \caption{C5.0}
    \end{subfigure}
    ~ 
    \begin{subfigure}[t]{0.99\textwidth}
    \captionsetup{font=scriptsize,labelfont=scriptsize}
        \centering        
        \includegraphics[width=0.24\textwidth]{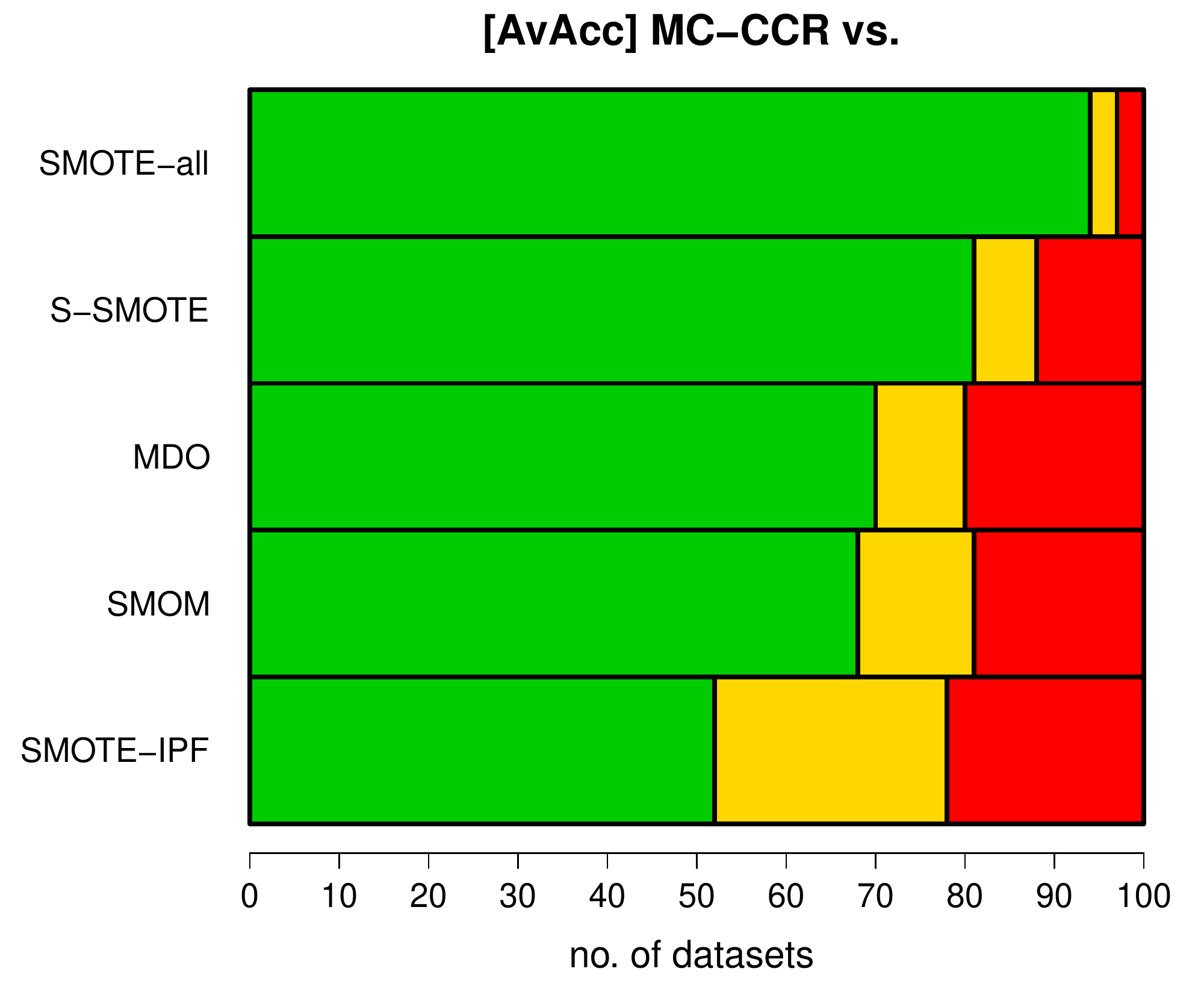}
        \includegraphics[width=0.24\textwidth]{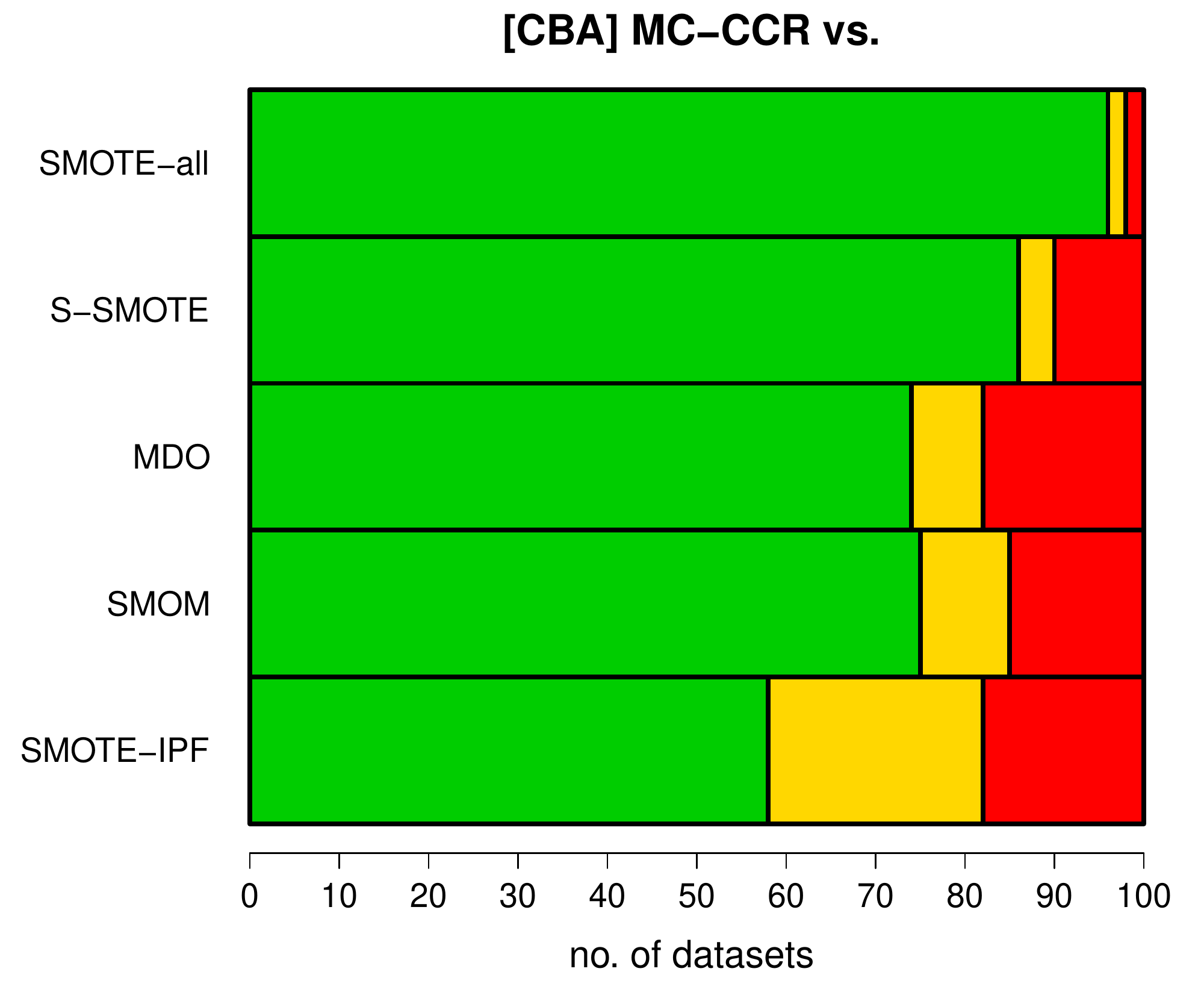}    
        \includegraphics[width=0.24\textwidth]{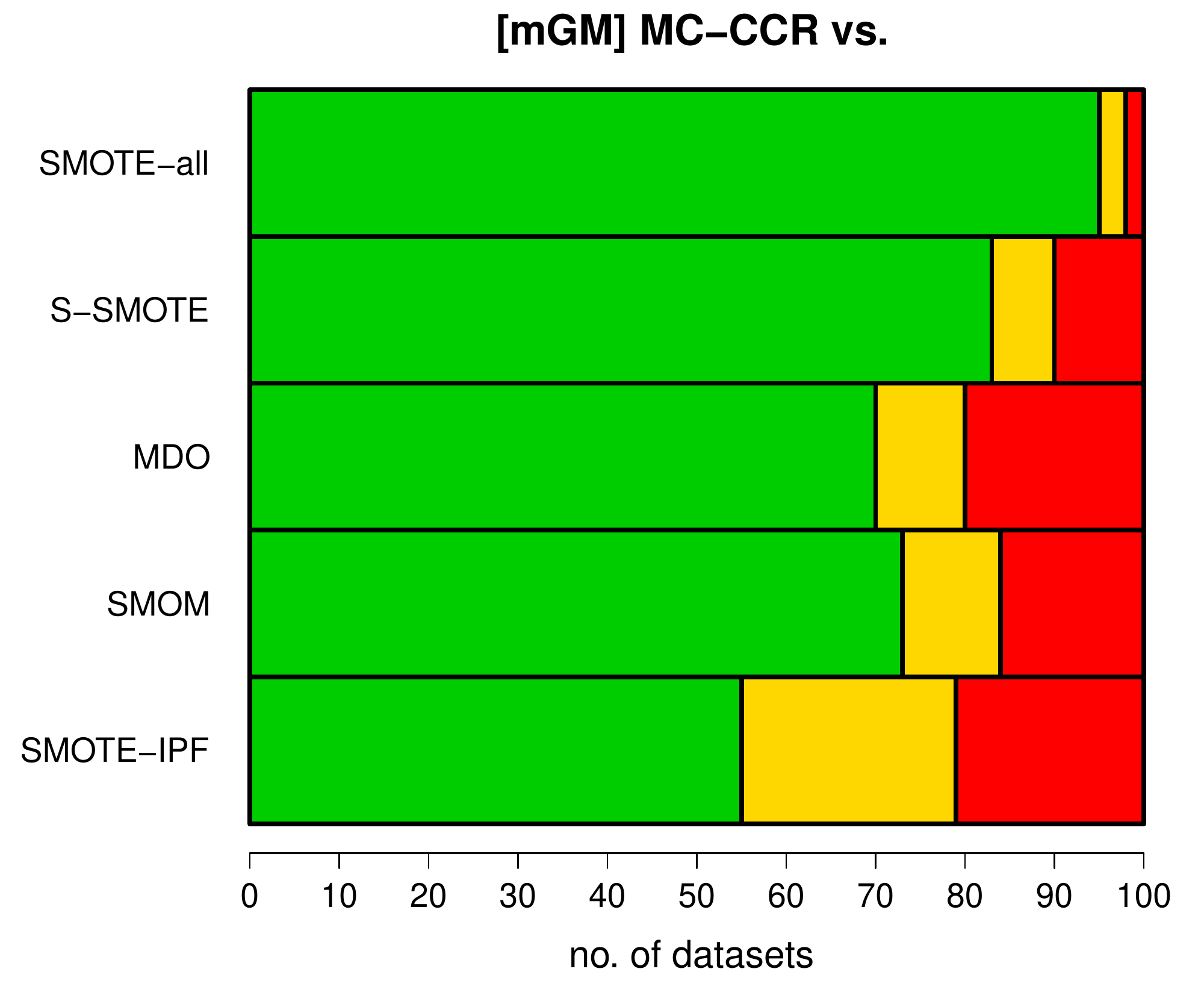}
        \includegraphics[width=0.24\textwidth]{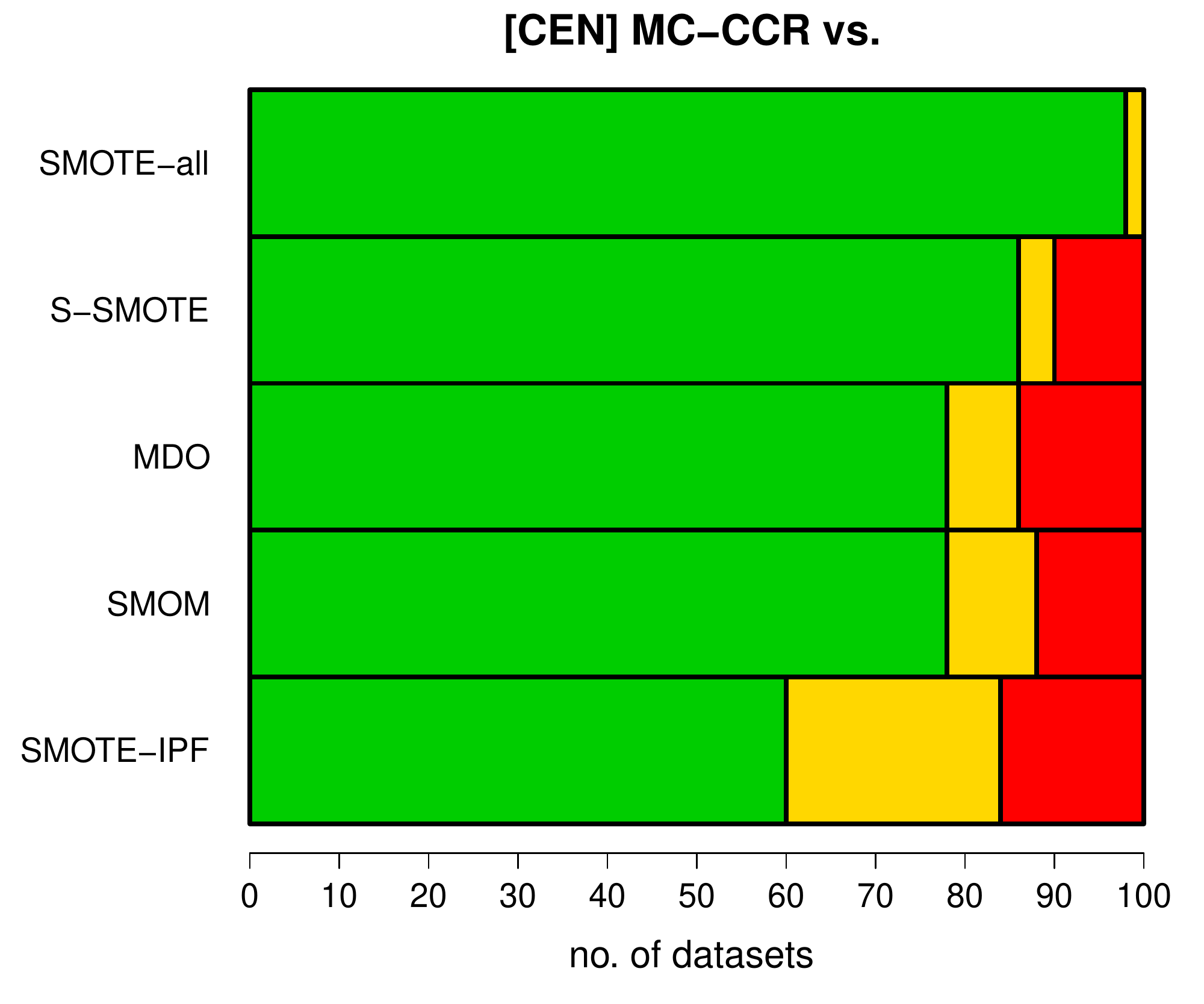}
        \caption{MLP}
    \end{subfigure}
        ~ 
    \begin{subfigure}[t]{0.99\textwidth}
    \captionsetup{font=scriptsize,labelfont=scriptsize}
        \centering        
        \includegraphics[width=0.24\textwidth]{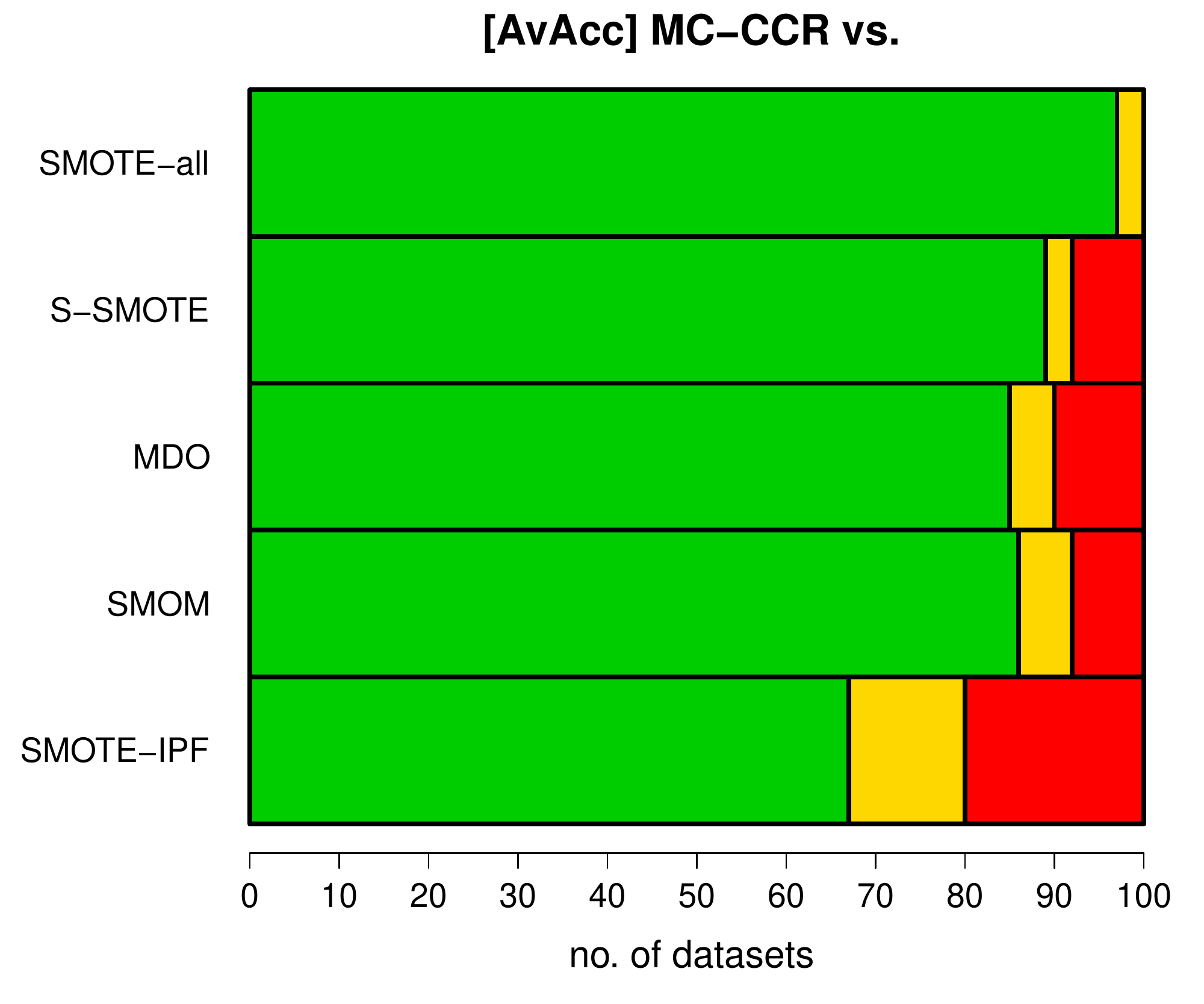}
        \includegraphics[width=0.24\textwidth]{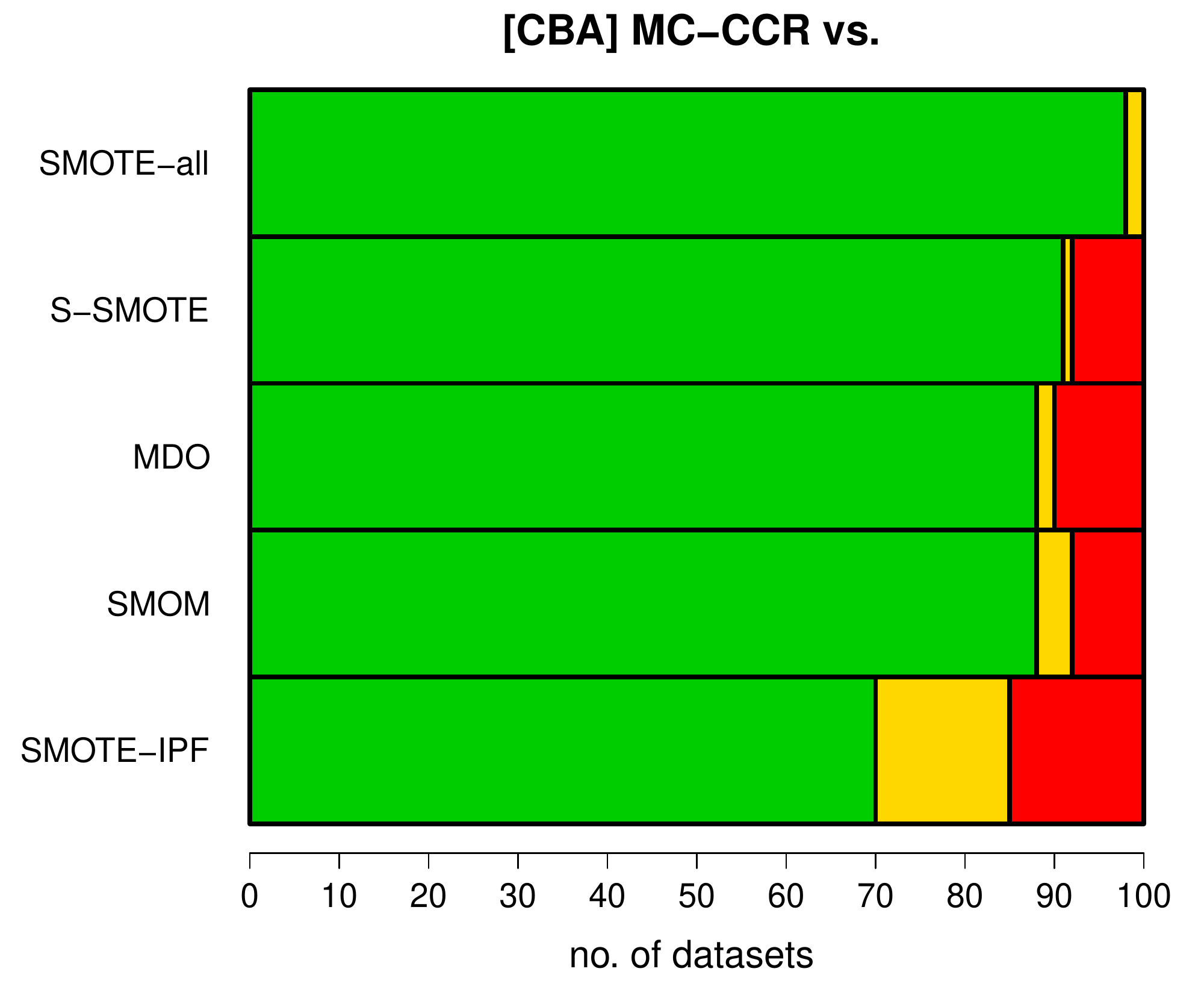}    
        \includegraphics[width=0.24\textwidth]{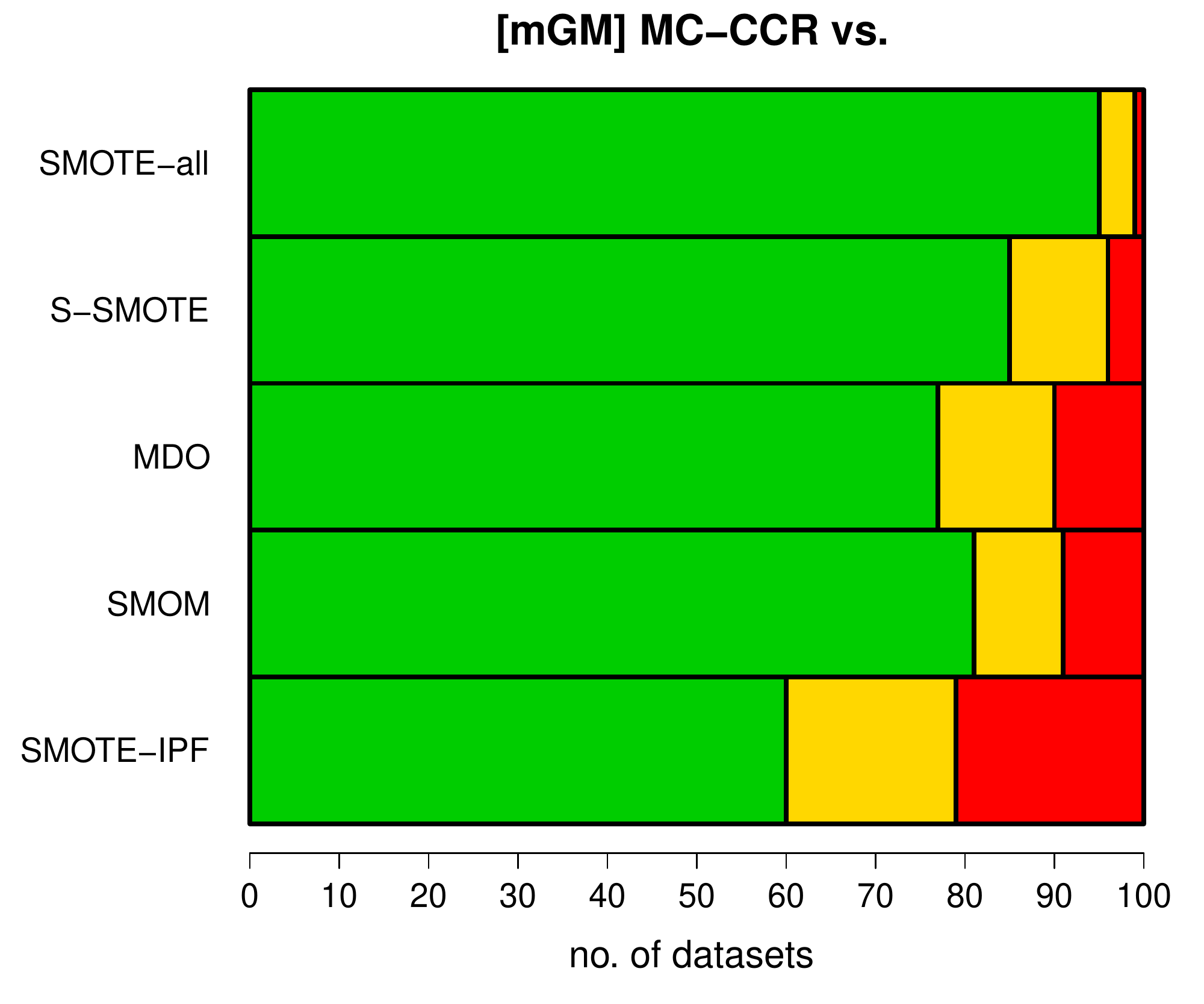}
        \includegraphics[width=0.24\textwidth]{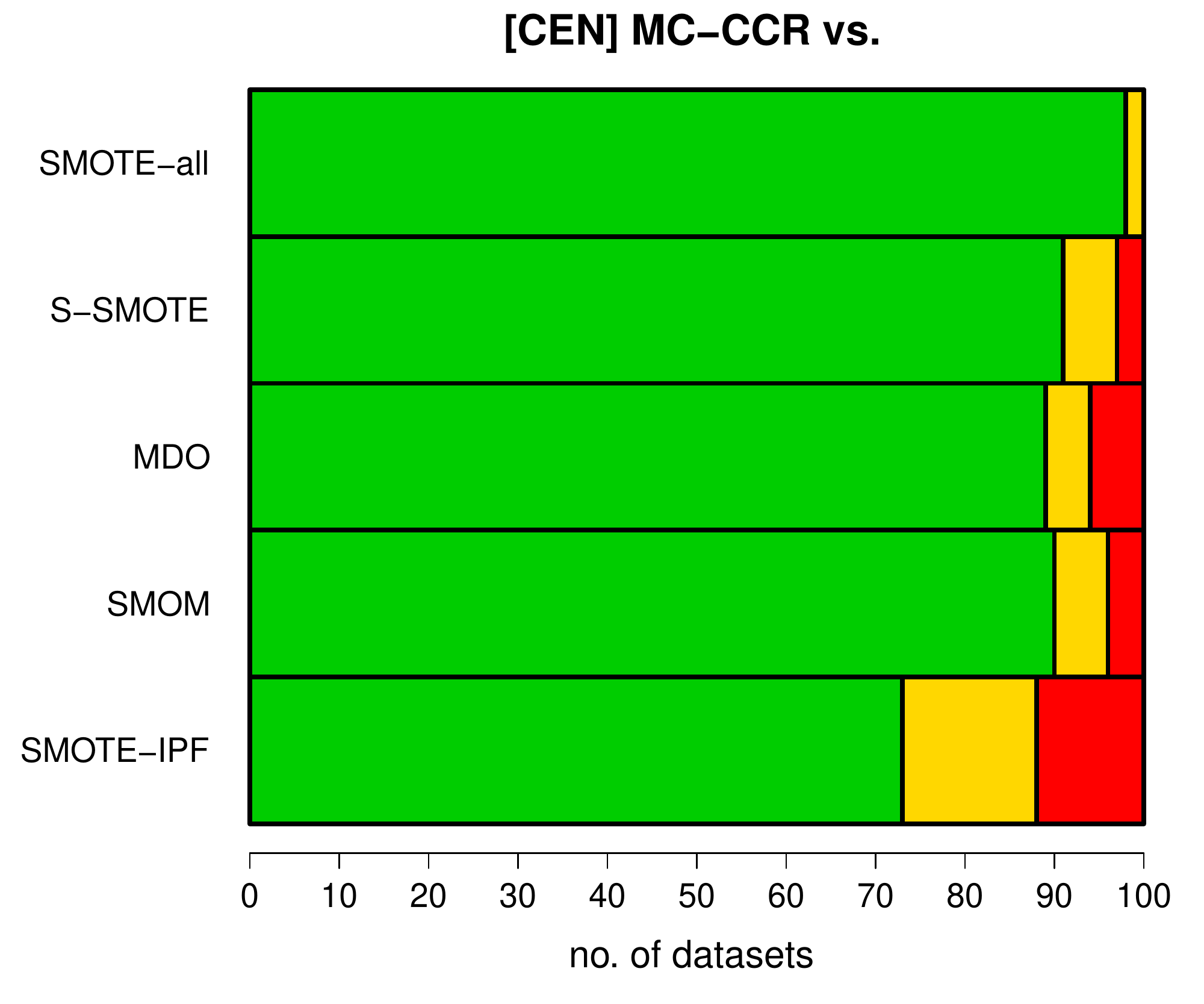}
        \caption{k-NN}
    \end{subfigure}
        ~ 
    \begin{subfigure}[t]{0.99\textwidth}
    \captionsetup{font=scriptsize,labelfont=scriptsize}
        \centering        
        \includegraphics[width=0.24\textwidth]{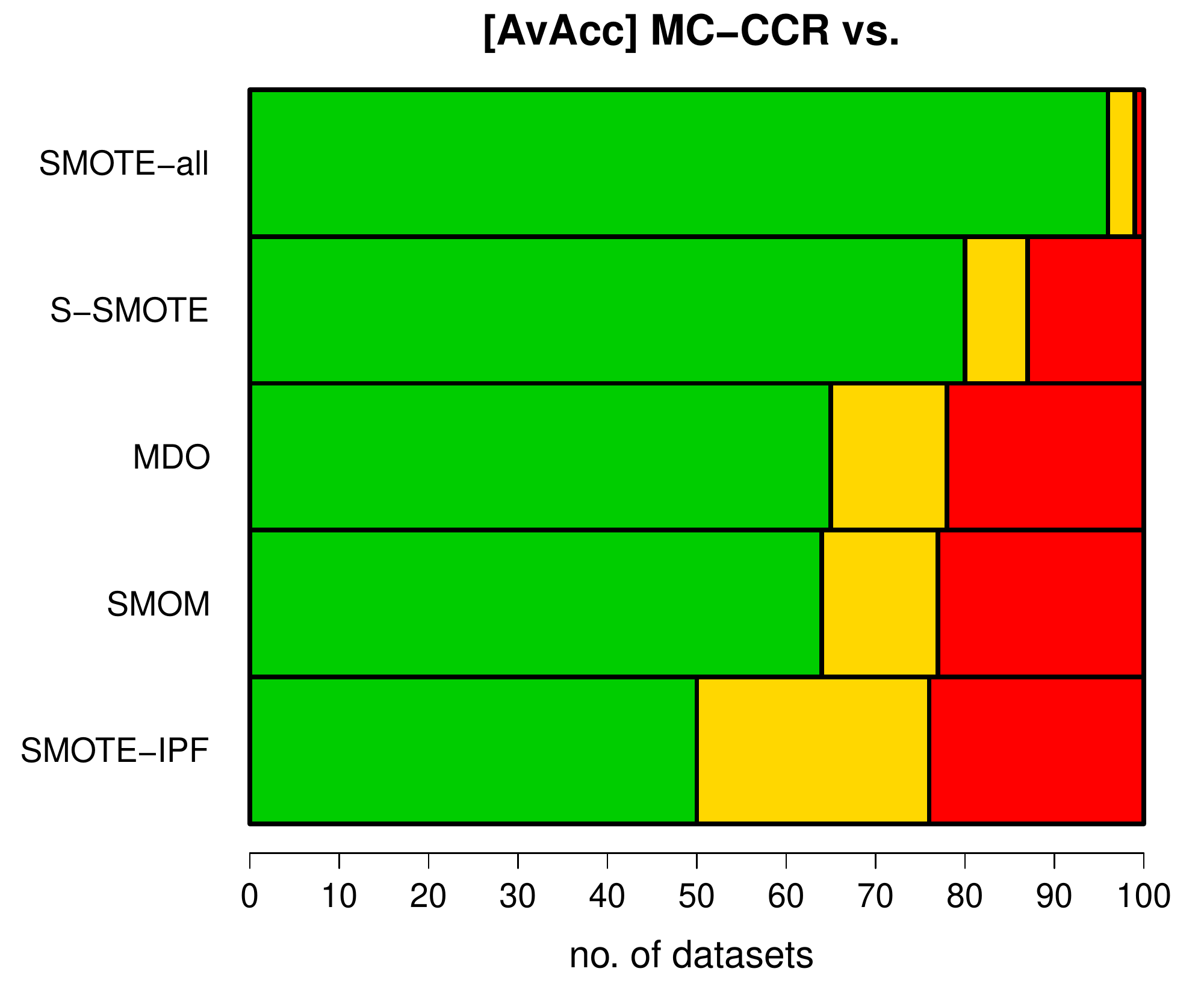}
        \includegraphics[width=0.24\textwidth]{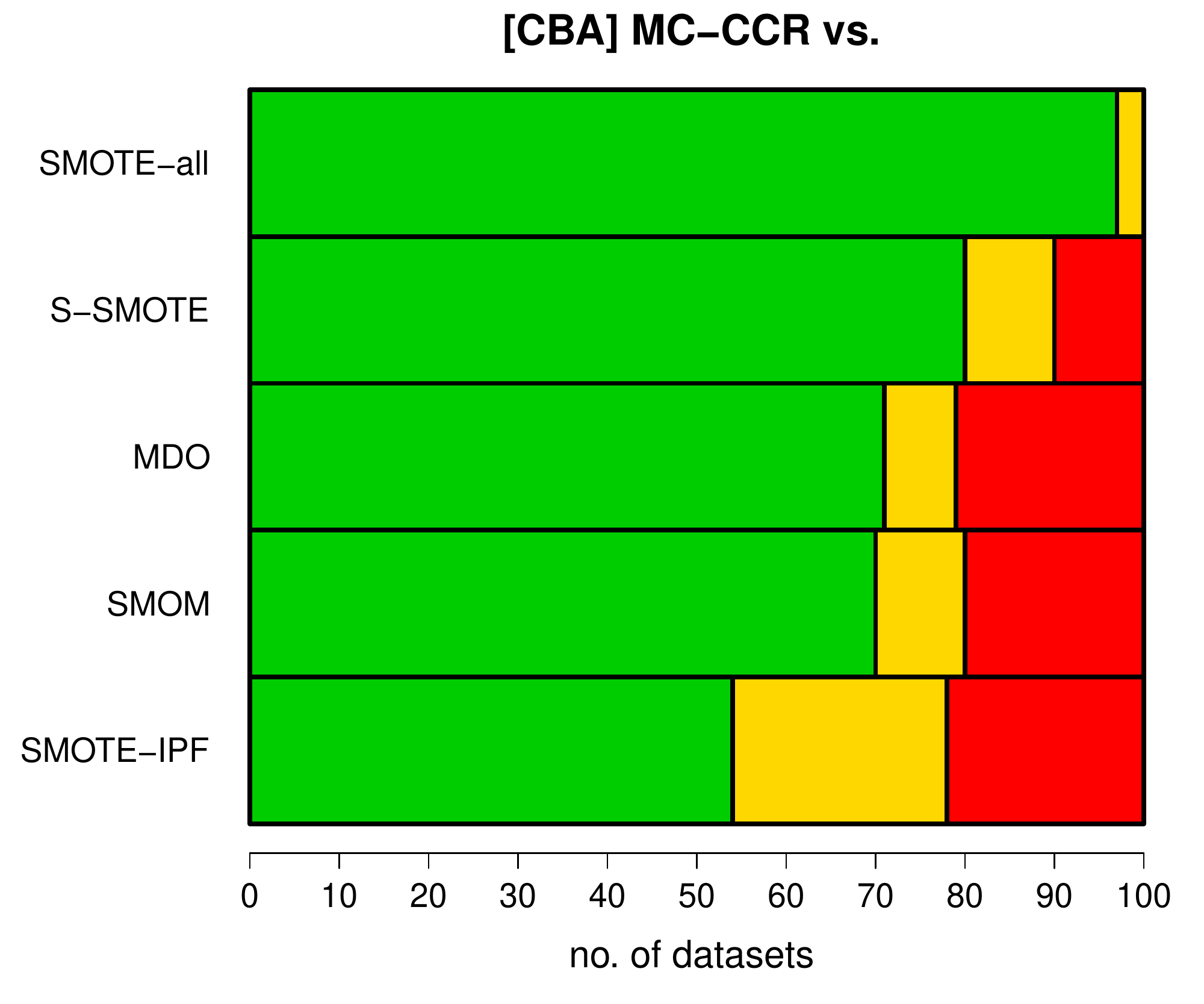}    
        \includegraphics[width=0.24\textwidth]{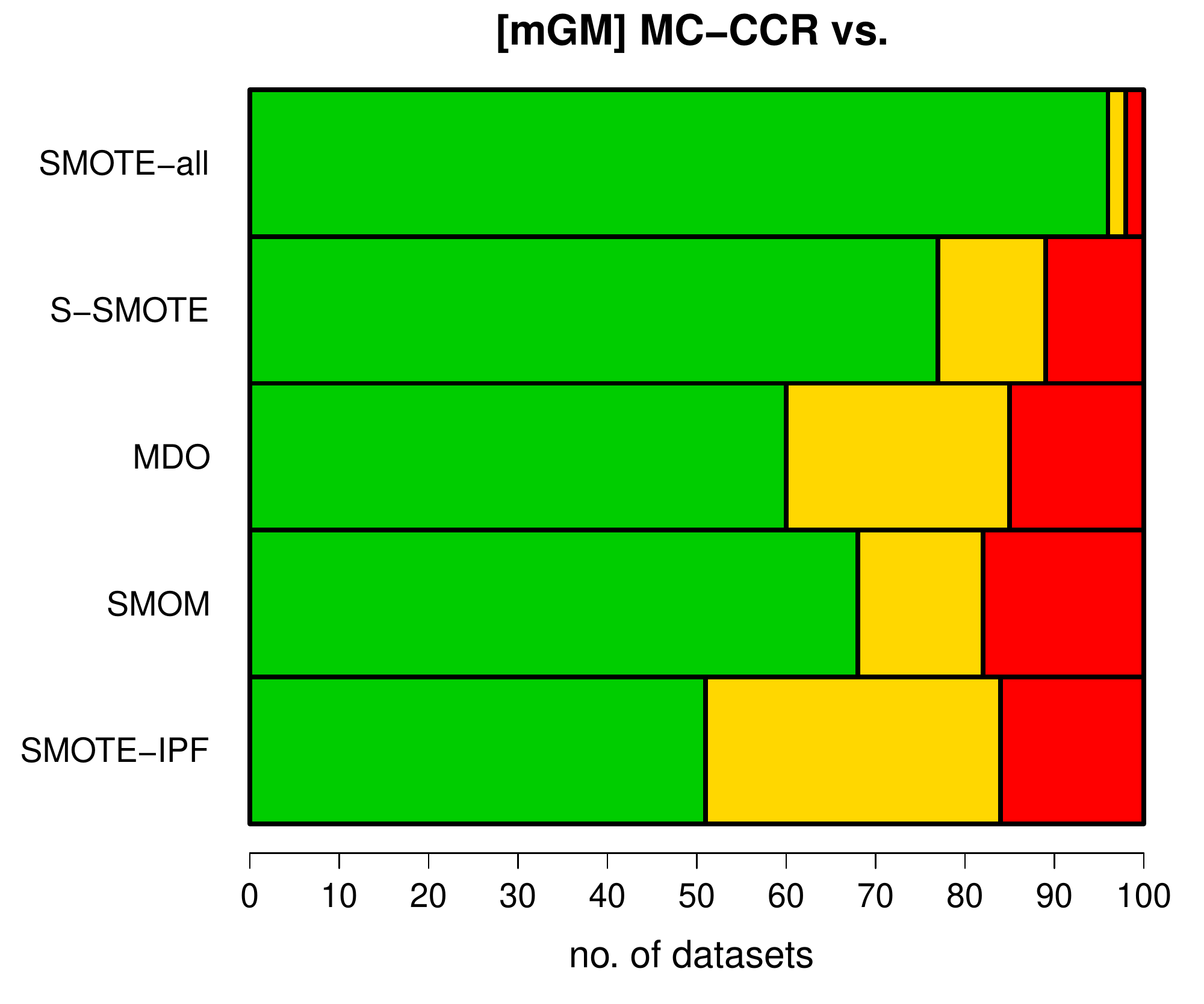}
        \includegraphics[width=0.24\textwidth]{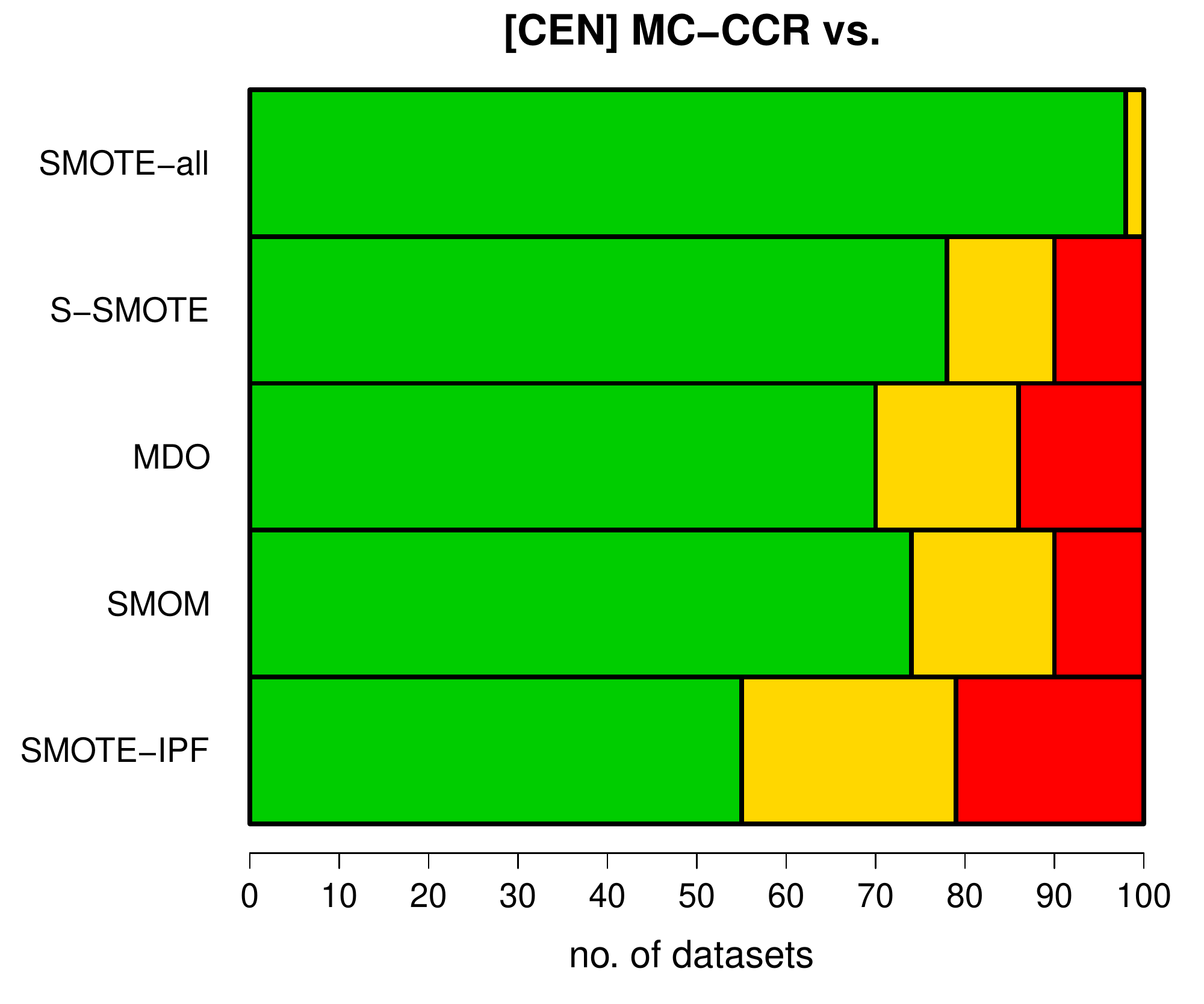}
        \caption{NB}
    \end{subfigure}
\caption{Comparison of MC-CCR with reference methods for four tested base classifiers with respect to the number of noisy datasets on which MC-CCR was statistically significantly better (green), similar (yellow), or worse (red) using combined 10-fold CV F-test over 100 datasets (20 benchmarks $\times$ 5 class label noise levels).}
\label{fig:wins-noise}
\end{figure*}
\begin{figure*}
\centering
        \includegraphics[width=0.24\textwidth]{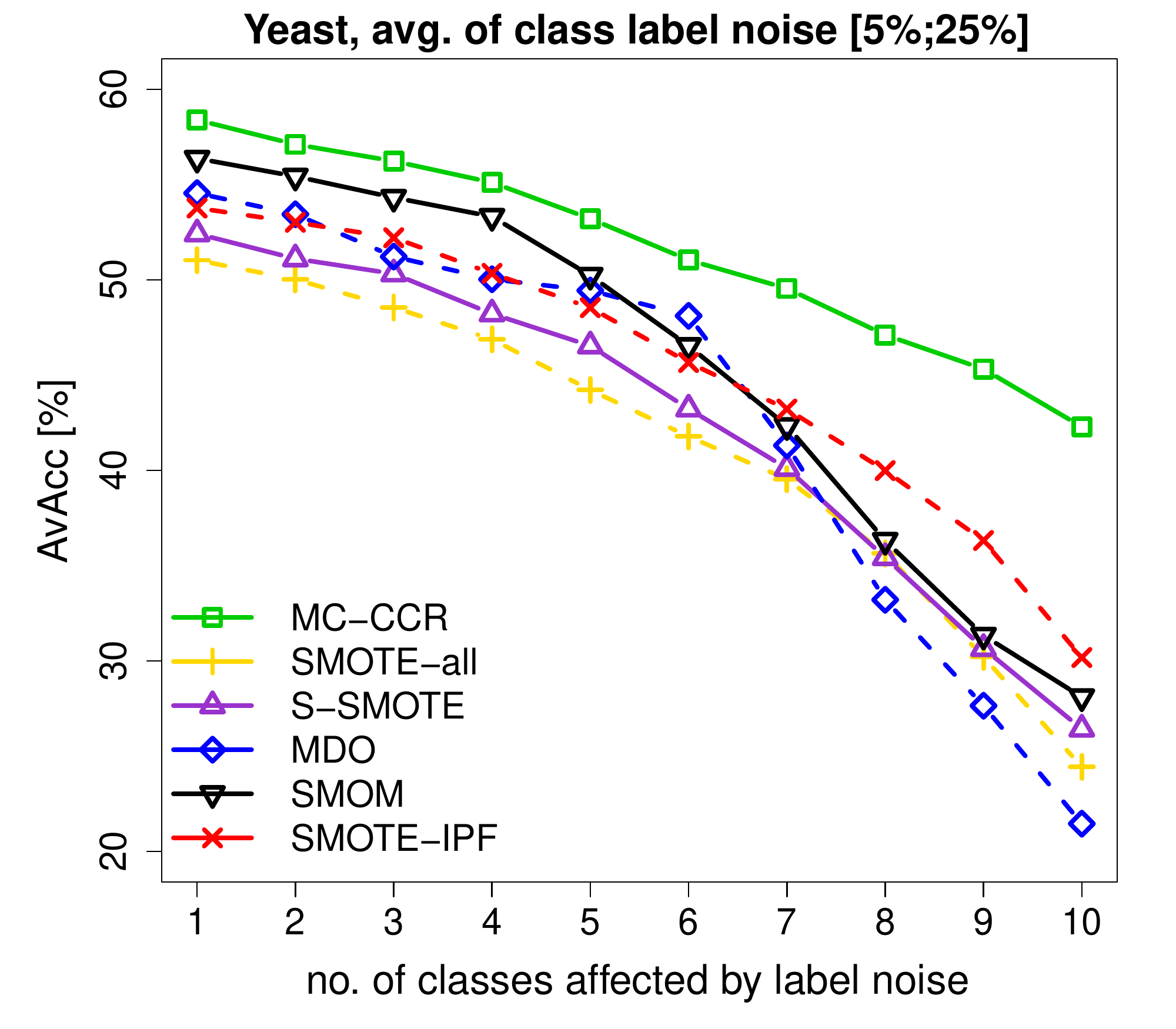}
        \includegraphics[width=0.24\textwidth]{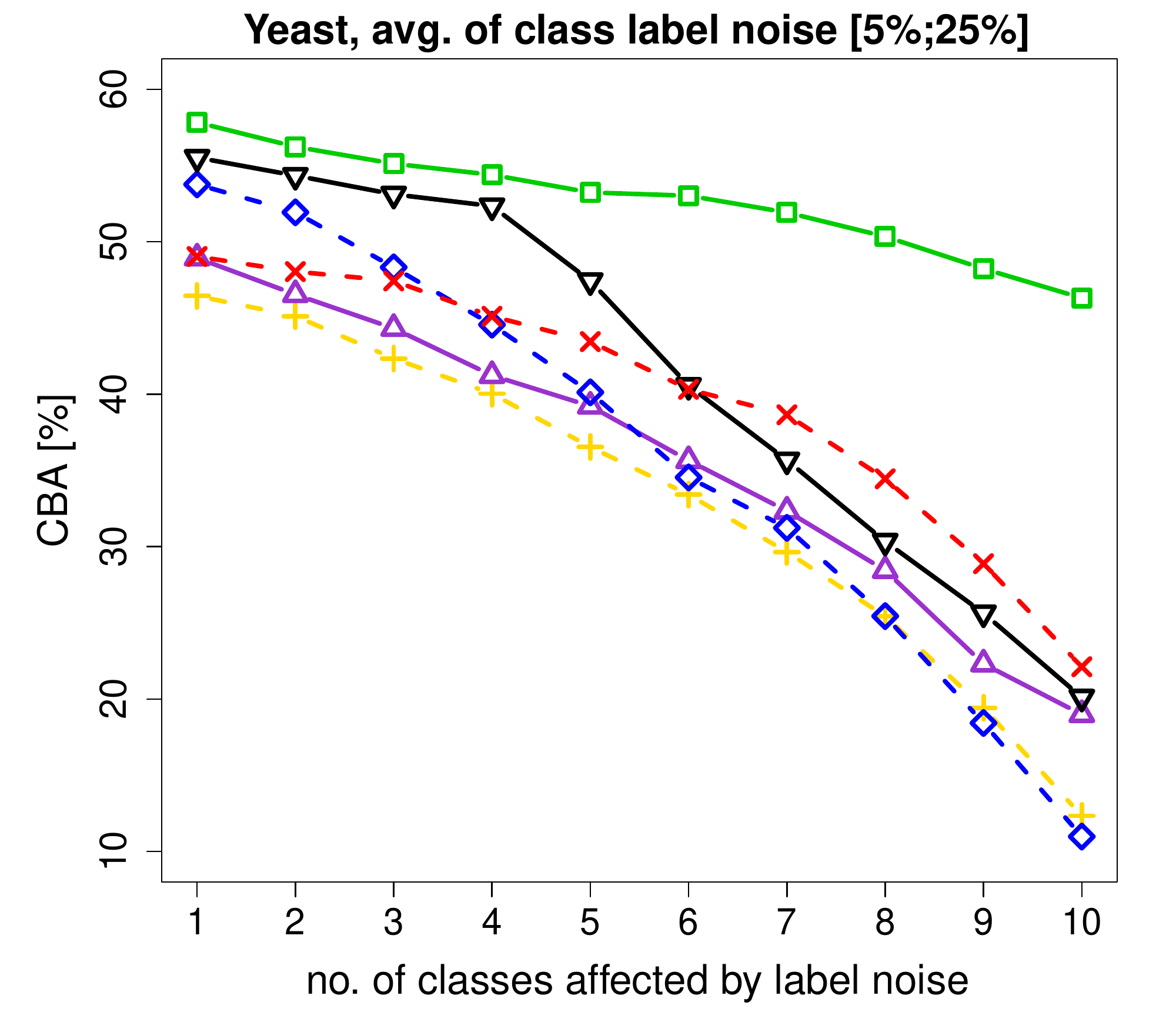}
        \includegraphics[width=0.24\textwidth]{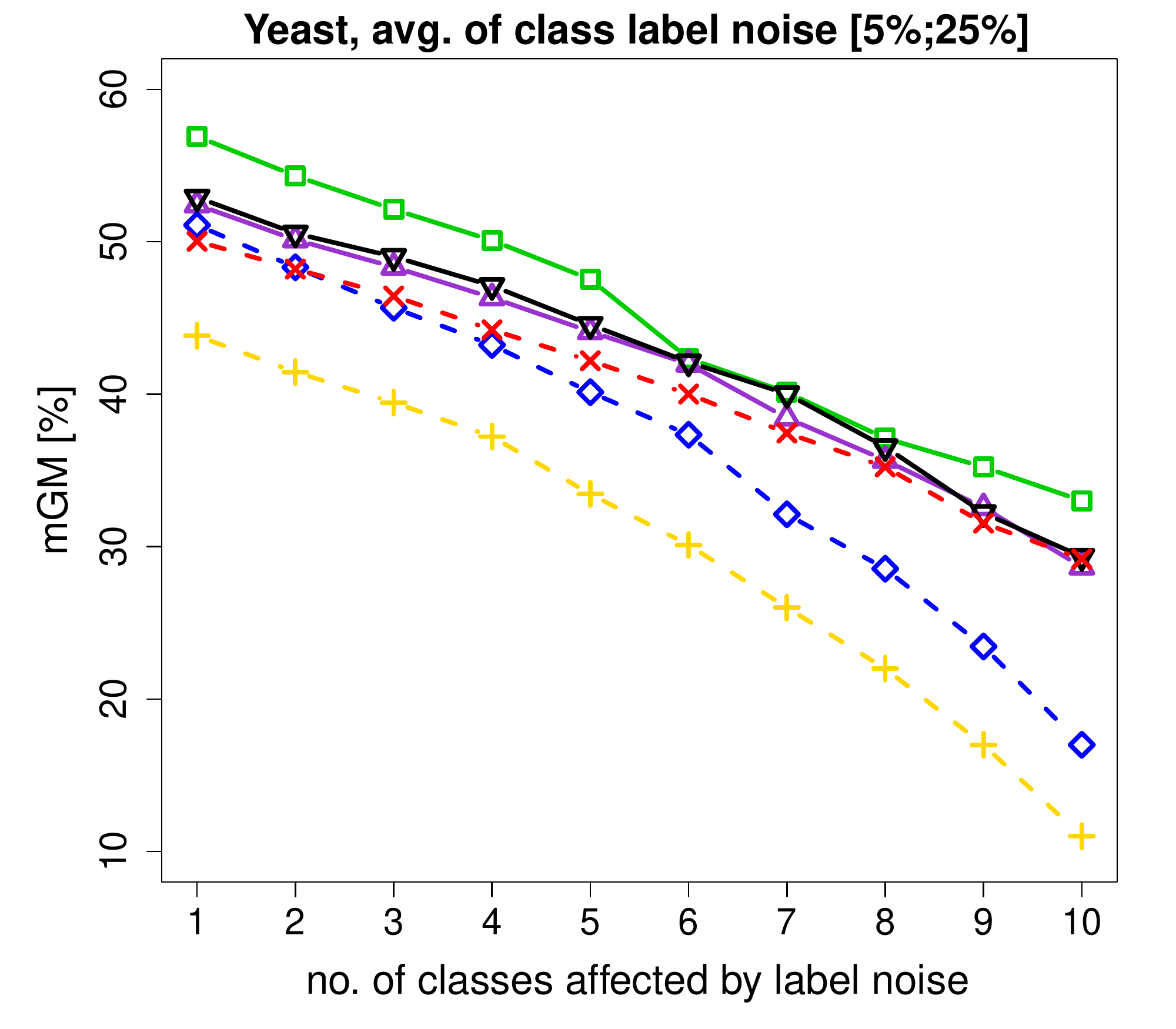}
        \includegraphics[width=0.24\textwidth]{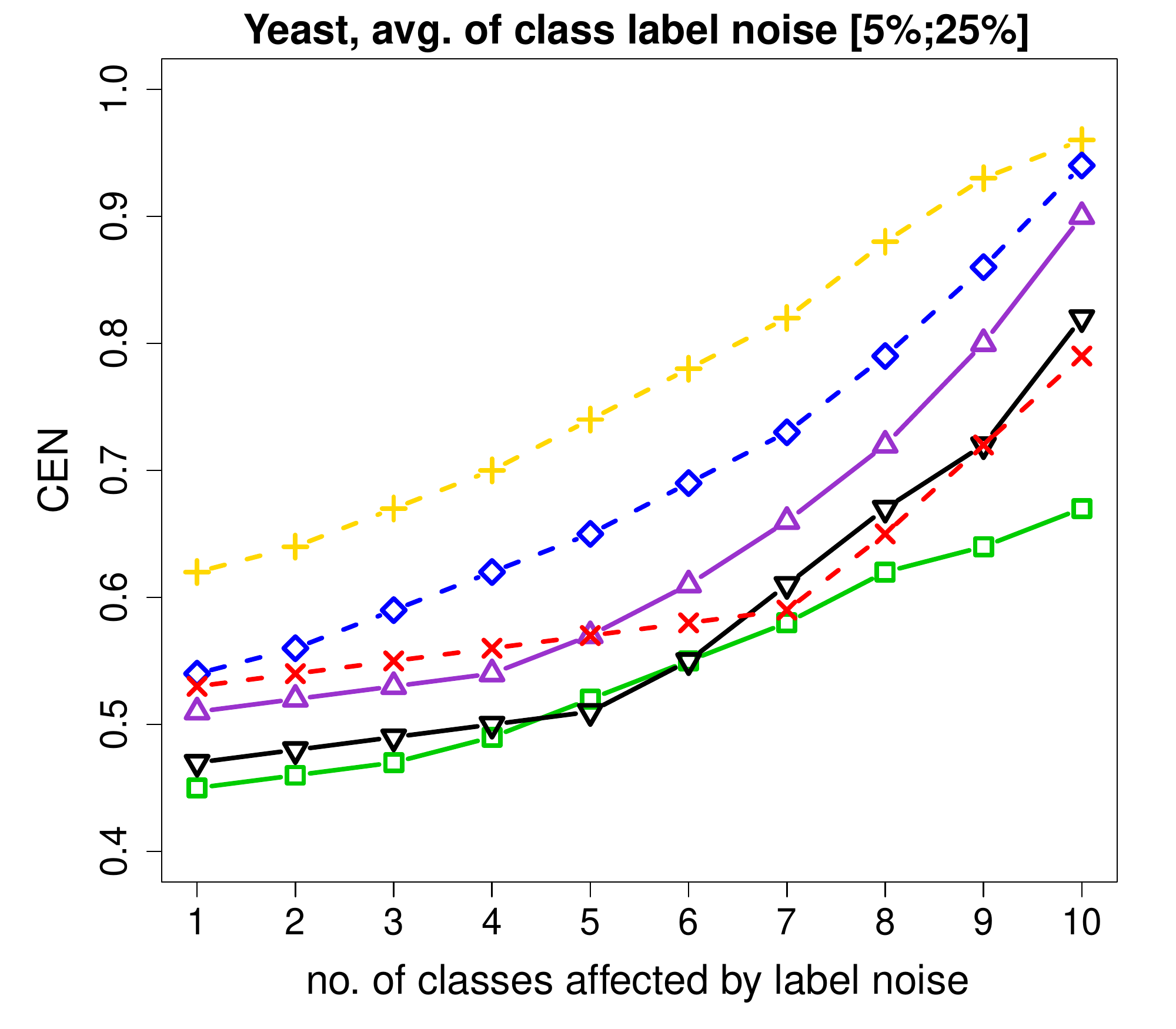}
\caption{Analysis of relationship between number of classes affected by label noise and performance metrics. Results presented for Yeast benchmark (10 classes), averaged over all noise levels (5\% - 25\%) with C5.0 as a base classifier.}
\label{fig:noise_av}
\end{figure*}

The results of the experiments were presented in Figures~\ref{fig:noise}--\ref{fig:noise_av} as well as in Table~\ref{tab:sh2}. When analyzing the relationship between the noise level and the predictive performance for different methods for different overampling methods, it should be noted that for most datasets one can notice the obvious tendency that quality deteriorates with the increase in noise level. MC-CCR usually has better predictive performance compared to state-of-art methods. It is also worth analyzing how quality degradation occurs as noise levels increase. Most benchmark algorithms report a sharp drop in quality after exceeding the label noise level of 10-15\% (except for SMOTE-IPF, which in many cases has fairly stable quality). However, MC-CCR, although the degradation of predictive performance depending on the noise level is noticeable, it is not so violent. It is linear for the whole range of experiments.

Similar MC-CCR behavior can also be seen when analyzing the relationship between the number of classes affected by noise and predictive performance. The decrease in the value of all the metrics is close to linear. In contrast, in the case of the remaining oversampling algorithms, we can observe a sharp deterioration in quality at the noise of a small number of classes, and the characteristics are close to quadratic.

When analyzing MC-CCR concerning various classifiers, it should be stated that for most databases and noise levels, the proposed method is characterized by much better predictive performance and, as a rule, is statistically significantly better than state-of-art algorithms. MC-CCR is best suited for use with minimum distance classifiers (as $k$-NN) and also with decision trees, although for other tested classification algorithms it also achieves very good results. Generalizing the observed predictive performance, MC-CCR is very robust to the label noise and it is characterized by the smallest decrease in predictive performance depending on the label noise level, or the number of classes affected by the noise. Due to this property, it can be seen that the proposed method is always statistically significantly better than other tested algorithms, especially for high noise levels. The benchmark methods may be ranked according to these criteria in the following order: SMOTE-IPF, SMON, MDO, S-SMOTE, and SMOTE-all.

\begin{table*}[!ht]
\small
 \centering
 \caption{Shaffer's tests for comparison between MC-CCR and reference oversampling methods on two levels of class noise (smallest and largest noise ratios) with respect to each metric and base classifier. We report obtained $p$-values. Symbol '$>$' stands for situation in which MC-CCR is statistically superior and '$=$' for when there are no significant differences.}
 \label{tab:sh2}
 \scalebox{0.88}{
 \begin{tabular}{lcccccccc}
 \toprule
 \tabhead{Hypothesis} & \multicolumn{2}{c}{\tabhead{AvACC}}  &  \multicolumn{2}{c}{\tabhead{CBA}}  &  \multicolumn{2}{c}{\tabhead{mGM}}  &  \multicolumn{2}{c}{\tabhead{CEN}} \\
\midrule
& 5\% noise & 25\% noise & 5\% noise & 25\% noise & 5\% noise & 25\% noise & 5\% noise & 25\% noise\\
\midrule
C5.0 & & & & \\
vs. SMOTE-all & $>$ (0.0000) & $>$ (0.0000) & $>$ (0.0003) & $>$ (0.0000) & $>$ (0.0000) &  $>$ (0.0000) & $>$ (0.0005) & $>$ (0.0000)\\
vs. S-SMOTE & $>$ (0.0311) & $>$ (0.0157) & $>$ (0.0096) & $>$ (0.0021)& $>$ (0.0156) & $>$ (0.0078) & $>$ (0.0099) & $>$ (0.0019)\\
vs. MDO & $>$ (0.0382) & $>$ (0.0179) & $>$ (0.0356) & $>$ (0.0199) & $>$ (0.0466) & $>$ (0.0203)& $>$ (0.0408)& $>$ (0.0275) \\
vs. SMOM & $>$ (0.0438) &$>$ (0.0198)& $>$ (0.0401) & $>$ (0.0158)&$>$ (0.0384) & $>$ (0.0127)& $>$ (0.0376) & $>$ (0.0122)\\
vs. SMOTE-IPF & $>$ (0.0285) & $>$ (0.0211)& $>$ (0.0151) & $>$ (0.0116)& $>$ (0.0159) & $>$ (0.0082)& $>$ (0.0109)& $>$ (0.0082)\\
\midrule
MLP & & & & \\
vs. SMOTE-all & $>$ (0.0011) & $>$ (0.0000) & $>$ (0.0038) & $>$ (0.0002)& $>$ (0.0035) & $>$ (0.0001)& $>$ (0.0021)& $>$ (0.0002)\\
vs. S-SMOTE & $>$ (0.0246) & $>$ (0.0138)& $>$ (0.0238) & $>$ (0.0111)& $>$ (0.0289) & $>$ (0.0162)& $>$ (0.0117)& $>$ (0.0076)\\
vs. MDO & $=$ (0.0977) & $>$ (0.0436) & $=$ (0.0698) & $>$ (0.0381)& $=$ (0.0933) & $>$ (0.0420) & $=$ (0.0515)& $>$ (0.0359)\\
vs. SMOM & $=$ (0.1286) & $>$ (0.0482)& $=$ (0.0964) & $>$ (0.0458)& $=$ (0.1003) & $>$ (0.0406)& $=$ (0.0602)& $>$ (0.0396)\\
vs. SMOTE-IPF & $>$ (0.0298) &$>$ (0.0209)& $>$ (0.0202) & $>$ (0.0178)& $>$ (0.0263) & $>$ (0.0207)& $>$ (0.0170)& $>$ (0.0099)\\
\midrule
k-NN & & & & \\
vs. SMOTE-all & $>$ (0.0000) & $>$ (0.0000) & $>$ (0.005) & $>$ (0.0000) & $>$ (0.009) & $>$ (0.0000)& $>$ (0.0004)& $>$ (0.0000)\\
vs. S-SMOTE & $>$ (0.0347) & $>$ (0.0236)& $>$ (0.0098) & $>$ (0.0056)& $>$ (0.0322) & $>$ (0.0217)& $>$ (0.0102)& $>$ (0.0068)\\
vs. MDO & $>$ (0.0349) & $>$ (0.0139)& $>$ (0.0291) & $>$ (0.0102)& $>$ (0.0127) & $>$ (0.0073) &$>$ (0.0351)& $>$ (0.0249)\\
vs. SMOM & $>$ (0.0402) & $>$ (0.0255) & $>$ (0.0300) & $>$ (0.0104)& $>$ (0.0297) & $>$ (0.0100)&$>$ (0.0256)& $>$ (0.0099)\\
vs. SMOTE-IPF & $>$ (0.0115) &$>$ (0.0096)& $>$ (0.0096) &$>$ (0.0072)& $>$ (0.0122) & $>$ (0.0091)&$>$ (0.0078)& $>$ (0.0055)\\
\midrule
NB & & & & \\
vs. SMOTE-all & $>$ (0.0000) & $>$ (0.0000) & $>$ (0.0000) & $>$ (0.0000) & $>$ (0.0001) & $>$ (0.0000)& $>$ (0.0000) & $>$ (0.0000)\\
vs. S-SMOTE & $>$ (0.0135) & $>$ (0.0094) & $>$ (0.0087) & $>$ (0.0042)& $>$ (0.0103) & $>$ (0.0052) & $>$ (0.0072) &$>$ (0.0044)\\
vs. MDO & $=$ (0.0791) &$>$ (0.0381) & $=$ (0.0616)  & $>$ (0.0201) & $=$ (0.0729) & $>$ (0.0417) & $>$ (0.0357) & $>$ (0.0138)\\
vs. SMOM & $=$ (0.1081) &  $>$ (0.0491) & $=$ (0.0537) & $>$ (0.0288) & $=$ (0.0599) & $>$ (0.0319)& $>$ (0.0447)& $>$ (0.04064)\\
vs. SMOTE-IPF & $>$ (0.0111) & $>$ (0.0099)& $>$ (0.0086) &$>$ (0.0044)&  $>$ (0.0117) & $>$ (0.0056) & $>$ (0.0076)& $>$ (0.0028)\\
\bottomrule
 \end{tabular}}
\end{table*}

\subsection{Lessons learned}
To summarize the experimental study, let us try to answer the research questions formulated at the beginning of this section.\\

\noindent \textit{RQ1: What is the best parameter setting for MC-CCR, and how they impact the behavior of the algorithm?}\\

\noindent MC-CCR is a strongly parameterized preprocessing method whose predictive performance depends on the correct parameter setting. The cleaning strategy, which is a crucial element of the algorithms, has the most significant impact on the quality of MC-CCR. Based on experimental research, it can be seen that the cleaning operation is essential to produce a classifier characterized by a high predictive performance. The best parameter setting seems to be (i) cleaning by translation, (ii) proportional seed observations selection, and (iii) using sampling during the multi-class decomposition.\\

\noindent \textit{RQ2: How robust is the MC-CCR to label noise in learning data?}\\

\noindent MC-CCR is very robust to the label noise and it is marked by the smallest decrease in predictive performance depending on the label noise level or the number of classes affected by the noise. It is worth emphasizing that the proposed method is always statistically significantly better than other tested algorithms, especially for high noise levels (higher than 10\%). Additionally, the degradation of the predictive performance according to noise level increase is not so violent, and resembles a linear, not a quadratic, trend.\\

\noindent \textit{RQ3: What is the predictive performance of the MC-CCR in comparison to the state-of-art oversampling methods?} \\

\noindent MC-CCR usually outperforms the state-of-art reference oversampling strategies considered in this work, which manifested in the highest average ranks concerning all of the performance metrics.\\

\noindent \textit{RQ4: How flexible is MC-CCR to be used with the different classifiers?}\\

\noindent Based on the conducted experiments, one can observe that the proposed method works very well for both noisy and no-noise data, especially in combination with classifiers using the concept of decision tree induction (C5.0) and minimal distance classifiers ($k$-NN). However, for the other two classification methods (Na\"ive Bayes and MLP) trained based on the learning sets preprocessed by MC-CCR, the results obtained are still good. Even if statistically significant differences at the significance level $ \alpha = 0.05$ are not observed, the $p$-values remain very small, indicating substantial differences.\\

\section{Conclusion and future works}
\label{sec:con}
The purpose of this study was to propose a novel, effective preprocessing framework for a multi-class imbalanced data classification task. We developed the \emph{Multi-Class Combined Cleaning and Resampling} algorithm, a method that utilizes the proposed energy-based approach to modeling the regions suitable for oversampling, and combines it with a simultaneous cleaning operation. Due to the dedicated approach to handling the multi-class decomposition, proposed method is additionally able to better utilize the inter-class imbalance relationships. The research conducted on benchmark datasets confirmed the effectiveness of the proposed solution. It highlighted its strengths in comparison with \emph{state-of-art} methods, as well as its high robustness to the label noise. It is worth mentioning that estimated computational complexity is acceptable and comparable to the state-of-art methods.
This work is a step forward towards the use of oversampling for multi-class imbalanced data classification. The obtained results encourage us to continue works on this concept. Future research may include:
\begin{itemize}
\item Propositions of new methods of cleaning the majority observations located in proximity to the minority instances, which may be embedded in MC-CCR. Especially, other shapes of the cleaning region could be considered.
\item Application of other preprocessing methods to the proposed framework.
\item Evaluation of how robust MC-CCR is to different distributions of the label noise, as well as assess its behavior if feature noise is present.
\item Embedding MC-CCR into hybrid architectures with inbuilt mechanisms, as classifier ensemble, especially based on dynamic ensemble selection.
\item Using MC-CCR on massive data or data streams requires a deeper study on the effective ways of its parallelization.
\item Application of MC-CCR to a real-world imbalanced data susceptible to the presence of label noise, i.e., medical data.
\end{itemize}

\section*{Acknowledgement}
Michał Koziarski was supported by the Polish National Science Center under the grant no. 2017/27/N/ST6/01705.

Michał Woźniak and Bartosz Krawczyk were partially supported by the Polish National Science Center under the Grant no. UMO-2015/19/B/ST6/01597.

This research was supported in part by PL-Grid Infrastructure.
\bibliographystyle{elsarticle-num}
\bibliography{main}

\end{document}